\documentclass[letterpaper]{article}
\usepackage[preprint]{aaai2027}
\ifdefined\AgentUCTAAMAS
\else
\usepackage{url,graphicx,natbib,caption,booktabs}
\makeatletter
\@ifpackageloaded{newtxmath}{}{\usepackage{amssymb}}
\makeatother
\fi
\usepackage{amsmath,amsthm}
\usepackage{tabularx}
\usepackage{algorithm,algpseudocode}
\newtheorem{theorem}{Theorem}
\newtheorem{proposition}{Proposition}
\newtheorem{corollary}{Corollary}
\newtheorem{lemma}{Lemma}
\newcommand{\E}{\mathbb{E}}

\newcommand{\ind}{\mathbf{1}}
\newcommand{\cref}{C_{\mathrm{ref}}}

\newcommand{\leanname}[1]{{\urlstyle{tt}\path{#1}}}
\newcommand{\papertitle}{Agent-UCT: Upper Confidence Bounds Applied to Trees for Cost-Aware Agentic Workflow Optimization}

\usepackage{placeins}
\newcommand{\AgentUCTArxiv}{}
\providecommand{\Description}[1]{}
\title{\papertitle}
\author{
    Yang Li\textsuperscript{\rm 1},
    Hai Liu\textsuperscript{\rm 2},
    Dian Shao\textsuperscript{\rm 3},
    Yu Wang\textsuperscript{\rm 4},
    Xiyu Chen\textsuperscript{\rm 1},
    Di Wan\textsuperscript{\rm 1},\\
    Sergey Volkov\textsuperscript{\rm 1},
    Bozhi Wang\textsuperscript{\rm 5},
    Ziyu Sun\textsuperscript{\rm 4},
    Sihang Liu\textsuperscript{\rm 2},
    Ye Luo\textsuperscript{\rm 1}\corresponding,
    Xiaowei Zhang\textsuperscript{\rm 3}\corresponding
}

\affiliations{
    \textsuperscript{\rm 1}The University of Hong Kong\\
    \textsuperscript{\rm 2}School of Artificial Intelligence, Jiangxi Science and Technology Normal University\\
    \textsuperscript{\rm 3}The Hong Kong University of Science and Technology\\
    \textsuperscript{\rm 4}University of Science and Technology of China\\
    \textsuperscript{\rm 5}New York University\\
    Yli9919@hku.hk, studenthailiu@gmail.com, dshao@connect.ust.hk,\\
    wangyu2002@mail.ustc.edu.cn, ethanchen2277@gmail.com, dwan.subs@gmail.com,\\
    sergey.volkov@connect.hku.hk, wbz13204977@gmail.com, everglow@mail.ustc.edu.cn,\\
    sihang\_liu@jxstnu.edu.cn, kurtluo@hku.hk, xiaoweiz@ust.hk
}

\begin{document}
\maketitle
\pagestyle{plain}
\thispagestyle{plain}
\ifdefined\AgentUCTAAMAS
\else
\begin{abstract}
Workflow search pays for complete quality observations, whose execution
cost changes as reusable prefixes accumulate. AgentUCT uses an economic
model of the remaining evaluation cost to guide UCT selection while backing
up raw quality. Query--prefix accounting connects the search decisions to
reusable execution state. For finite fixed-reward trees with bounded adaptive
cold-cost estimates and successful persistent caching, we prove an
$O(\log T)$ bound on cumulative quality regret over $T$ complete evaluations.
Across five paired seeds, AgentUCT achieves $69.57\%$ test F1 versus
$68.01\%$ for UCT at a 32M weighted-token budget on a retrospective HotpotQA
panel. On GSM8K, it reaches $93.07\%$ mean accuracy at the first budget,
compared with $92.60\%$ for UCT and $91.60\%$ for AFlow. Single-factor RAG
controls show that online prices and historical continuations improve the
selected workflows at equal retained evaluation counts.
\end{abstract}

\fi
\section{Introduction}
Workflow search pays for each quality observation. A retrieval or
language-model workflow may require several calls before its quality becomes
visible. Compatible execution prefixes can be reused, so evaluating one
workflow also changes the cost of evaluating another. A search decision
therefore determines both the feedback acquired now and the execution state
available for later decisions.

The budget buys complete quality feedback. Two cheap next steps can lead
to very different bills before a score arrives, and a previously evaluated
prefix can lower those bills. AgentUCT uses an economic model to predict
this remaining work at each tree decision and subtracts its price from the
quality-and-exploration index. Raw quality is backed up through the tree:
a workflow's quality mean retains its meaning as the price of obtaining
another observation changes. Query-specific checkpoints supply the execution
state needed to update that price.

UCT recursively allocates observations through quality and exploration
\citep{kocsis2006bandit}. FLASH and EEIPU combine pipeline costs with
memoization \citep{zhang2016flash,essofi2024memoization}, while cascade
optimizers allocate stage evaluations \citep{torresi2026multistage}.
Our question is how complete-feedback prices interact with recursive UCT
allocation when each search trajectory creates its own cache. Cost guidance
changes which descendants are evaluated, so even fixed terminal rewards
produce changing internal-node returns. The allocation tree records these
returns; the execution graph records paid resources and restorable inputs.
Their joint evolution is the object of our algorithm and whole-tree analysis.

We make three contributions:
\begin{itemize}
\item Complete-evaluation pricing for UCT allocation with raw-quality backup.
Exact resource accounting connects prices to reusable execution state; a
demand-profile result identifies the cache summaries sufficient for pricing.
\item An $O(\log T)$ cumulative quality-regret bound under bounded adaptive
cold-cost estimates, finite fixed rewards and successful persistent caching.
The analysis establishes asymptotic robustness of recursive allocation to
cost guidance. Conditional histories identify both favorable and adverse
finite-budget net-value outcomes.
\item Five-seed comparisons in two search spaces: fixed RAG workflows and
generated GSM8K programs. AgentUCT achieves higher mean quality than UCT in
both and attains a higher early-budget mean than AFlow on GSM8K.
Single-factor RAG controls improve the selected quality at equal retained
evaluation counts, locating the effect in which workflows are evaluated.
\end{itemize}

These experiments test the same allocation idea under two forms of workflow
search. RAG holds the workflow grammar and executor fixed; GSM8K holds the
program proposal, repair and scoring framework fixed. Within each system,
the comparison asks which search strategy finds better workflows for the
same actual search budget.

\section{Execution State and Tree Allocation}
\label{sec:setting}
\subsection{A search tree and a cache graph}
A search tree describes legal decisions; an execution graph describes saved computation. WTB is the persistent workflow-execution and checkpoint service used by the implementation. For the theory, let $\mathcal X$ be a finite nonempty
prefix-antichain of terminal workflows and $\mathcal T$ its legal-prefix
tree, rooted at $\varnothing$. Child $pa$ extends prefix $p$ by action $a$,
and $\mathcal X(pa)$ contains its terminal descendants. For query universe
$\mathcal U$, active finite nonempty panel $\mathcal Q$ and execution
prefixes $\mathcal P$, the cache state is
\begin{equation}
G_t=(\mathcal U,\mathcal P,K_t),\qquad K_t\subseteq\mathcal U\times\mathcal P.
\label{eq:graph}
\end{equation}
Edge $(q,u)$ identifies the compatible object that restores input $q$ after
prefix $u$. This input-specific execution state is separate from tree visit
counts and quality means.

Let $\mathcal D(x)$ be the query--prefix resources demanded by terminal $x$,
including any fixed execution prefix, with nonnegative weights $w_r$.
For persistent, prefix-complete publication,
\begin{equation}
b_t^{\mathrm{cold}}(x)=\sum_{r\in\mathcal D(x)\setminus K_t}w_r,
\qquad K_{t+1}=K_t\cup\mathcal D(x_t).
\label{eq:cold}
\end{equation}
Equation~\eqref{eq:cold} charges each query's missing suffix after its
deepest checkpoint. Empty initialization and publication of successful
intermediates preserve prefix completeness. Sparse checkpoints, eviction
and failed publication change this execution model.
Nonreusable work $h_t(x)\ge0$ gives the full physical bill
\begin{equation}
b_t(x)=b_t^{\mathrm{cold}}(x)+h_t(x).
\label{eq:bill}
\end{equation}
The second term includes repeatedly incurred proposal, repair and failure
work. Such receipts remain in the actual bill. A declared token-cost unit
provides the price scale; wall time is a separate measurement.

\begin{figure*}[t]
\centering
\includegraphics[width=\textwidth]{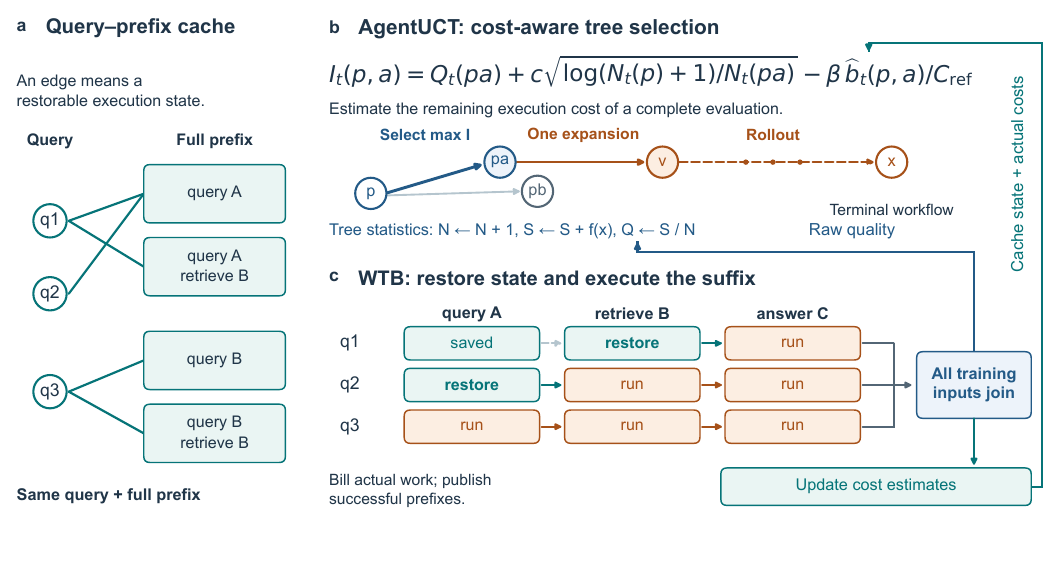}
\caption{AgentUCT prices the work required for complete feedback. (a) Query-specific compatible prefixes identify reusable state; the forecast includes the unpaid cold suffix and recurring work. (b) Prices guide branch selection. Raw scores update expanded ancestors. (c) Execution records actual bills and publishes successful prefixes. The full panel joins before price fitting and the next decision. Theorem~\ref{thm:main} establishes whole-tree robustness for a finite fixed-reward tree with bounded cold predictions and successful persistent publication. Cache membership and tree depth are illustrative.}
\Description{Compatible saved states are identified by query and full prefix. Complete-feedback costs enter a tree selection index. The diagram shows a selected branch, one expanded node, a terminal completion and upward raw-quality backup through expanded ancestors. Execution restores saved prefixes, finishes unpaid suffixes, records actual bills and publishes successful prefixes. All training inputs join before price fitting and the next decision. Separate feedback arrows carry raw quality to tree statistics and cache plus paid-cost observations to the next cost forecast.}
\label{fig:agentuct-overview}
\end{figure*}

\subsection{Allocation and complete feedback}
A complete evaluation supplies one full-panel reward. In the theoretical
model this reward is $f(x)$ for a fixed table
$f:\mathcal X\to[0,1]$. The state indexed by $t$ precedes evaluation $t$ and contains the first
$t-1$ completed returns, starting with $N_1=S_1=0$. Event $t$ inserts at most
one node and completes a terminal $x_t$. Let $v_t$ be the deepest expanded
node: the inserted node after expansion or an already expanded terminal on
a replay. Rollout extends $v_t$ to $x_t$. Every ancestor $p\preceq v_t$
receives the same raw return:
\begin{equation}
\begin{aligned}
N_{t+1}(p)&=N_t(p)+1,\\
S_{t+1}(p)&=S_t(p)+f(x_t),\qquad Q_t(p)=S_t(p)/N_t(p).
\end{aligned}
\label{eq:backup}
\end{equation}
A rollout may publish execution prefixes that never become tree nodes.
A replay of an already expanded terminal sets $v_t=x_t$ and adds its saved
outcome to every ancestor. Cache restoration during rollout leaves $v_t$
at the new expanded node. The implementation waits for all panel outcomes
before updating allocation; a candidate error that returns a score remains
scored and billed. The theorem concerns successfully completed evaluations.

Unvisited legal children are taken first. Among initialized children,
AgentUCT maximizes
\begin{equation}
\begin{aligned}
I_t(p,a)&=Q_t(pa)+c\sqrt{\frac{\log(N_t(p)+1)}{N_t(pa)}}
-\gamma\widehat b_t(p,a),\\
\gamma&=\beta/C_{\rm ref}.
\end{aligned}
\label{eq:index}
\end{equation}
The exploration coefficient is $c>0$ and the price coefficient uses
$\beta\ge0$ with calibrated scale $C_{\rm ref}>0$. Pricing reads only the
pre-decision state. MCTS-UCT sets $\beta=0$ and uses the same execution
service, making the price term the difference in the primary comparison.

\begin{algorithm}[t]
\caption{AgentUCT with complete feedback and actual-cost snapshots}
\label{alg:agentuct}
\begin{algorithmic}[1]
\Require Legal workflows, panel, cache, estimator, $c,\beta,C_{\rm ref}$
\State Initialize root statistics and actual-cost ledger
\While{the evaluation safety horizon has not been reached}
\State Traverse initialized nodes using~\eqref{eq:index}
\State Expand one unvisited legal child, when available
\State Complete a terminal by the declared rollout rule
\State Restore checkpoints and finish the whole training panel
\State Record all physical receipts; publish successful objects
\If{the complete panel returns a score}
\State Update cost observations and continuation counts
\State Back up raw quality through expanded ancestors
\State Save tree, cache and cumulative actual-cost snapshot
\EndIf
\If{actual search cost reaches the largest reporting budget}
\State \textbf{break}
\EndIf
\EndWhile
\State Retain the last complete affordable snapshot per budget
\State Select an endpoint by the declared held-out protocol
\end{algorithmic}
\end{algorithm}

\section{An Economic Model of Complete Evaluations}
\label{sec:estimation}
A branch is priced by the bill needed to obtain its next complete reward.
Given the current history and selected child, the target is
\begin{equation}
m_t(p,a)=\E_{\rho_{\mathrm{exec},t}}[b_t(X)\mid p,a].
\label{eq:actualmean}
\end{equation}
Cache membership is observed. Missing-resource prices and future demand
are predicted. A normalized reference law $\rho_{C,t}$ over legal terminal
descendants gives the cold-cost estimate
\begin{equation}
\widehat b_t^{\mathrm{cold}}(p,a)=
\sum_{x\in\mathcal X(pa)}\rho_{C,t}(x\mid p,a)
\sum_{r\in\mathcal D(x)\setminus K_t}\widehat w_{t,r}.
\label{eq:reference}
\end{equation}
Reference continuation can differ from the execution law in
\eqref{eq:actualmean}. Theoretical predictions satisfy
$0\le\widehat w_{t,r}\le M_r$ for fixed finite bounds. These bounds constrain
forecasts; actual charges are recorded after an evaluation completes.

For the finite RAG grammar, a resource ending in action $a$ has calibrated
pseudo-observation $\mu_a^0$. Online prices combine it with observed
physical costs $c_i$:
\begin{equation}
\widehat\mu_{t,r}=\frac{\mu_a^0+\sum_{i\in I_t(r)}\min\{c_i,M_a\}}
{1+|I_t(r)|},\qquad 0\le\mu_a^0\le M_a.
\label{eq:online-mean}
\end{equation}
The estimator uses query--prefix receipts when available, then pools by
prefix across queries and by action across prefixes. A restored object
supplies no new price sample, and a known zero-work slot has zero forecast.

If $n_t(p,a)$ counts complete training paths taking $a$ after $p$, the
historical next-action law is
$(n_t(p,a)+1/|A(p)|)/(\sum_b n_t(p,b)+1)$.
Uniform-complete chooses each next action with probability $1/|A(p)|$.
The induced terminal probabilities depend on the branching along each path.
With $d_q(x,K_t)$ the deepest
saved prefix depth, $m=16$ sampled legal completions estimate
\begin{equation}
\widehat b_t^{\mathrm{cold}}(p,a)=\frac1m\sum_{i=1}^m
\sum_{q\in\mathcal Q}\sum_{d_q(X_i,K_t)<j\le|X_i|}
\widehat w_{t,(q,X_{i,1:j})}.
\label{eq:sample-cost}
\end{equation}
Each sampled completion contributes with its multiplicity. Cache membership,
price means and continuation counts are frozen for this calculation, which
uses saved execution state. For the sampled estimate, let $\pi_{t,r}$ be the
empirical frequency with which the sampled paths demand a missing resource
$r\notin K_t$; resources already cached have zero incidence.
Then the same realized estimate is
\begin{equation}
\widehat b_t^{\mathrm{cold}}(p,a)=\sum_r\pi_{t,r}(p,a)\widehat w_{t,r}.
\label{eq:resourceform}
\end{equation}
Using inclusion probabilities under $\rho_{C,t}$ instead gives its
expectation. Linearity gives both sums even when resource demands are dependent.
Price learning, continuation mismatch and finite sampling each introduce
error. A growing cache removes work, while revised prices or demand can still
raise the forecast. This one-step model prices the next complete observation. Future cache
investment would require a multi-observation objective.
The GSM8K editing and execution forecast is defined in
Section~\ref{sec:gsm-main-forecast}.

AgentUCT-NextC retains only the first positive unpaid resource per query.
If its position on completion $X$ is $j_t(q,X)$, with price zero when none exists,
\begin{equation}
\widehat b_t^{\mathrm{unpaid}}(p,a)=
\E_{X\sim\rho_{C,t}}\sum_{q\in\mathcal Q}
\widehat w_{t,(q,X_{1:j_t(q,X)})}.
\label{eq:next-unpaid}
\end{equation}
On the same sampled paths, this first-unpaid estimate is bounded by the
complete estimate. It removes suffix pricing while preserving live execution
reuse. Pricing only the direct child would be less informative: full-panel
publication has already saved an initialized child prefix.

\begin{table}[t]
\centering\small
\caption{Implemented pricing and the scope of Theorem~\ref{thm:main}.}
\label{tab:estimator-scope}
\begin{tabular}{@{}p{0.24\columnwidth}p{0.41\columnwidth}p{0.27\columnwidth}@{}}
\toprule Variant & Priced work & Theory setting\\\midrule
RAG AgentUCT & All missing cold resources; recurring forecast is empty & Fixed-table execution assumptions\\
RAG NextC & First positive missing resource per query & Same cold-penalty envelope\\
RAG StaticFull & Frozen prices and law, empty pricing cache & Persistent penalty\\
GSM8K & Proposal/repair plus complete execution; uncapped point forecast & Growing programs; recurring work\\
\bottomrule
\end{tabular}
\end{table}
The implemented RAG overhead forecast is empty: ``complete'' there means
the entire unpaid cold suffix. The more general target~\eqref{eq:actualmean}
also permits recurring work. NextC shares the RAG samples and live execution
cache; StaticFull freezes its pricing reference while execution still reuses.
The finite-tree proof uses bounded cold penalties that vanish after
coverage. The experiments use $m=16$ and $\beta=2$ as fixed design settings.

\section{Whole-Tree Guarantees under Adaptive Execution Costs}
\label{sec:theory}
Let $V(p)=\max_{x\in\mathcal X(p)}f(x)$ and $V^*=V(\varnothing)$.
Fixed terminal scores still produce changing internal averages $Q_t(p)$,
because descendants are selected adaptively. Define cumulative quality loss by
\begin{equation}
 R_Q(T)=\sum_{t=1}^{T}[V^*-f(x_t)].
 \label{eq:qualityregret}
\end{equation}
Here $T$ counts complete full-panel evaluations, including free cache replays.

\begin{theorem}[Asymptotic whole-tree robustness to adaptive cold costs]
\label{thm:main}
Fix a finite nonempty prefix-antichain of terminal workflows and a training
reward table $f:\mathcal X\to[0,1]$. Assume legal incremental expansion,
unvisited-child priority, raw-return backup, successful full-panel publication
with persistent compatible, prefix-complete initial cache state, $c>0$, and fixed $\gamma\ge0$.
Completion laws may depend on the observed history, provided they are
normalized on legal terminal descendants. Resource predictions may also
adapt arbitrarily within fixed finite bounds
$0\le\widehat w_{t,r}\le M_r$. AgentUCT using the complete cold prediction
in~\eqref{eq:reference} satisfies, for every finite valid history,
\begin{equation}
 R_Q(T)\le A_D\log(T+1)+B_D.
 \label{eq:mainbound}
\end{equation}
The nonnegative constants are fixed before the history and horizon. They
depend on the finite tree, positive child-value gaps, exploration coefficient,
depth, and $H=\gamma\sum_{r\in\mathcal R} M_r$, where the finite resource universe is $\mathcal R=\bigcup_{x\in\mathcal X}\mathcal D(x)$.
\end{theorem}

This extends UCT's known asymptotic logarithmic behavior
\citep{coquelin2007bandit} to bounded history-dependent cold penalties
with explicit incremental expansion and ancestor backup.

\subsection{Two inductions over actual execution and backup}
An under-visited child's exploration bonus eventually exceeds every bounded
penalty. A finite-window count and depth induction therefore force coverage
of every terminal by a uniform threshold. Successful persistent publication
then removes all cold penalties, under any allowed price or law adaptation;
compatible shared publication can accelerate coverage.
For the second induction, define the actual backed-up deficit by
\begin{equation}
 D_t(p)=\sum_{s<t:\,p\preceq v_s}[V(p)-f(x_s)].
 \label{eq:deficit}
\end{equation}
Partitioning events by their first backed-up child gives
\begin{equation}
\begin{aligned}
 D_t(p)=d_t^{\mathrm{birth}}(p)
   +\sum_a[N_t(pa)\Delta_{pa}+D_t(pa)],\\
 \Delta_{pa}=V(p)-V(pa).
\end{aligned}
 \label{eq:decomposition}
\end{equation}
Both $D_t$ and $N_t$ precede evaluation $t$. At the root,
$D_{T+1}(\varnothing)=R_Q(T)$. The birth term is at most one; terminal deficit is zero. The exact identity
$Q_t(p)=V(p)-D_t(p)/N_t(p)$ exposes selection's internal-return bias.
Depth induction bounds the optimal child's deficit and downward bias.
After finite burn-in, index maximality bounds positive-gap visits
logarithmically. Zero-gap children contribute their subtree deficits through
\eqref{eq:decomposition}, closing the root bound. Thus the bias bound comes
from actual backup at every depth. The supplement gives explicit constants
and the Lean-checked chain.

\subsection{A comparator charged against its own cache}
Let $C_A(T)$ and $C_\pi(T)$ be cumulative search charges generated by AgentUCT
and any comparator $\pi$ using their own caches and the same fixed reward
table over $T$ complete evaluations. For fixed nonnegative physical resource
weights, let $W$ be the total weight of resources missing from AgentUCT's
initial cache, and let $\eta\ge0$.
\begin{corollary}[Own-cache cumulative net regret]
\label{cor:own-cache-net}
Under Theorem~\ref{thm:main}, if execution is cold-only, or the two
trajectories have equal total recurring overhead, then
\begin{align}
 R_{\mathrm{net}}(T)
 &=\sum_{t=1}^{T}[f(x_t^\pi)-f(x_t)]
       +\eta[C_A(T)-C_\pi(T)] \notag\\
 &\le A_D\log(T+1)+B_D+\eta W.
 \label{eq:own-cache-net}
\end{align}
\end{corollary}
Each comparator return is at most $V^*$; each own history pays for a cold
resource at most once. Arbitrary nonnegative recurring overhead adds
AgentUCT's total overhead to the bound; equal totals cancel. This term
includes recurring proposal, repair and failed work.

Penalties bounded by the same cold envelope and vanishing after coverage,
including AgentUCT-NextC, inherit this structural rate. Early allocation
still depends on the forecast, and coverage constants can grow severely
with depth~\citep{coquelin2007bandit}. Online-mean and sampled-law lemmas establish the estimator's boundedness.
The theorem analyzes fixed rewards with successful persistent publication;
the generated-program study tests allocation empirically.

For the highest-scoring observed training workflow after $T_B\ge1$
complete evaluations in a budget-retained prefix, the same bound yields
\begin{equation}
 V^*-f(\widehat x_B^{\rm train})
 \le\frac{A_D\log(T_B+1)+B_D}{T_B}.
 \label{eq:budget-training-gap}
\end{equation}
A rate in money or tokens also requires a bound on $T_B$, which includes
free replays. Validation-selected test quality and endpoint $J$ remain
empirical outcomes.

\begin{table}[t]
\centering\small
\caption{Four objectives and the evidence supporting each. Selection uses
$\beta=2$; the endpoint diagnostic uses $\lambda=0.05$.}
\label{tab:objective-map}
\begin{tabular}{@{}p{0.24\columnwidth}p{0.34\columnwidth}p{0.34\columnwidth}@{}}
\toprule Objective & Reward and charged work & Horizon and evidence\\\midrule
Training net & Best observed $f$; retained search bill & Actual affordable prefix; conditional histories\\
Cumulative net & Sum of $f$; each policy's own search bill & Equal $T$; Corollary~\ref{cor:own-cache-net}\\
Test quality & Validation-selected test F1/accuracy & Actual search cutoff; benchmark tables\\
Endpoint $J$ & Test quality minus scaled full optimization bill & Search, crossing, validation, failures/retries; empirical\\
\bottomrule
\end{tabular}
\end{table}

\section{Reuse Information and Finite-Budget Histories}
\subsection{Sufficient information for reusable cost}
\label{sec:graph-information}
Physical restoration requires compatible saved objects. Cost calculation can
use a smaller representation when the allowed query demands make some
queries indistinguishable. The following result identifies that distinction
for summaries formed by grouping residual bills.

\begin{theorem}[Demand-profile quotient]
\label{thm:demand-quotient}
Fix the family of allowed panels $\mathcal Q_1,\ldots,\mathcal Q_m$.
Group queries by their membership profiles
$q\sim r$ if $\ind\{q\in\mathcal Q_j\}=\ind\{r\in\mathcal Q_j\}$ for every $j$.
For each group $C$ and prefix $u$, store
\begin{equation}
 R_K(C,u)=\sum_{q\in C}\ind\{(q,u)\notin K\}w_{q,u}.
 \label{eq:group-residual}
\end{equation}
The bill for an allowed demand $\mathcal Q_j\times P_x$ equals
\[
 \sum_{u\in P_x}\sum_{C\subseteq\mathcal Q_j}R_K(C,u).
\]
Publishing that demand sets precisely these group--prefix cells to zero.
Conversely, merging two queries that some allowed panel separates loses the
ability to recover every panel bill from group residual sums, already with
unit weights and arbitrary binary residual states.
\end{theorem}

\paragraph{Proof.}
Every allowed panel contains either all or none of a profile group. Partitioning
the query sum therefore gives its exact bill, and publication removes precisely
the demanded cells. For the converse, swap one unit of uncached demand between
two merged queries. The group total stays fixed, while a separating panel's bill
changes. A common summary cannot recover both bills. Lean checks both directions.

\begin{proposition}[Full-panel compression]
\label{prop:graph-compression}
Fix the panel $\mathcal Q$ and resource weights. For an arbitrary cache $K$, define the residual prefix bill
\begin{equation}
 r_K(u)=\sum_{q\in\mathcal Q}\ind\{(q,u)\notin K\}w_{q,u}.
 \label{eq:residual-prefix}
\end{equation}
For a full-panel demand $\mathcal Q\times P_x$, its cold bill is $\sum_{u\in P_x}r_K(u)$. Publishing this demand sets $r_K(u)$ to zero on $P_x$ and leaves the other residual bills unchanged. If $K=\mathcal Q\times S$, the bill further reduces to $\sum_{u\in P_x\setminus S}\bar w_u$, where $\bar w_u=\sum_qw_{q,u}$. Empty initialization followed by full-panel publication preserves this rectangular form.
\end{proposition}

When every evaluation demands the full panel, all queries have the same
membership profile: per-prefix residual totals suffice for exact pricing.
This is the RAG protocol used below. Query-level checkpoint objects
are still needed for restoration, and query-conditioned receipts can still
update prices. The pricing summary and the restorable objects serve different roles.

Selective demands can require finer information. Consider two queries, prefixes
$u,v$, unit prices, and the caches
\[
 K_1=\{(q_0,u),(q_1,v)\},\quad
 K_2=\{(q_1,u),(q_0,v)\}.
\]
Both prefixes have one full-panel residual unit in either state.
On query $q_0$, their bills are $(0,1)$ and $(1,0)$. Equal-quality allocation
therefore reverses. The supplement gives a separation reachable from empty
caches through complete overlapping-panel evaluations. The minimal partition therefore follows the allowed query demands. Lossless
encodings such as bitsets preserve the same demand distinctions.

\subsection{A conditional early-budget mechanism}
\label{sec:budget-histories}
Finite-budget net value depends on both the best observed training score
and the retained search bill: $\Delta\mathcal J^{\rm tr}=\Delta f-
\lambda\Delta C/\cref$. Supplementary Appendix D constructs two reward
tables with common tariffs and completion inputs, each policy using its own
cache. At 13M, AgentUCT's quality differences from NextC are $+0.1$ and
$-0.1$, with a higher retained bill in both. Their net-value signs are
positive and negative. At 32M, favorable quality saturates at $0.5$ for both
policies, and NextC's lower bill gives higher net value. Full inputs, generated
histories and crossing identities are given in the supplement.

\section{Empirical Evaluation}
\label{sec:experiments}
\subsection{Two search spaces and a common comparison logic}
We ask whether complete-evaluation pricing improves the workflows found
within an actual search budget. RAG tests selection in a fixed workflow
space; GSM8K tests search in a generated-program space. AgentUCT--UCT
isolates the pricing term in each system. AgentUCT--AFlow compares search
strategies using the same workflow executor or program-generation framework.
The RAG mechanism controls then test which parts of the price model
contribute. We use five paired seeds $\{11,23,37,53,71\}$ and report equally
weighted means and sample SD ($n-1$ denominator).

HotpotQA distractor inputs contain supplied question contexts
\citep{yang2018hotpotqa}. The panels comprise 150 hard training questions,
300 disjoint hard validation questions and 300 hard primary test questions.
Training and validation use the official training split; primary test uses
the official development split. This original test panel was viewed during
development, so its results are retrospective held-out measurements. Within
each reported run, only training and validation scores selected workflows.

The common one-round RAG grammar contains 61,344 syntactic workflows.
Eight query actions include identity, document generation, two Query2doc
mixtures~\citep{wang2023query2doc}, decomposition, rewriting, synonymous
rewriting and HyDE~\citep{gao2023hyde}. Retrieval, fusion, evidence selection,
reranking, width and ordering complete a workflow. Generated retrieval
passages are excluded from answer evidence. Qwen3.5-4B provides all generative
roles~\citep{qwen2026qwen35}; E5-base-v2, BGE-M3 and Qwen3-Embedding-0.6B
provide encodings~\citep{wang2022e5,chen2024bgem3,zhang2025qwenembedding}.

Each trial owns its cache. The four original MCTS methods share eight-root
initialization, uniform expansion and rollout, raw-return backup and execution
service. MCTS-Random selects children uniformly; uniform-terminal search
samples complete workflows. Every training panel joins before one score,
price update or next search decision. Greedy bfloat16 generation uses fixed
groups of at most four queries. Settings are $c=1.4$, $\beta=2$ for the two
cost-aware methods and independently calibrated $C_{\rm ref}\approx1.471$M
weighted tokens. Model weights remain fixed throughout search.

At 16M, 24M and 32M actual search-cost cutoffs, up to five distinct completed
workflows with the highest training F1 enter validation. Validation F1 chooses
one endpoint, with fixed training order resolving ties. Test outcomes neither
select workflows nor update search. The primary metric is that endpoint's
test F1. Direct training-Top1 testing checks dependence on validation selection.

\paragraph{Actual-cost accounting.}
RAG bills multiply each request's actual unpadded input and output tokens by
its model's parameter count in billions. This declared resource proxy has no
wall-time or FLOP equivalence. Evaluation finishes before actual bills identify
the last complete snapshot affordable at each cutoff. Crossing work remains
charged; its later rewards and cache state are excluded from earlier snapshots.
Full optimization cost $C_{\rm opt}$ includes search, crossing evaluations,
validation and paid failures or retries. Preparation and primary-test
measurement are separate. We report $J=F-0.05C_{\rm opt}/C_{\rm ref}$, with $F\in[0,1]$, as an
endpoint diagnostic.

\paragraph{Comparators and interventions.}
Eleven declared policies share the RAG executor. AgentUCT-NextC removes suffix
pricing. AgentUCT-StaticFull jointly freezes prices, zero-history continuation
and an empty pricing cache while retaining live execution reuse.
AFlow--RAG proposes, edits and selects RAG workflows, using its Top4
score/uniform parent mixture. Qwen proposals and repairs are charged. FLASH*--RAG uses D-optimal design, linear EIPS and TPE;
EEIPU uses objective/stage-cost GPs and cached-prefix proposals; LaMBO uses
slowly moving regions and local GP-UCB; MSBO uses cascade start/advance
decisions with full-panel feedback barriers. All policies use the same
workflow grammar, models and execution service; their initialization and
allocation rules define the comparison. LaMBO's switching-cost helper is unused.

\subsection{Search quality in a fixed RAG workflow space}
\label{sec:stage-results}
AgentUCT finds the highest mean-F1 endpoints in the eleven-policy cohort at 32M
(Table~\ref{tab:direct-algorithm-coverage}). Its mean exceeds MCTS-UCT by
1.56 percentage points. The paired differences are positive for all seeds;
seed 11 reuses the same endpoint and differs by 0.067 points under fresh
numerical execution.
\begin{table}[t]
\centering\small
\caption{RAG test F1 at 32M weighted-token search cost, mean $\pm$
sample SD across five paired seeds. All rows share validation selection and
the test panel. All budgets and seeds remain in the supplementary tables.}
\label{tab:direct-algorithm-coverage}
\begin{tabular}{lr}
\toprule
Method & Test F1 (\%)\\
\midrule
AgentUCT & $69.57\pm0.28$\\
MCTS-UCT & $68.01\pm1.12$\\
AgentUCT-NextC & $68.22\pm1.96$\\
AgentUCT-StaticFull & $67.36\pm2.67$\\
MCTS-Random & $68.80\pm0.87$\\
AFlow--RAG & $67.25\pm1.57$\\
FLASH*--RAG (TPE) & $67.29\pm0.85$\\
EEIPU--RAG & $67.29\pm0.78$\\
LaMBO--RAG & $66.11\pm2.62$\\
MSBO--RAG & $67.44\pm2.53$\\
Uniform-terminal & $67.18\pm1.59$\\
\bottomrule
\end{tabular}
\end{table}
The mean numbers of retained complete evaluations are 17.6 for AgentUCT,
17.8 for MCTS-UCT and 21.0 for AFlow. Across the original four MCTS methods,
all 214 saved adaptive RAG decisions occur at the root before uniform
expansion and rollout. In AgentUCT's retained 32M histories, pricing changes
28 of 48 choices relative to maximizing the saved $Q+\mathrm{bonus}$ at the
same state. One seed-53 choice exchanges 0.0353 quality-index units for a
predicted 0.760M complete-cost saving. The price term therefore changes the observed allocation sequence.

Initialization dominates the first budget cutoff. Eight-root
initialization costs 12.287--16.626M, so at 16M the original MCTS methods
select the same workflow in each seed; only seed 23 retains adaptive
completions. For each of the five paired seeds, AgentUCT and MCTS-UCT reuse
their endpoint and test measurement at 24M and 32M, making those cutoffs
correlated measurements.
\begin{figure*}[t]
\centering
\includegraphics[width=0.96\textwidth]{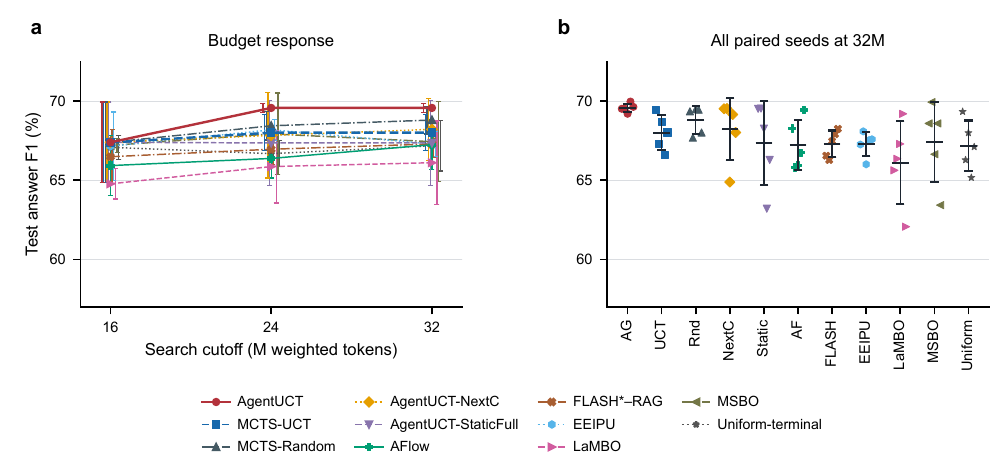}
\caption{RAG test F1 after validation selection at actual search-cost cutoffs.
(a) Eleven method means and $\pm$ one sample SD across five paired seeds
share axes. SD bars are displaced horizontally for visibility; means stay at
the true cutoffs. (b) Every paired 32M outcome; left-to-right dots are seeds
11/23/37/53/71, and black bars show mean $\pm$ SD. Endpoint reuse correlates
measurements across budgets.}
\Description{Eleven methods share test F1 versus actual search budget axes with sample SD error bars. A second panel retains all matched seed outcomes at 32 million weighted-token units.}
\label{fig:budget-quality}
\end{figure*}
Individual seeds and earlier cutoffs show reversals. MSBO leads
AgentUCT in seed 23 by 0.41 points; at 16M all five sampling/surrogate/stage
controls lead in seed 53, and LaMBO leads by 0.39 points in seed 37 at 24M.
Fifty prespecified
random workflows give $60.58\pm5.22\%$ F1, median $62.11\%$; the SD here
is across workflows. Fixed Query2doc--BM25, HyDE--E5 and RankGPT-style
implementations~\citep{sun2023rankgpt} give 61.60, 62.89 and 62.53\%.
The supplement reports every budget, seed and reference workflow.

\subsection{Search quality in a generated-program space}
GSM8K~\citep{cobbe2021verifiers} uses 150 training, 300 validation and
300 primary test questions. AgentUCT, MCTS-UCT and AFlow share one program-generation
system: Qwen3.5-4B, fixed-eight groups, Python operators, proposals, repairs,
code sandbox, scorer and counted-error histories. AgentUCT and MCTS-UCT share
progressive widening $\min(8,\max(1,\lceil\sqrt N\rceil))$, depth limit 12
and no rollout; completion pricing distinguishes them. AFlow uses its Top4
score/uniform parent mixture. AgentUCT--UCT isolates the pricing term;
AgentUCT--AFlow compares the search strategies within this common system.
AgentUCT's pre-proposal point forecast combines parent source and prompt size,
proposal/repair factors and parent execution bills, using completed training edits.
Independent fixed-path
calibration freezes $C_{\rm ref}=150{,}954.5$ raw tokens and the three budgets
in Table~\ref{tab:generated-complete-main}. Calibration scales are derived from actual model-call
charges. Up to five training-ranked workflows enter validation.
\begin{table}[t]
\centering\small
\caption{Three search strategies in the shared GSM8K program-generation system:
accuracy (\%), mean $\pm$ sample SD across five paired seeds at raw-token
budgets 6.423/12.772/19.122M.}
\label{tab:generated-complete-main}
\begin{tabular}{@{}lrrr@{}}
\toprule
Method & 6.423M & 12.772M & 19.122M\\
\midrule
AgentUCT & $93.07\pm1.52$ & $93.00\pm1.49$ & $93.07\pm1.48$\\
MCTS-UCT & $92.60\pm1.09$ & $92.60\pm0.86$ & $92.67\pm0.78$\\
AFlow & $91.60\pm1.95$ & $92.93\pm1.26$ & $93.07\pm1.40$\\
\bottomrule
\end{tabular}
\end{table}
AgentUCT has higher mean accuracy than UCT at all three cutoffs. At 6.423M,
it already reaches $93.07\%$, compared with UCT's $92.60\%$ and AFlow's
$91.60\%$. This equals its own largest-budget mean and exceeds AFlow by
1.47 percentage points at the first budget. The early paired comparison has
two wins, two losses and one tie, with a 7-point gain in seed 23.
At the largest cutoff, the paired
AgentUCT--UCT difference is
$0.40\pm0.80$ percentage points across seeds. AgentUCT leads in three,
ties in one and trails in one. Its mean equals AFlow's before rounding.
Saved counted-v3 traversals show recursive cost-induced choices. AgentUCT seed 71 selects
initial node 0 at all three cutoffs. These traversals show completion pricing changing allocation within the
generated tree.

The supplement retains all 45 GSM8K cells and full phase bills. Physical
failures, scored candidate errors, repairs and complete crossing evaluations
remain recorded in the full bills. Shared preparation costs 109,983 raw
tokens once. GSM8K uses raw tokens; RAG weights tokens by model size.

\subsection{What complete-evaluation prices change}
\label{sec:skip-validation}
Directly testing the highest training-F1 workflow gives
$69.24\pm0.74\%$ for AgentUCT and $67.37\pm0.53\%$ for MCTS-UCT at 32M.
The paired gain is $1.87\pm1.24$ points. Thus the original-panel ordering
also holds under direct training-Top1 selection. The supplement reports all sixty training-Top1 cells.

\label{sec:static-full-results}
AgentUCT's mean test F1 exceeds AgentUCT-NextC's by 1.73 points at 24M and
1.35 at 32M; three seeds improve and two tie at 32M. Direct training-Top1 selection
reduces the largest-budget gap to 0.35 points. StaticFull's paired 32M gap is
$2.21\pm2.89$ points, combining price learning, continuation updates and
cache conditioning.

\paragraph{Single-factor completion-pricing controls.}
We separately froze calibrated action prices or replaced historical
continuation sampling by uniform legal-child sampling, retaining live cache
conditioning and all other policy settings. Table~\ref{tab:single-factor-main}
compares AgentUCT with ten completed control searches on the same RAG panel.
At 32M, AgentUCT has higher F1 in three seeds and ties two against frozen
prices; against uniform continuation it has four higher scores and one tie.
All five 16M comparisons tie. At 32M, every pair retains the same number of
complete evaluations. Online price updates raise the mean by 0.47 points;
historical continuation raises it by 1.17 points relative to the corresponding
control. Equal retained counts locate the effect in which workflows are chosen.

\begin{table}[t]
\centering\small
\caption{Single-factor RAG controls. Test F1 (\%) is mean $\pm$ sample SD
over five paired seeds on the RAG panel. Budgets are M
weighted-token search units; reused endpoints correlate the cutoffs.}
\label{tab:single-factor-main}
\begin{tabular}{@{}lccc@{}}
\toprule
Method & 16M & 24M & 32M\\
\midrule
AgentUCT & $67.40\pm2.54$ & $69.57\pm0.28$ & $69.57\pm0.28$\\
Frozen prices & $67.40\pm2.54$ & $69.06\pm0.86$ & $69.10\pm0.79$\\
Uniform continuation & $67.40\pm2.54$ & $69.15\pm0.77$ & $68.40\pm1.80$\\
\bottomrule
\end{tabular}
\end{table}

Validation cost changes the $J$ ranking. Mean
$C_{\rm opt}$ is 59.003M for AgentUCT, 59.624M for MCTS-UCT and 58.319M for
NextC. AgentUCT's mean $J$ is 0.03670 above MCTS-UCT. It is 0.00978 below NextC, chiefly because of validation cost. Logical cache hit rates are 6.79\% and
6.81\% for AgentUCT and MCTS-UCT. The common executor provides reuse to
both policies. A matched cache-disabled timing study would measure the
execution saving from that reuse. Full-panel publication admits the exact
prefix-total cost summary of Proposition~\ref{prop:graph-compression}.

\section{Related Work}
\paragraph{Tree search with adaptive returns.}
UCT applies bandit selection recursively to descendant returns
\citep{kocsis2006bandit}. Its internal rewards adapt with descendant selection;
UCB1~\citep{auer2002finite} analyzes stationary arms.
Coquelin and Munos exhibit severe depth-dependent UCT transients
\citep{coquelin2007bandit}, while Shah et al. use polynomial exploration in
recursive planning~\citep{shah2020mcts}. Our result keeps the logarithmic
bonus and treats bounded history-dependent execution penalties through
finite resource coverage and actual ancestor deficits. Its fixed-table
scope is separate from stochastic planning guarantees.

\paragraph{Memoization and computation allocation.}
FLASH combines cost-sensitive acquisition with prefix caching
\citep{zhang2016flash}; EEIPU estimates memoized stage costs through GPs
\citep{essofi2024memoization}; MSBO decides when to start and advance
cascade candidates~\citep{torresi2026multistage}. LaMBO models modular switching from
the preceding configuration~\citep{lin2021lambo}. AgentUCT couples a tree's
raw-quality statistics to the complete observation cost after restoration
from any earlier compatible state. AFlow edits programs from execution
feedback~\citep{zhang2024aflow}; our common-system comparison retains its distinct
parent-selection policy. Metareasoning relates a final decision's value to
computation expenditure~\citep{hay2012selecting}. The own-cache corollary
specifies reusable charges on each policy's separate trajectory, at a common
number of complete observations.

\section{Discussion}
The economic model improves which workflows a search budget evaluates.
Equal retained counts in the RAG controls connect its gains to allocation.
RAG exercises the price term at the root; the GSM8K trajectories also show
recursive choices.

The theorem establishes asymptotic robustness: persistent reuse eventually
removes bounded cold penalties, preserving the quality-regret rate through
actual ancestor backups. Lean verifies this mathematical execution model;
the Python-to-model refinement remains unverified. Finite-budget value
depends on the quality gain covering the added bill. The conditional
histories and RAG endpoint diagnostic both expose cases favoring NextC.

The measured advantage over AFlow is concentrated at the first GSM8K
budget; the largest-budget means agree. Across the two systems, complete
prices supply current allocation information while raw-quality statistics
retain the workflow objective. This separation lets execution reuse change
the next decision without changing the meaning of earlier quality returns.

\section{Conclusion}
AgentUCT uses reusable execution state to price complete feedback while
preserving raw-quality backup. Exact accounting and asymptotic whole-tree
robustness support this design. Five-seed RAG and GSM8K comparisons show
higher budgeted mean quality than UCT; the RAG controls locate the gains in
the workflows chosen for evaluation.

\clearpage
\onecolumn
\newcommand{\AgentUCTReading}{}
\setlength{\emergencystretch}{2em}
\ifdefined\AgentUCTProofOnly
\else
\appendix
\section*{Supplementary Guide}
This supplement contains complete proofs, estimator identities, experimental
protocols and reproduction details. \ifdefined\AgentUCTArxiv
Equations, tables and figures continue the numbering of the main text.
\else
Equations, tables and figures are numbered independently from one. A \textbf{M.} prefix identifies a reference to the main manuscript.
\fi

\paragraph{Proofs and formal verification.}
Appendix~\ref{app:operational} gives the execution and backup identities.
The whole-tree quality-regret proof is in Appendix~\ref{app:rate}.
Additional cost-estimation results are in Appendix~\ref{sec:additional-estimation};
proofs for the other objectives and estimators are in Appendix~\ref{app:other}.
Appendix~\ref{app:lean} maps the results to the Lean sources and reproduction
evidence.

\paragraph{Experimental detail.}
The current generated-program study is the fifteen-run counted-v3 GSM8K
cohort in Appendix~\ref{sec:gsm-current-study}: one protocol, all forty-five
quality cells, saved checkpoints and complete phase bills. RAG protocols
are in Appendix~\ref{sec:followup-design}; the original RAG results and
algorithm specifications are in Appendix~\ref{sec:rag-full-results}.
Historical pilots and the retired GSM8K context-overflow cohort have their
own appendix, Appendix~\ref{sec:gsm-history}. The current comparison uses
the five-seed counted-v3 cohort.

\paragraph{Four different objectives.}
Table~\ref{tab:objective-map} in the main manuscript defines training net,
equal-horizon cumulative net, search-budget test quality and the full-bill
endpoint diagnostic. Training net combines the best observed training score
with its retained search bill. The quality bound counts complete training
evaluations $T$, including free replays. Its own-cache net corollary compares
cumulative returns and each policy's execution bill at the same $T$, with
cold-only costs or equal total recurring overhead. The experiments report
held-out quality of a validation-selected endpoint at actual search cutoff
$B$. The auxiliary RAG endpoint $J=F-0.05C_{\rm opt}/C_{\rm ref}$ charges full
optimization work, including crossing and validation; preparation and test
measurement are reported separately.

\fi
\section{Operational Execution and Backup Identities}
\label{app:operational}
The formal model represents a completed evaluation by an expanded path $v_t$, a terminal completion $x_t$ extending it, and its reward. The legal-tree recurrence starts with the root. An event either evaluates an already expanded terminal or inserts one new child of an existing node. Inserting a child preserves prefix closure. The allocation predicate additionally requires unvisited children first and index maximality whenever all legal children have positive counts.

\subsection{Counts, Returns, and Birth Events}
We use the main manuscript's pre-evaluation convention: state $t$ contains
events $s<t$. Lean's event-list length counts completed events, so the state
after a list of length $n$ is manuscript state $t=n+1$. Inducting over the
event list in~\eqref{eq:backup} gives
\begin{align}
 N_t(p)&=\sum_{s<t}\ind\{p\preceq v_s\},\\
 S_t(p)&=\sum_{s<t}\ind\{p\preceq v_s\}f(x_s).
\end{align}
Every backed-up terminal extends $p$, so $0\le Q_t(p)\le V(p)\le1$ for positive counts. Subtracting the return sum from $N_t(p)V(p)$ proves~\eqref{eq:deficit} and $Q_t(p)=V(p)-D_t(p)/N_t(p)$. At a terminal, prefix-antichain legality forces all its completed returns to equal its fixed value. Hence its deficit is zero.

For internal $p$, partition backed-up events by $v_s=p$ or by the first action $a$ after $p$ in $v_s$. The former can occur only once, when a non-root internal node is inserted. The latter are exactly the child visits. This proves~\eqref{eq:decomposition}, with a birth deficit between zero and one. The local action stream at $p$ contains only the latter events. If it has length $n$, its child counts are exactly the global backed-up child counts, while the parent count is $n$ at the root and $n+1$ at an already born non-root internal node. Projecting a global history onto this stream therefore preserves the index with a fixed offset $o\in\{0,1\}$. This projection accounts for progressive expansion and the history-dependent returns at internal nodes.

\subsection{Query-Specific Execution and Conservation}
For each query and terminal, saved depths form an initial segment under prefix completeness. Its largest saved depth exists because the path is finite. A demanded key is absent precisely when its depth exceeds that maximum. Summing its nonnegative weight proves equivalence between the set-difference bill and deepest-checkpoint suffix execution. Publishing a complete path preserves prefix completeness.

For one successful evaluation, disjointness of the old ledger and the newly demanded keys gives
\[
 \sum_{r\in M_{t+1}}w_r-\sum_{r\in M_t}w_r
       =\sum_{r\in\mathcal D(x_t)\setminus M_t}w_r.
\]
The current and historical graphs remain equal because both receive the same union update from equal initial states. Telescoping proves~\eqref{eq:conservation}. Nonreusable overhead adds $\sum_t h_t(x_t)$ and never enters this finite resource budget. The proof applies to the same events used for reward backup, independently of how they were selected. Distinct query identifiers make their resource sets disjoint.

For~\eqref{eq:sequentialcache}, apply the union accounting identity first to $K,A$ and then to $K\cup A,B$, and cancel the intermediate cache weight. This allows $A$ and $B$ to overlap. The operational transition publishes $\mathcal D(x_t)$ regardless of the depth of $v_t$; a prefix of $x_t$ outside the expanded ancestor chain enters the cache and leaves its tree statistics unchanged. The formal statement is \leanname{rollout_publication_without_backup}.

\section{Explicit Proof of the Tree Regret Bound}
\label{app:rate}
This section gives the constants used by the Lean main theorem. Let $K_p$ be the number of legal children of $p$, $K=\max_p K_p$, and $D=\max_{x\in\mathcal X}|x|$. Choose $\delta>0$ no larger than every positive gap $\Delta_{pa}$. Finiteness supplies such a number; when every gap is zero, any positive $\delta$ suffices. This choice is made before quantifying over histories or horizons.

The penalty in~\eqref{eq:reference} is bounded by
\[
 H=\gamma\sum_{r\in\mathcal D(\mathcal X)}M_r.
\]
For a target child count $m\in\mathbb N$, define
\begin{align}
 L_p(m)=\Biggl\lceil\max\Biggl\{&K_p+1,
 \exp\!\left(\frac{(H+2)^2m+1}{c^2}\right),\nonumber\\
 &(8K_p\max\{1,c^2\}+2)^2\Biggr\}\Biggr\rceil,
 \label{eq:window}\\
 F_p(m)&=2L_p(m).\nonumber
\end{align}

\begin{lemma}[Finite local exploration window]
\label{lem:window}
At an internal node, suppose means lie in $[0,1]$, penalties lie in $[0,H]$, unvisited children are taken first, and subsequent choices maximize the actual index with $c>0$ and parent offset $o\in\{0,1\}$. After $F_p(m)$ local child choices, every child has more than $m$ visits. Means and penalties may otherwise vary arbitrarily with the history.
\end{lemma}
\begin{proof}
Initialization visits each child within the first $K_p$ choices. Let $L=L_p(m)$ and suppose some child has at most $m$ visits at local time $2L$. Throughout the interval $[L,2L)$, its count is positive and at most $m$. The exponential term in~\eqref{eq:window} gives
\[
 c^2\log(L+o+1)>(H+2)^2m.
\]
Its exploration bonus therefore exceeds $H+2$. Its index exceeds two, since its mean is nonnegative and its penalty is at most $H$. A selected child's mean is at most one and its penalty is nonnegative, so its bonus must exceed one. Thus each selected child has prior count at most $c^2\log(2L+o+1)$. Each of the $K_p$ children can be selected at such counts at most $c^2\log(2L+o+1)+1$ times.

The square term in~\eqref{eq:window} ensures
\[
 K_p\bigl(c^2\log(2L+o+1)+1\bigr)<L.
\]
For completeness, $\log z\le2\sqrt z$ for $z>0$, $2L+o+1\le4L$, and $\sqrt L\ge8K_p\max\{1,c^2\}+2$ imply this inequality. There are $L$ choices in the interval, contradicting the count bound. All arguments concern a finite observed history; no infinite continuation is assumed.
\end{proof}

\subsection{Coverage and Disappearance of Cold Penalties}
Define
\begin{equation}
 C_0=1,\qquad C_{d+1}=1+\max_{p\in\mathcal T}F_p(C_d),
 \qquad T_{\mathrm{cache}}=C_D.
 \label{eq:coverageconstant}
\end{equation}
We prove by remaining depth that $C_d$ visits to a node suffice for every terminal below it to be visited. At depth zero, a positive visit evaluates that terminal. At an internal node, discarding its possible single birth event leaves at least $F_p(C_d)$ local choices. Lemma~\ref{lem:window} gives each child more than $C_d$ visits, and the induction hypothesis applies. The root count is the number of completed evaluations, so every terminal has been visited once $T\ge T_{\mathrm{cache}}$.

Each such visit materializes all resource pairs for its terminal and complete panel. Therefore $K_T$ covers $\mathcal D(\mathcal X)$ after $T_{\mathrm{cache}}$, and every cold forecast becomes zero. Before that time, normalization and nonnegative forecast weights bound it by $H$. The execution model supplies both boundedness before coverage and disappearance after it.

\subsection{Closing the Deficit Induction}
Set $A_0=B_0=0$. Given $A_d,B_d$, define
\begin{align}
 m_d&=\left\lceil\max\left\{1,
             (32A_d/\delta)^2,8B_d/\delta\right\}\right\rceil,
 \label{eq:biasthreshold}\\
 \tau_d&=\max_{p\in\mathcal T}\max\{T_{\mathrm{cache}},F_p(m_d)\},\nonumber\\
 A_{d+1}&=K\left(A_d+\frac{4c^2}{\delta^2}\right),\nonumber\\
 B_{d+1}&=1+K(B_d+\tau_d+1).\nonumber
\end{align}
We prove simultaneously for every finite legal event history and every node of remaining depth at most $d$ that
\begin{equation}
 D_t(p)\le A_d\log(N_t(p)+1)+B_d.
 \label{eq:subtreebound}
\end{equation}
The depth-zero statement follows from the exact terminal mean. Assume the result for all child subtrees and all history prefixes. For $n\ge m_d$, the inequalities $\log(n+1)\le4\sqrt n$ and~\eqref{eq:biasthreshold} give
\[
 \frac{A_d\log(n+1)+B_d}{n}\le\delta/4.
\]
After $\tau_d$ local choices at an internal node, every child exceeds this count threshold by Lemma~\ref{lem:window}. Global time is at least local time, so the cold penalties have also disappeared. An optimal child $a^*$ has
\[
 Q_t(pa^*)\ge V(p)-\delta/4,
 \qquad Q_t(pa)\le V(pa)
\]
for any child $a$. If a child with positive gap is selected, index maximality and the nonnegative optimal-child bonus imply that its own bonus is at least $\delta/2$. Its prior count is therefore at most
\[
 \frac{4c^2\log(N_t(p)+1)}{\delta^2}.
\]
Monotonicity of counts and of the logarithm bounds all late selections by the final parent count. Counting the at most $\tau_d$ early selections and the last late increment yields
\begin{equation}
 N_T(pa)\le\tau_d+\frac{4c^2\log(N_T(p)+1)}{\delta^2}+1
 \label{eq:childcount}
\end{equation}
for each positive-gap child. Zero-gap children contribute nothing to $N_T(pa)\Delta_{pa}$.

Insert~\eqref{eq:childcount} and the child induction hypotheses into~\eqref{eq:decomposition}. Use $0\le\Delta_{pa}\le1$, $N_T(pa)\le N_T(p)$, the birth bound one, and at most $K$ children. This gives exactly~\eqref{eq:subtreebound} with the constants in~\eqref{eq:biasthreshold}. At the root, $N_T=T$ and $D_T=R_Q(T)$, proving Theorem~\ref{thm:main}. All constants were fixed before the arbitrary history and are independent of $T$.

Ordinary UCT uses the same induction with $H=0$ and $T_{\mathrm{cache}}=0$. Proposition~\ref{prop:next} transfers that bound to the legacy direct-edge estimator without adding a cache-coverage delay: initialized direct edges are already published. The current first-unpaid estimator instead uses its domination by the complete cold bill and the coverage argument. For the full-cold policy, the finite instance-dependent coverage constants can be enormous at greater depth.

\subsection{Reading the Lean Proof in Dependency Order}
The following steps identify the actual proof obligations. The English guide
\path{theory/lean/AgentUCTProofGuideEN.lean} follows the same order; its
complete-proof companion \path{theory/lean/AgentUCTCompleteProofsEN.lean}
contains all definitions and proof bodies in dependency order, with English
step comments and source-module locations.
Each \texttt{\#check} displays a declaration and its explicit hypotheses.
\begin{enumerate}
\item The backup recurrence gives counts and return sums.
\leanname{operational_mean_deficit} converts the actual mean to $V-D/N$ for
positive $N$. The parent partition retains the single birth deficit and
the local-stream offset; these are operational identities.
\item Nonnegative normalized completion laws and a fixed price envelope give
a history-independent penalty bound. It constrains predictions. Physical receipts are recorded after model calls.
\item \leanname{finite_window_explores} proves the finite-window contradiction.
\leanname{finite_explicit_window_explores} supplies the explicit exponential
and square thresholds. \leanname{actual_children_explored} transfers the
statement to the actual local action stream.
\item \leanname{actual_subtree_coverage} inducts on remaining depth, retaining
the possible birth visit. \leanname{actual_restorable_coverage} converts
complete terminal publications to persistent executable resources.
\item \leanname{coldForecastPenalty_after_coverage} derives zero penalties
from empty missing-resource sets. Bounded penalties alone establish
exploration, and do not discharge this obligation.
\item \leanname{logarithmic_deficit_small} divides the child induction bound
by its positive count to obtain $D/N\le\delta/4$. The optimal child then
has mean at least $V(p)-\delta/4$.
\item \leanname{actual_suboptimal_child_count} combines this mean bound,
zero penalties, index maximality and the bonus-square identity. It retains
the early threshold and the last increment.
\item \leanname{parent_deficit_log_bound} sums the parent decomposition;
\leanname{actual_subtree_log_deficit} closes the remaining-depth induction
simultaneously for every history prefix. Root count and deficit identities
give the cumulative training-quality bound.
\item \leanname{onlineMean_envelope} bounds the online price mean.
\leanname{sampledLaw_mass} retains multiplicity and requires a positive
sample count. They discharge estimator premises of
\leanname{online_mean_shared_root_log_bound}; fixed scores and successful
persistent publication remain execution assumptions.
\end{enumerate}

\begin{table*}[t]
\centering\small
\caption{Proof-to-claim map. The operational tree rate uses the actual-backup chain. The local recurring-cost counterexample and conditional confidence results have their own hypotheses.}
\label{tab:lean}
\begin{tabularx}{\textwidth}{@{}>{\raggedright\arraybackslash}p{0.27\textwidth} >{\raggedright\arraybackslash}X@{}}
\toprule
Paper result & Representative Lean declaration in namespace \texttt{AgentUCTPaper}\\
\midrule
Full-panel compression & \leanname{full_panel_residual_quotient}, \leanname{full_panel_history_rectangular}\\
Demand-profile compression & \leanname{demand_group_cost_quotient}, \leanname{demand_group_publication}, \leanname{demand_separating_merge_loses_information}\\
Query-selective information loss & \leanname{GraphWitness.no_count_only_optimal_policy}, \leanname{GraphWitness.budget_feasibility_reverses}\\
Actual reward backup & \leanname{replayBackup_visits}, \leanname{replayBackup_mean}, \leanname{operational_mean_deficit}\\
Graph accounting and suffix & \leanname{replayMetered_cold_conservation}, \leanname{coldWork_eq_deepest_suffix}\\
Tree/stream correspondence & \leanname{localBefore_parent_count}, \leanname{projected_index_max}\\
Exploration and coverage & \leanname{actual_children_explored}, \leanname{actual_restorable_coverage}\\
Online-price tree rate & \leanname{checkpoint_conditioned_shared_root_log_bound}, \leanname{online_mean_shared_root_log_bound}\\
Fixed-price special case & \leanname{cold_forecast_uniform_root_log_bound}\\
Shared fixed-panel family & \leanname{shared_cache_uniform_root_log_bound}, \leanname{fixed_panel_family_log_bound}\\
Completion index and budget & \leanname{completion_index_error}, \leanname{own_cache_budget_coverage}\\
Same-state pricing benefit & \leanname{priced_choice_tradeoff}, \leanname{physical_saving_certificate}, \leanname{actual_net_gain_certificate}\\
Exclusive trajectory costs & \leanname{own_cache_cost_difference}\\
Own-cache cumulative objective & \leanname{own_cache_policy_regret_bound}, \leanname{own_cache_cost_difference_common_overhead}\\
Recommendation and endpoint & \leanname{validation_shortlist_gap}, \leanname{equal_quality_positive_endpoint_gap}\\
Full cost and its errors & \leanname{referenceColdCost_resource_form}, \leanname{full_reference_cost_error}\\
Prediction and optimism & \leanname{full_cost_mean_unique_mse_minimizer}, \leanname{selected_net_gap_from_cost_radii}\\
Legacy direct-edge reduction & \leanname{next_edge_allocation_iff_uct}, \leanname{next_edge_uniform_root_log_bound}, \leanname{legacyContinuationCold_common_shift}\\
First unpaid execution & \leanname{nextUnpaidBill_le_complete}, \leanname{dominated_cold_shared_root_log_bound}\\
Coupled trajectories & \leanname{resolvedTreeHistory_valid}, \leanname{next_edge_coupled_history_eq_uct}, \leanname{next_edge_coupled_state_eq_uct}\\
Local recurring-cost failure & \leanname{next_edge_linear_regret}\\
\bottomrule
\end{tabularx}
\end{table*}

\ifdefined\AgentUCTProofOnly\else
\section{Additional Cost Estimation Results}\label{sec:additional-estimation}
\label{app:cost-estimation}
\subsection{The Full Evaluation Target}
The conditional completion mean $m_t$ is defined in Section~\ref{sec:estimation}.

For any scalar $z$, the exact identity
\begin{equation}
 \E[(b_t(X)-z)^2]=\E[(b_t(X)-m_t)^2]+(z-m_t)^2
 \label{eq:mse}
\end{equation}
shows that this full conditional mean uniquely minimizes squared prediction error. The prediction target is the next complete reward-bearing evaluation, including its nonreusable work.

When the actual continuation law or bills are unknown, a declared reference law gives the computable surrogate
\begin{equation}
 \widehat b_t(p,a)=\sum_{x\in\mathcal X(pa)}\rho_{C,t}(x\mid p,a)
  \left[\sum_{r\in\mathcal D(x)\setminus K_t}\widehat w_{t,r}+\widehat h_t(x)\right].
 \label{eq:fullestimate}
\end{equation}
Uniform legal-terminal enumeration is one choice of $\rho_C$. A continuation model matching actual traversal and rollout is better justified for predicting~\eqref{eq:actualmean}. A frozen per-action calibration mean can supply $\widehat w_{(q,u)}$, but it discards query and prefix variation. More detailed predictors should condition on relevant execution features when supported by calibration data.

For $x=(a_1,\ldots,a_L)$, let $d_q(x,K)$ be the largest saved prefix depth, or zero if none is saved. The predicted cold bill is
\[
 \sum_{q\in\mathcal Q}\sum_{j>d_q(x,K)}\widehat w_{t,(q,x_{1:j})}.
\]
Current cache membership is observed. The quantities to predict are future resource demand and the cost of executing missing resources.

Checkpoint contents refine the resource-price model. If an earlier observation
of $(q,p)$ certifies that retrieval has halted, every later query/retrieval slot
under $p$ has zero model-token cost for that query, including slots whose own
checkpoints have never been materialized. We therefore set
\begin{equation}
 \widehat w_{t,r}=\begin{cases}0,&\text{a prior state certifies a skipped slot},\\
 \widehat\mu_{t,r},&\text{otherwise},\end{cases}
 \qquad 0\le\widehat w_{t,r}\le M_{a(r)},
 \label{eq:state-price}
\end{equation}
where $M$ is a fixed resource-price envelope. Only observations available
before selection enter this prediction. Each sample consults query-specific
zero-work observations. The membership-only variant retains the online mean
outside physical hits. The main loop-zero grammar has no extra retrieval rounds.

This distinguishes two effects of observed execution state. Restorability
removes materialized work from the bill. A known stop predicts zero model work
for unmaterialized descendants; it creates no cache hit. The specified no-reuse control would
retain the same observed stop facts while rerunning every non-skipped call.
Those cache-disabled measurements remain deferred
(Section~\ref{sec:followup-design}); the two active price/law controls
retain execution reuse.
Token-free state advancement still takes measured CPU time.

\begin{proposition}[Prediction and law error]
\label{prop:error}
Suppose $|\widehat w_{t,r}-w_r|\le e_r$, $|\widehat h_t(x)-h_t(x)|\le e_h(x)$, and $|b_t(x)|\le B_{\max}$ on the child subtree. With $L_t=\sum_x|\rho_{C,t}(x)-\rho_{\mathrm{exec},t}(x)|$,
\begin{equation}
 |\widehat b_t-m_t|\le
 \E_{\rho_{C,t}}\!\left[\sum_{r\in\mathcal D(X)\setminus K_t}e_r+e_h(X)\right]
 +B_{\max}L_t.
 \label{eq:costerror}
\end{equation}
\end{proposition}
The complete horizon removes suffix truncation; the bound separates the
resource-price and continuation-law terms.

\subsection{What Cache Accumulation Changes}
The graph has two distinct effects. Physical membership determines which query-prefix states can be restored and which work is billed. The estimator reads the same membership to remove that work from each candidate's predicted bill. Forecast weights and the reference law determine selection scores; published objects determine physical cache hits. Executing the selected completion changes the graph and thereby changes later forecasts.

Write $c_w(K,A)=\sum_{r\in A\setminus K}w_r$. Sequential publication satisfies
\begin{equation}
 c_w(K,A\cup B)=c_w(K,A)+c_w(K\cup A,B).
 \label{eq:sequentialcache}
\end{equation}
Thus the full bill from the pre-rollout snapshot already accounts for work saved as that rollout executes. The implementation chooses a symbolic completion, executes it, and publishes physically completed prefixes.

There is also a precise effect on later forecasts. Holding the candidate, law, and weights fixed, publishing a completed demand $A$ gives
\begin{equation}
 \widehat b^{\mathrm{cold}}(K)-\widehat b^{\mathrm{cold}}(K\cup A)
 =\E_{\rho_C}\!\left[\sum_{r\in(\mathcal D(X)\cap A)\setminus K}\widehat w_r\right].
 \label{eq:cacheforecast}
\end{equation}
The right side is the expected weight of newly saved resources shared with a later completion. It is nonnegative. Comparing successive iterations without fixing the law or predictor need not be monotone, because those quantities can also change.

Equation~\eqref{eq:fullestimate} prices the next evaluation under the accumulated cache state. Valuing an action's cache investment over later evaluations requires a continuation value on joint tree statistics and cache state, an explicit remaining budget and an endpoint objective. Equation~\eqref{eq:cacheforecast} supplies the one-step reduction in predicted work for that extension. Richer query-prefix features and calibrated cost lower bounds are possible additions to the current historical full-bill model.

\subsection{Two Short-Horizon Estimates}
Let $\widehat d_t(p)$ be the predicted cold demand of prefix $p$. The historical
\texttt{immediate} implementation, labeled next-edge in earlier experiments,
estimates the marginal cold cost of the tree's direct child:
\begin{equation}
 \widehat b_t^{\mathrm{next}}(p,a)=\max\{0,\widehat d_t(pa)-\widehat d_t(p)\}.
 \label{eq:next}
\end{equation}
It is one possible cost approximation. Write a full estimate as $\widehat b_t^{\mathrm{next}}+\tau_t(p,a)$. If the omitted tail $\tau_t(p,a)$ is common to all candidates, subtraction changes every index by the same constant and preserves the maximizer set. Otherwise the pairwise index difference changes by $\gamma(\tau_t(p,a)-\tau_t(p,b))$. A small common tail is unnecessary; small variation across candidates is the relevant condition for preserving comparisons.

\begin{proposition}[Immediate-child reduction]
\label{prop:next}
Under successful persistent full-panel execution, every visited child $pa$ has its complete prefix demand cached. For nonnegative resource predictions, its penalty in~\eqref{eq:next} is zero. This immediate-child allocation rule is equivalent to ordinary UCT allocation on the same history. Their complete trajectories agree when the initial state is identical and a shared resolver couples expansion, rollout, tie-breaking, evaluation outcomes, and stopping. In the fixed-table setting, this variant inherits the ordinary-UCT instance-dependent quality bound.
\end{proposition}
The reduction leaves qid-prefix caching and reward backup in place. It concerns
the direct-child formula~\eqref{eq:next} among initialized children. Expansion
and rollout can still execute uncached suffixes, whose predicted cost can be
positive. Estimating the next unpaid execution instead gives a different rule.

For
example, a saved child followed by an uncached edge priced four gives zero
in~\eqref{eq:next} and four in~\eqref{eq:next-unpaid}.
Combining this reduction with own-cache accounting gives the immediate-child result: under the same finite fixed-table conditions, and cold-only costs or equal total recurring overhead,
\begin{equation}
 R_{\mathrm{net,next}}^\pi(T)
 \le A_D^{\mathrm{UCT}}\log(T+1)+B_D^{\mathrm{UCT}}+\eta W.
 \label{eq:nextnet}
\end{equation}
This equal-horizon bound follows from ordinary-UCT quality control and finite total cold work.

For a local comparison evaluated at the same state, omitted-tail variation gives a separate approximation bound. If the chosen action's omitted tail exceeds the comparator's by at most $\omega_t$, direct subtraction gives $R_{\mathrm{full}}^{\mathrm{local}}(T)\le R_{\mathrm{next}}^{\mathrm{local}}(T)+\eta\sum_{t\le T}\omega_t$. Hence a logarithmic proxy bound transfers when the accumulated tail variation is also logarithmic. Constant nonzero variation only supplies a linear error allowance. This relation compares local choices at one state; Equation~\eqref{eq:nextnet} compares policies with their own caches.

The historical \texttt{continuation\_mean} averages
$\max\{0,\widehat d_t(x)-\widehat d_t(p)\}$ over child terminals. Prefix-complete persistent caching implies $\widehat d_t(x)\ge\widehat d_t(p)$, so clipping is inactive. With a normalized reference law this equals the full cold mean minus the common parent bill. It therefore preserves the full-cold selection ordering. The implemented \texttt{acquisition\_mean} uses the full bill and includes configured future overhead. The common-shift identity assumes prefix-complete caching and the same omitted parent bill across actions.

Omitted recurring cost has a different consequence from finite cold work. A verified two-arm example has deterministic rewards $0.8,0.7$, immediate costs $0.4,0.1$, and full costs $0.4,0.4$. With cost weight one, its full-net regret is at least $T/10-40\log T-2/5$ under the initialized next-edge UCB rule with bonus $\sqrt{2\log(t+1)/N}$, where $t$ counts previous pulls. The second arm appears cheaper because its recurring tail of $0.3$ is omitted, although its full net value is smaller. This local recurring-cost example disproves a general logarithmic full-net guarantee for next-edge estimation. This is a recurring-cost bandit construction with its stated bonus and constants.

\subsection{When Is the Index a UCB?}
For a stationary bandit with known fixed action cost $b_a$, subtracting $\gamma b_a$ from the reward index is exactly UCB on the shifted returns $Y_{a,s}-\gamma b_a$. That stationary reduction assumes a fixed action price. Graph bills depend on execution history, and UCT returns depend on descendant selection.

If a cost estimate has a valid radius $e_a$, a more principled optimistic net index uses $\underline b_a=\max\{0,\widehat b_a-e_a\}$. Given a valid reward upper bound $\mu_a\le Q_a+r_a$, it follows that
\begin{equation}
 \mu_a-\gamma b_a\le Q_a+r_a-\gamma\underline b_a.
 \label{eq:optimism}
\end{equation}
With two-sided reward error for the selected action, maximizing this index yields selected net gap at most $2r_a+2\gamma e_a$. Net optimism here requires valid reward and cost intervals. The implemented estimator uses point forecasts; Theorem~\ref{thm:main} establishes its robustness through deterministic ancestor-backup analysis.

\section{Conditional Finite-Budget Histories and Accounting Inputs}
\label{si:conditional-budget-histories}
\label{si:candidate-budget-histories}
This section supplies the task class and construction for
Theorem~\ref{thm:conditional-budget}. It uses the native finite workflow
grammar and a homogeneous panel of $n=150$ queries. Prices are stationary,
evaluations succeed, prefix publication persists, and recurring model-billed
overhead is zero. The reward tables and static invoice counts are constructed
inputs. They specify a finite accounting task, with no assumption about the
workflow selected by either policy.

\begin{theorem}[Conditional history example]
\label{thm:conditional-budget}
There is a nonempty stationary accounting-task family with 150-query panels,
common legal MC16 inputs, role-valid invoice counts and initial-price error
at most $10^{-6}$ on covered actions. For $\beta=2$, $c=1.4$,
$\lambda=0.05$, $B=13\mathrm{M}$, $\tau=10^{-7}$, cap at least eleven
and $\cref\in[1470000,1471000]$, the native ordered allocator generates
both policies' histories from eight common charged seeds, updating each
policy's own cache and receipts. Then
\begin{align}
 T_B^{\rm AgentUCT}&=10,\qquad T_B^{\rm NextC}=9, \notag\\
 \mathcal J_{\rm AgentUCT}^{\rm tr}-\mathcal J_{\rm NextC}^{\rm tr}
 &=\sigma\,0.1-\lambda\frac{2831.6322489}{\cref},
 \label{eq:conditional-budget-net}
\end{align}
where $\sigma=1$ and $\sigma=-1$ select favorable and adverse reward tables
with common tariffs, controls, priors, tape and cutoff. The net signs are
respectively positive and negative.
\end{theorem}

\subsection{Quote projection}
\label{si:quote-projection}
\begin{lemma}[Per-query quote equivalence]
At a fixed pre-decision state in the exact-arithmetic execution model, let
$s_i(q)$ be the bill returned by the per-query suffix loop for query $q$
and sampled completion $i$, using either Complete or NextC. With sixteen
samples, the full-panel quote is exactly
\[
 \widehat b=\frac1{16}\sum_{i=1}^{16}\sum_{q\in\mathcal Q}s_i(q).
\]
For the homogeneous $150$-query construction, the product cache and common
completion inputs give $s_i(q)=s_i$, hence
$\widehat b=(150/16)\sum_i s_i$. For its common completion tape this reduces
to $150s_1$. Each identity uses the policy's own cache.
\end{lemma}
\paragraph{Proof.}
The per-query loop restores the deepest available prefix and sums the
remaining charged slots. Complete retains the whole suffix; NextC ends at
the first positive-price slot. Summing these loop outputs over queries and
averaging over samples gives the first identity. The product cache and
homogeneous prices make all query outputs equal in the construction;
counting the $150$ queries and sixteen identical completions gives the
reduction. The Lean projection keeps both factors. The online-price error
bound is handled separately in the choice certificates, including NextC's
zero/positive support condition.

\subsection{Execution, retention and training net value}
\label{si:conditional-retention}
Each resource is a query--prefix pair. A newly materialized resource publishes
a fresh injective immutable receipt, including a formal row for a zero-price
pure step. Chronological replay reconstructs its policy's saved resources,
seen IDs, price pools and paid sum. Query sums and the mean of sixteen legal
completions give the forecast. Actual fees are not clipped; caps constrain
predictions and observed pools. The construction keeps clipping inactive.

Both modes use ordered unvisited-child priority, first ordered visited
maximum, $c\sqrt{\log(N+1)/N_a}$ exploration, one-child expansion and raw
ancestor backup. All eight query roots receive distinct legal nine-action
seed workflows and their full charges. Seeds use width $k2$ and answer $m2$;
HyDE uses E5 and the other roots use BM25. One query action is the legal
identity. Each policy subsequently uses only its own completed receipts.

Let $L_m$ be the length of the completed raw history, including its crossing
evaluation, and $C_m(t)$ the actual search charge of its first $t$ evaluations.
For $\tau\ge0$ and $\cref>0$, define
\begin{align}
 T_B^m&=\max\{t\in\{0,\ldots,L_m\}:C_m(t)\le B+\tau\},
 \label{eq:conditional-retention}\\
 \mathcal J_m^{\rm tr}(B)
 &=\max_{1\le s\le T_B^m}f(x_s^m)-\lambda C_m(T_B^m)/\cref,
 \quad\lambda\ge0.
 \label{eq:conditional-training-net}
\end{align}
The budget window below retains all eight seeds, so $T_B^m\ge8$.
A complete evaluation can replay an existing terminal. It changes raw tree
visits and averages even when it publishes no resource and adds no model
charge. $T_B$ is therefore an evaluation count, not a distinct-workflow count.
The generator checks actual cumulative charge after completing each
evaluation. Reporting then takes the last affordable complete prefix; the
crossing evaluation's rewards, receipts and cache state are not used in it.

\subsection{Task gap, overlap and forecast-error conditions}
Let $b$ denote the HyDE--BGE tariff, $h$ the initialized HyDE--E5 tariff,
$e$ the normal-E5 tariff, and $q$ the HyDE--Qwen tariff. Width and answer
tariffs are $w_2,w_4,a_2,a_6$, and reranking costs $r$. Write
\[
 \kappa=\beta n/\cref,\quad A=w_4+a_2,\quad F=w_4+r+a_6,
\]
where $\kappa$ is the index coefficient for a per-query tariff; the
panel-bill coefficient remains $\gamma=\beta/\cref$. For $c>0$, put
\[
 D_2=c(\sqrt{\log10}-\sqrt{(\log10)/2}),\qquad
 D_3=c(\sqrt{\log11}-\sqrt{(\log11)/3}).
\]
Require $0<\delta<g\le1$, $\kappa,b,w_4>0$, nonnegative remaining tariffs,
and the input margins
\begin{align}
 0&<\kappa(w_4-b)<\delta,\notag\\
 \delta&<\kappa(F-(b+A)),\notag\\
 \delta+D_2&<\kappa F,\quad \delta+D_3<\kappa F,\notag\\
 g&<\kappa(w_4-b)+D_2.
 \label{eq:conditional-root-margins}
\end{align}
The conditions specify tariffs, initial seed quality and exploration.
The proof derives the allocation sequence from these inputs.

Positive-action priors differ from their tariff by at most
$\varepsilon>0$; zero-tariff priors are exactly zero, and the positive
tariff floor exceeds $\varepsilon$. Complete own-history receipts preserve
this support and error bound. A suffix has at most ten actions. Summation
over queries and MC16 averaging therefore bound the same-tape index error
by $E=10\kappa\varepsilon$. Require $2E$ below each strict residual
\begin{equation}
\begin{gathered}
 \delta,\quad\delta-\kappa(w_4-b),\quad\kappa(w_4-b),\quad
 \kappa(F-(b+A))-\delta,\\
 \kappa F-\delta-D_2,\quad\kappa F-\delta-D_3,\quad
 \kappa(w_4-b)+D_2-g.
\end{gathered}
\label{eq:conditional-error-margins}
\end{equation}
This is a same-path price-error bound. Continuation-law mismatch contributes
a separate error term, as in Proposition~\ref{prop:error}.

\subsection{The exact common legal task resolver}
The resolver is \texttt{AgentUCTNativeBudget.budgetTape} in
\texttt{NativeBudgetResolver.lean}. Before eight evaluations it supplies
the explicit legal seed preference. After initialization, HyDE expansion
orders available retrievers BGE, E5, then Qwen; branch-B expansion orders
normal E5, then BM25. It uses the first listed available candidate, with
the first available candidate as fallback.

The rollout goal is the first of four terminal paths extending the current
prefix: HyDE--BGE/$k4$/$m2$, initialized HyDE--E5/$k2$/$m2$,
HyDE--Qwen/$k2$/$m6$, and branch-B/normal-E5/$k4$/$m2$.
They use native selection index 0, stop, reranker index 0 and context index 0;
top-$m$ and answer use matching index 0 for $m2$ or 2 for $m6$.
At a root comparison, all sixteen HyDE samples use its BGE/$k4$/$m2$
path. Every other root uses its normal-BM25/$k4$/rerank/$m6$ path, with
selection index 0, width index 1, reranker index 1, top-$m$ index 2,
context index 0 and answer index 2. Non-root MC goals use the first
extending four-path rollout goal or \texttt{finish}. The
\texttt{safeCompletion} guard accepts only a terminal goal extending the
requested prefix, and otherwise returns the native legal completion.

This common tape supplies task inputs independently of mode and reward
table. Each sample has positive same-history smoothed support. The supported tape is the declared input of this deterministic construction.
At the priced checkpoints used for the early 13M window, all eight roots
are present and each has an unexpanded child. Traversal performs the root
comparison and stops at that child. The source's zero-coefficient guard
skips estimator calls at initialization, expansion and rollout. The 32M
continuations also compare descendants below the root; their per-step
certificates are covered in Section~\ref{si:conditional-32m-continuation}.

\subsection{Generated histories and the strict reporting window}
Let $u_j$ be the eight query tariffs and define
\[
 I=\sum_{j=1}^8u_j+8(w_2+a_2)+h,\qquad X=q+w_2+a_6,
 \qquad B=n(I+b+A+A/2).
\]
For $n,A>0$, $b\ge0$, $e<b+A/2$ and $\tau\ge0$, assume
\begin{equation}
 \tau<n(X-A/2),\qquad\tau<n(e+A/2).
 \label{eq:conditional-window}
\end{equation}
The evaluation cap is at least eleven. The native allocator then generates
eight seeds followed, in complete mode, by cold HyDE--BGE/$k4$/$m2$,
warm HyDE--E5/$k2$/$m2$, and cold HyDE--Qwen/$k2$/$m6$. NextC follows
the seeds with cold branch-B/normal-E5/$k4$/$m2$ and cold
HyDE--BGE/$k4$/$m2$.
\begin{equation}
\begin{array}{c|c|c|c}
 \text{mode}&T_B&\text{retained bill}&\text{raw bill}\\\hline
 \text{complete}&10&n(I+b+A)&n(I+b+A+X)\\
 \text{NextC}&9&n(I+e+A)&n(I+e+2A+b)
\end{array}
\label{eq:conditional-bills}
\end{equation}
The warm replay expands a genuine E5 tree child and changes visits and raw
means, while leaving resources, receipt means and charges unchanged.
The next choice follows from the changed tree. Both final cold evaluations
cross $B+\tau$ and remain in their raw histories. Their future rewards and
publications do not enter the earlier retained endpoints.

\subsection{Reward tables and two net directions}
The initial branch-B seed has quality $\delta$. In the favorable task,
the new HyDE--BGE completion has quality $g$ and the branch-B completion
has quality $\delta$. In the adverse task their qualities are zero and
$g$. All other terminal qualities are zero. Both tables have optimum $g$.
Each mode is regenerated on each table with the same initial seeds,
tariffs, controls, priors, tape and cutoff. Their retained best qualities
are $(g,\delta)$ and $(\delta,g)$ respectively, hence
\begin{equation}
 \Delta\mathcal J^{\rm tr}
 =\pm(g-\delta)-\lambda n(b-e)/\cref.
 \label{eq:conditional-net-general}
\end{equation}
The strict bracket
$\delta-g<\lambda n(b-e)/\cref<g-\delta$ gives opposite signs.
Shared-prefix charges cancel only where the two own histories actually
overlap. The exclusive BGE/E5 demand contributes their retained bill
difference, so a quality gain must exceed that extra charge to improve net.

\subsection{A nonempty role-valid natural-invoice family at $\beta=2$}
Let the per-field weights be
\[
 \boldsymbol\omega=(G,G,Q,0,E,H,0,Q),\quad
 G=4.205751296,\ E=0.10948224,\ H=0.568805377,\ Q=0.595776512.
\]
For each action $a$, its bill is
$p(a)=\sum_{f=0}^7\omega_f z_{a,f}$ with
$z_{a,f}\in\mathbb N$. On the actions observed in the histories and all
root samples, the fixed covered invoices give
\begin{align*}
 u_{\rm identity}&=0,&u_{\rm other}&=512G,&h&=37E,\\
 b&=38H,&e&=25E,&q&=39Q,\\
 w_2&=512G,&w_4&=320G,&a_2&=1307G,\\
 a_6&=2048G,&r&=16384G.
\end{align*}
The control values are $c=7/5$, $\beta=2$, $\lambda=1/20$,
$\delta=2/5$, $g=1/2$, $\tau=10^{-7}$ and cap 256.
The root and resource-window margins hold. There is an
$\varepsilon\in(0,1/2)$ satisfying all error margins; positive priors use
tariff plus $\varepsilon/2$ and covered caps use tariff plus
$\varepsilon$. The initial observed BGE forecast error is genuinely nonzero.

For actions outside the covered domain, allow any natural invoice satisfying
\texttt{StaticUnusedInvoices}: nonzero fields belong to the native action's
model roles, every field count respects its source bound, and its weighted
bill is within its static cap. Completed priors equal these unused bills;
unused prediction caps are the static caps. These choices preserve agreement
on all observed actions and initial covered priors/caps. One token in every
allowed unused role supplies a feasible completion, establishing nonemptiness.
All thirty normal-retriever role/cap rows are included in the static table.

Chronological equality preserves complete receipt state, resource means,
cache membership and actual paid sums. It also preserves all eight root
MC16 scores and their first ordered maximum at every priced checkpoint.
The margin theorem first derives the candidate choices; these equalities
transfer them and native generation derives all four full histories.
Neither completed-run equality nor a winner is an input assumption.
Arbitrarily large unused real prices are inert under fixed covered agreement;
the role/cap/count-valid natural family just defined is more restricted.

\subsection{Independent reference and exact constructed bills}
The reference uses three distinct legal paths with query roots 0, 4 and 7,
each evaluated from its own empty cache. Each cold path has 2331 generator
tokens. Their median times 150 gives
\begin{equation}
 \cref=150(2331G)=1470540.9406464.
 \label{eq:conditional-reference}
\end{equation}
This statistic is defined independently of the task winners. The constructed
cutoff is $B=12984796.0484025$. Its field-total accounting is
\begin{center}
\begin{tabular}{lrrr}
\toprule
Mode & Retained evaluations & Retained bill & Raw bill\\
\midrule
AgentUCT & 10 & 12471589.2465081 & 14090083.0367673\\
AgentUCT-NextC & 9 & 12468757.6142592 & 13498413.4086969\\
\bottomrule
\end{tabular}
\end{center}
The retained bill difference is $2831.6322489>0$, and
\[
 \Delta\mathcal J^{\rm tr}=\pm0.1
       -\frac{0.05\,(2831.6322489)}{1470540.9406464}.
\]
The favorable value is positive and the adverse value negative. These
invoice sums instantiate the accounting task at
$\cref=1470540.9406464$. Both raw bills lie below 32M. The next two
subsections extend the inputs to the saved reference and the complete
32M continuation.

\subsection{A reference-control interval, a 13M window and saved-input boundaries}
\label{si:conditional-saved-controls}
Keep the invoice family, native tape and controls above, and allow
$\cref\in[1470000,1471000]$. This interval contains both the independently
constructed statistic and the saved scalar. Set
$\varepsilon=10^{-6}$ and $\zeta=10^{-3}$. Uniformly throughout the interval,
the root margins hold and every strict residual in
\eqref{eq:conditional-error-margins} exceeds $2(E+\zeta)$.
Thus the causal quote allowance leaves an additional algebraic index
margin $\zeta$ in the real-arithmetic construction.

The strict budget window can also be written
\begin{align}
 \max\{n(I+b+A),n(I+e+A)\}&<B,\notag\\
 B+\tau&<\min\{n(I+b+A+X),n(I+e+2A+b)\}.
 \label{eq:conditional-control-window}
\end{align}
Both the original $12984796.0484025$ cutoff and $B=13\mathrm{M}$ with
$\tau=10^{-7}$ satisfy it. For every cap of at least eleven, the same
native choices generate all four histories, with every proper pre-crossing
prefix below $B$ and each crossing completed and charged atomically.
Retained counts remain $10/9$, the bills in the preceding table are unchanged,
and the training-net difference is
$\pm0.1-\lambda(2831.6322489)/\cref$, with the same opposite signs
throughout the reference interval.

The saved 32M target lies beyond this early window. The eleven- and
ten-evaluation histories are affordable generated prefixes with a native
continuation. For the specified completed invoice extension and initial
prices, Corollary~\ref{si:conditional-32m-result} below determines the
complete continuation and final retained net value. A cap-eleven complete
instance stops below the target; the full continuation uses a common cap
of at least 45.

Saved initial priors also need their own correspondence. The eight seeds
leave the width-$k4$ own price pools empty. With saved prior
$4201.545544703999$ and cap $86133.78654208$, its unclipped quote differs
from the constructed true fee $1345.84041472$ by $2855.705129983999$.
These saved inputs exceed this family's uniform error envelope. The
calibration parser's default cardinality is four; the constructed panel
has 150 queries. Mapping a saved run to the construction requires its
invocation cardinality and initial price table as well as its reference
scalar.

The added declarations are:
\begin{itemize}
\item \texttt{uniform\_\allowbreak nonzero\_\allowbreak error\_\allowbreak inputs}\\
\texttt{SavedControlMargins.lean}.
\item \texttt{control\_\allowbreak window\_\allowbreak full\_\allowbreak histories\_\allowbreak generated}\\
\texttt{WindowFiniteBudget.lean}.
\item \texttt{control\_\allowbreak window\_\allowbreak positive\_\allowbreak and\_\allowbreak adverse\_\allowbreak net}\\
\texttt{WindowFiniteBudget.lean}.
\item \texttt{saved32M\_\allowbreak favorable\_\allowbreak reported\_\allowbreak quality\_\allowbreak saturated}\\
\texttt{SavedQualitySaturation.lean}.
\item \texttt{saved\_\allowbreak width\_\allowbreak four\_\allowbreak prior\_\allowbreak not\_\allowbreak uniform\_\allowbreak admissible}\\
\texttt{SavedInitialPriorBoundary.lean}.
\end{itemize}
Independent checking covers nine
new modules, 52 theorems/68 declaration targets and 20 diagnostic
theorems/20 targets, using only the standard logical axioms.

\subsection{Complete continuation at 32M}
\label{si:conditional-32m-continuation}
\begin{corollary}[32M continuation for the specified accounting inputs]
\label{si:conditional-32m-result}
Keep the integer tariffs, reward tables, common MC16 completion tape and
150-query panel above. Complete unused invoice rows with one token in each
permitted model role. Use the declared perturbed initial prices with
$\varepsilon=10^{-6}$, inactive caps, $\beta=2$, $c=1.4$,
$\cref=1470690.0198555747$, $B=32\mathrm M$, a common cap of at least
45 and reporting tolerance $\tau\in[0,1]$. The native allocator generates
the following four complete histories and stops on the first evaluation
whose actual cumulative charge reaches $B$.
\begin{center}
\small
\begin{tabular}{llrrr r}
\toprule
Task & Mode & $L_m$ & $T_B^m$ & $C_m(T_B^m)$ & $C_m(L_m)$\\
\midrule
Favorable & AgentUCT & 44 & 43 & 31346070.3166905 & 32171869.5836601\\
Favorable & NextC & 37 & 36 & 31183651.0658382 & 32331275.62775835\\
Adverse & AgentUCT & 45 & 44 & 31850129.6095161 & 32675928.8764857\\
Adverse & NextC & 36 & 35 & 31183651.0658382 & 32331275.62775835\\
\bottomrule
\end{tabular}
\end{center}
Every proper prefix has actual charge below $B$; the whole crossing bill
exceeds $B+1$. Reporting therefore excludes the crossing for every declared
tolerance. Retained best true qualities are $0.5/0.5$ on the favorable
task and $0.4/0.5$ on the adverse task, in AgentUCT/NextC order. Writing
$\Delta\mathcal J^{\rm tr}=\mathcal J_{\rm AgentUCT}^{\rm tr}
-\mathcal J_{\rm NextC}^{\rm tr}$ gives
\begin{align*}
 \Delta\mathcal J_{\rm favorable}^{\rm tr}(32\mathrm M)
   &=-\lambda\,(162419.2508523)/\cref,\\
 \Delta\mathcal J_{\rm adverse}^{\rm tr}(32\mathrm M)
   &=-0.1-\lambda\,(666478.5436779)/\cref.
\end{align*}
NextC has strictly higher training net value for $\lambda>0$ in the
favorable task and for $\lambda\ge0$ in the adverse task.
\end{corollary}
\paragraph{Proof.}
Each step follows from the current raw ancestor backup, the policy's own
paid-prefix cache, per-query quote projection and source-order maximum.
Certified rational bounds on real logarithms and square roots establish
the strict comparisons; equal quotes use the same pre-decision state.
The initial checkpoint and these step equations determine the full native
history. Monotone actual charges establish the first crossing and final
affordable prefix. Exact retained invoice sums and reward maxima give
the displayed net identities.

The positive 13M window and the 32M continuation describe the same causal
mechanism at different cutoffs. At 32M, quality saturation removes the
favorable early gain while its additional search charge persists.

The formal linkage is in \texttt{NativeConcreteEndpoints.lean} and
\texttt{NativeConcreteReporting.lean}. The native event equations determine
all four runs; the quote, invoice and retained-net modules certify the
corresponding prices and endpoints. The proof uses real arithmetic and the
declared common legal completion tape.

\subsection{Failure conditions and correspondence to the experiments}
Zero-price support is needed to stabilize NextC's first positive quote;
assigning a small positive prior to a zero first step can change that quote
discontinuously. Active clipping can keep predictions small while actual
fees are large. A cap of eight can stop at initialization before the target,
and a larger tolerance can include the crossing. Dynamic changes on covered
actions break price agreement; incomplete receipts break the price-pool and
paid-sum correspondence. Positive recurring replay overhead changes the
free-replay accounting. Each change affects one of the quote, receipt or reporting identities used
in the proof.

The construction uses the stated role-valid integer tariffs and mathematical
priors. It illustrates the effect of an allocation sequence under those
inputs. Applying it to a service trace requires the service's own priors,
completion law, arithmetic and receipts. Positive recurring proposer,
repair or replay work also changes the bill.

Equation~\eqref{eq:conditional-training-net} evaluates best observed training
quality minus retained search charges. The benchmark protocol selects an
endpoint through validation, measures test quality, and charges all declared
optimization phases in endpoint $J$. These quantities have separate reward
and cost definitions, as Table~\ref{tab:objective-map} records in the main
paper.

\paragraph{Proof and formal evidence.}
The proof connects input margins to ordered native choices, complete causal
receipts to all-candidate observations, and those choices to generated
histories and actual retrospective reporting. The independently checked
observability batch contains nine modules including the saved-target
boundary, with 50 theorems/72 audit targets and 32 new diagnostic
theorems/34 targets; only the standard logical axioms occur. Its key
declarations are:
\begin{itemize}
\item \texttt{observed\_\allowbreak both\_\allowbreak full\_\allowbreak histories\_\allowbreak generated}.
\item \texttt{completed\_\allowbreak whole\_\allowbreak history\_\allowbreak prefix\_\allowbreak observations}.
\item \texttt{beta2\_\allowbreak all\_\allowbreak unused\_\allowbreak completions\_\allowbreak full\_\allowbreak trajectories}.
\item \texttt{companion\_\allowbreak whole\_\allowbreak bills\_\allowbreak below\_\allowbreak saved\_\allowbreak target}.
\end{itemize}
The predecessor conditional trajectories, selected-footprint diagnostics
and whole-tree results retain their respective scopes.

\section{Cache Information: Updates and Reachable Separations}
\label{app:graph-information}
The cache graph changes the bill of an evaluation through the intersection of its demand with saved resources. Whether the full incidence relation must be retained depends on the demands that future evaluations can make. A full-panel protocol and a query-selective protocol have different sufficient statistics.

Thus full-panel trials admit an exact prefix-level implementation. Query-dependent weights and nonrectangular warm caches also admit this quotient. To determine which information selective evaluation requires, we characterize a larger class of exact compressions.

The theorem identifies the coarsest partition-based residual-sum summary that
works for every unit-price cache state. A single full panel has one group;
the three cyclic pairs $\{q_1,q_2\},\{q_2,q_3\},\{q_3,q_1\}$ have three.
Lossless bitsets, demand-profile tables, and incidence graphs implement the
same information. The partition converse quantifies over arbitrary valid warm caches on
one-step prefixes. The reachable example below establishes information loss
under the overlapping-panel protocol itself. More general encodings than group
sums can preserve the same information.
Physical restoration still requires the saved execution objects; the quotient
compresses cost accounting, not their contents.
The residual-sum formula also applies to the predictor's prices at a fixed
decision. A state-conditioned predictor refreshes those prices from saved
execution observations. Its feature state and the residual accounting table
have distinct roles; the fixed-weight publication update above concerns the latter.
For live prices $\widehat w_t$, write
$r_t(u)=\sum_q\ind\{(q,u)\notin K_t\}\widehat w_t(q,u)$.
After publishing $\mathcal Q\times P_x$ and revising prices, the exact update is
\begin{equation}
\begin{split}
 r_{t+1}(u)&=0 \quad (u\in P_x),\\
 r_{t+1}(u)&=r_t(u)+\sum_{q\in\mathcal Q}\ind\{(q,u)\notin K_t\}
       \bigl(\widehat w_{t+1}(q,u)-\widehat w_t(q,u)\bigr)
       \quad (u\notin P_x).
\end{split}
\label{eq:live-residual-update}
\end{equation}
This identity, checked in \texttt{FullPanelLivePrices.lean}, separates new
cache hits from new information about future execution. A saved stop state
can lower an uncached descendant's predicted price. Computing that revision
requires the predictor's observations. For the main fixed panel, the resulting
residual totals suffice to compute completion costs under any supplied workflow law.

\begin{proposition}[Query identity can be decision-relevant]
\label{prop:graph-separation}
There are two prefix-complete caches with the same per-prefix hit counts and the same full-panel residual bills, for which the cost-minimizing action for one fixed active query is different. No deterministic policy using only these identical summaries is optimal in both states. A randomized policy with the same action distribution in both states incurs at least $1/2$ unit of expected excess cold cost in one state. A single estimate of the same action's bill has absolute error at least $1/2$ in one state.
\end{proposition}
Take two queries $q_0,q_1$, two one-step terminal prefixes $u,v$, unit resource weights, and caches
\[
 K_L=\{(q_0,u),(q_1,v)\},\quad
 K_R=\{(q_1,u),(q_0,v)\}.
\]
Each prefix has one hit and one unit of full-panel residual cost in either state. For a demand on $q_0$, the bills of $(u,v)$ are $(0,1)$ under $K_L$ and $(1,0)$ under $K_R$. With identical candidate quality, the preferred action reverses. Adding one unit of recurring work and imposing a remaining budget of one makes only $u$ feasible in the first state and only $v$ feasible in the second. This witness establishes finite-budget information loss at one decision; assessing the current UCT index requires a separate policy analysis.

\begin{proposition}[Separation reachable from an empty cache]
\label{prop:reachable-separation}
Let the allowed panels include $A=\{q_0,q_1\}$ and $B=\{q_1,q_2\}$,
with two one-step execution prefixes $u,v$ and unit prices.
Starting empty, the complete evaluations $(A,u),(B,v)$ produce
$K_L=(A\times\{u\})\cup(B\times\{v\})$; the evaluations
$(A,v),(B,u)$ produce $K_R=(B\times\{u\})\cup(A\times\{v\})$.
Both histories use the same panel order and cost four. Both have two hits and one residual unit at each
prefix over the three-query universe. For active panel $A$, the completion
bills of $(u,v)$ are $(0,1)$ in $K_L$ and $(1,0)$ in $K_R$.
With one unit of common overhead and one unit of remaining budget, the unique
affordable action reverses.
\end{proposition}
The proof enumerates the four publications in each history and their set
differences. All five claims, including reachability and equal warm-up expenditure,
are checked in \texttt{ReachableDemandSeparation.lean}. This witness uses
complete evaluations of overlapping panels and establishes why global prefix
totals cannot serve as the cost state for that selective-demand protocol.
A policy retaining each query's membership, or the equivalent demand-profile
quotient, computes both states' bills exactly.

A bitset or any other lossless encoding of incidence preserves the required information. A stronger prefix summary that preserves the needed conditional bills also avoids the counterexample. For query-selective evaluation, let $S$ denote the next query subset and $X$ the completed workflow. The prediction target is
\begin{equation}
 \begin{split}
 m_t(p,a)=\mathbb E\Bigl[
   &\sum_{q\in S}\sum_{u\preceq X,\,u\ne\varnothing}
       \ind\{(q,u)\notin K_t\}w_{q,u}\\[-2pt]
   &+h_t(S,X)\mid\mathcal H_t,p,a\Bigr].
 \end{split}
 \label{eq:joint-panel-cost}
\end{equation}
The joint law of $S$ and $X$ matters; separate query and path marginals need not determine their resource-inclusion probabilities. This identity follows by summing the conditional inclusion probability of each uncached resource. The full-panel estimator below sets $S=\mathcal Q$.

Executing a completion also changes later candidates' bills by publishing their shared resources. Equation~\eqref{eq:cacheforecast} quantifies this effect. A full next-evaluation estimator uses the current graph. A planner that invests in reusable work also needs the remaining budget, the endpoint objective, and a value for future cache states.

Propositions~\ref{prop:graph-compression}--\ref{prop:graph-separation} identify when query incidence affects cost-aware allocation. The main experiment uses the fixed-full-panel compression case. Section~\ref{sec:finite-budget} connects cost errors to decisions and evaluation coverage. Theorem~\ref{thm:shared} extends the quality bound to overlapping fixed panels with separate reward backups and a shared cache; this is a theoretical extension beyond the main one-tree protocol.

\section{Further Allocation and Budget Comparisons}
\label{app:finite-budget}
\subsection{The allocation tradeoff at a fixed state}
\label{sec:finite-budget}
At a common tree and cache state, let $U_a$ be ordinary UCT's quality index.
Complete prices allow a precise exchange between that index and evaluation cost:
\begin{equation}
 U_a-\gamma b_a\ge U_b-\gamma b_b
 \quad\Longleftrightarrow\quad
 U_b-U_a\le\gamma(b_b-b_a).
 \label{eq:completion-threshold}
\end{equation}

\begin{proposition}[What pricing buys at the current state]
\label{prop:priced-tradeoff}
Let $b$ maximize ordinary UCT's $U$, and let $a$ maximize
$U-\gamma\widehat b$ over the same initialized candidates, with $\gamma>0$.
Then
\begin{equation}
 \widehat b_a\le\widehat b_b,\qquad
 0\le U_b-U_a\le\gamma(\widehat b_b-\widehat b_a).
 \label{eq:priced-tradeoff}
\end{equation}
If $|\widehat b_j-b_j|\le\epsilon_j$ for $j=a,b$, the actual expected saving obeys
\begin{equation}
 b_b-b_a\ge\widehat b_b-\widehat b_a-\epsilon_a-\epsilon_b.
 \label{eq:physical-saving}
\end{equation}
\end{proposition}

\paragraph{Proof.}
Ordinary-UCT maximality gives $U_b\ge U_a$; priced maximality gives
$U_b-U_a\le\gamma(\widehat b_b-\widehat b_a)$. Nonnegative quality sacrifice
implies predicted cost savings. The two cost-error bounds give the physical
saving inequality. A predicted saving exceeding their sum certifies cheaper
expected execution.

Equal first-edge prices leave this decision unresolved when suffix prices
differ. Complete pricing changes the choice once the predicted suffix saving
exceeds the index sacrifice divided by $\gamma$. A first-unpaid rule retains
a signal beyond already saved prefixes and omits the rest of the bill.
The actual-saving certificate in \eqref{eq:physical-saving} uses
cost-error bounds. The implementation reports point estimates;
their error radii remain uncalibrated.

The proposition concerns alternatives at one common state. Relative test
quality and realized policy-level savings require whole-trajectory comparisons
because entire search policies create different caches. Under fixed weights and persistent
publication, their cold-charge difference is the weight of resources exclusive
to their respective histories. Recurring work and endpoint measurement remain
separate costs.

\subsection{Local error certificates and own-history comparisons}
The cost term prices the next completed reward evaluation. Let
$U_a=Q_a+c\sqrt{\log(N+1)/N_a}$ and $\gamma=\beta/\cref$ at a fixed decision state.
Write $b_a$ for the expected completion bill under the declared continuation law.
Equation~\eqref{eq:completion-threshold} gives the exact exchange condition.
With equal optimistic quality and $\gamma>0$, the rule chooses a cheapest
completion. With unequal quality, that condition specifies the saving required to
offset the quality disadvantage. This is a resource-priced allocation rule;
$U_a$ retains its tree-exploration meaning and need not be a calibrated
confidence bound on a stationary arm mean.

Proposition~\ref{prop:priced-tradeoff} compares the priced choice $a$ with
the ordinary-UCT choice $b$ at the same initialized state. If
$|\widehat b_j-b_j|\le\epsilon_j$ for $j=a,b$, pricing bounds the sacrifice
in optimistic quality and a predicted saving greater than
$\epsilon_a+\epsilon_b$ certifies a strict reduction in expected physical cost.
The statement is local to one common tree and cache state. Realized stochastic
bills need not follow their expectations on every call.

For a net-value interpretation, let $v_j$ be the actual expected reward of the
declared completion, and additionally require $|U_j-v_j|\le s_j$.
Writing $M=(U_a-\gamma\widehat b_a)-(U_b-\gamma\widehat b_b)$ gives
\begin{equation}
 (v_a-\gamma b_a)-(v_b-\gamma b_b)
 \ge M-s_a-s_b-\gamma(\epsilon_a+\epsilon_b).
 \label{eq:net-certificate}
\end{equation}
A positive right side certifies a net-value improvement. This certificate requires per-decision error radii beyond a tree bonus or average prediction error. We use
Equation~\eqref{eq:priced-tradeoff} to audit the implemented rule and reserve
the physical and net certificates for settings with valid error bounds.

\begin{proposition}[Forecast error and allocation]
\label{prop:allocation-error}
Let $a$ maximize $U-\gamma\widehat b$, let $a^*$ maximize $U-\gamma b$ at the
same state, and suppose $|\widehat b_j-b_j|\le\epsilon_j$ for these two actions.
For $\gamma\ge0$,
\begin{equation}
 (U_{a^*}-\gamma b_{a^*})-(U_a-\gamma b_a)
 \le\gamma(\epsilon_{a^*}+\epsilon_a).
 \label{eq:allocation-error}
\end{equation}
\end{proposition}
The inequality follows by inserting the two estimated indices and using
their ordering. Summing it bounds cumulative loss against the true-index
choice at each visited state by the corresponding sum of error radii.
The comparator in this statement uses the same observed state. Comparisons
between whole search policies must also account for their different histories
and caches.

Two consequences explain the choice of horizon and cache information. First,
suppose both candidates have the same next-edge bill $e$, with remaining tails
$d_a,d_b$. If $0<U_b-U_a<\gamma(d_b-d_a)$, next-edge chooses $b$, while the full
bill chooses $a$. The absolute truncation error of $e$ is exactly $d_j$.
Second, scale every cold bill in the two-cache witness of
Proposition~\ref{prop:graph-separation} by $H>0$. Any allocation that sees only
the identical prefix totals incurs expected excess cold cost at least $H/2$
in one of the two states. Incidence-aware allocation has zero excess in both.
Thus aggregation and horizon truncation can each erase the cost difference
needed by~\eqref{eq:completion-threshold}.

\begin{proposition}[Coverage under separate cache histories]
\label{prop:budget-coverage}
Consider two search policies with their own realized, nonnegative evaluation
bills. Suppose each of the first $m$ bills of the first policy is at most
$\ell$, and each of the first $k$ bills of the second is at least $h$.
If $m\ell\le B<kh$, every prefix of the first $m$ evaluations fits budget $B$,
whereas the second policy cannot complete its first $k$ evaluations within $B$.
\end{proposition}
Summing the respective bounds proves the claim; nonnegativity bounds each
intermediate prefix cost. Bills include whatever recurring work the declared
budget charges. This conditional statement gives a directly testable benefit:
lower realized completion costs can fund additional reward observations.
The quality of those additional observations is a separate empirical question.
The experiments report completed evaluations, cost and final quality together;
estimator speed remains untested.

\paragraph{Different trajectories.}
Let $A$ and $B$ be the unions of resources materialized by two policies from
the same initial cache $K$. With fixed prices and successful persistent reuse,
their cold-cost difference is exactly
\begin{equation}
 c_w(K,A)-c_w(K,B)
 =c_w(K\cup B,A)-c_w(K\cup A,B).
 \label{eq:exclusive-cost}
\end{equation}
Common materializations cancel; the policies pay differently for resources
exclusive to their own trajectories. Recurring costs add their own difference.
Whole-run costs depend on the resources exclusive to each trajectory,
and endpoint quality requires evaluating the workflows each policy selects.
For example, take current priorities $(U_a,U_b)=(0.6,0.7)$, exact completion
costs $(1,2)$, and $\gamma=0.5$. UCT chooses $b$ and pricing chooses $a$;
both fit a budget of two. If their unknown next rewards are $(0,1)$, pricing
loses both next-reward quality and net value; for $(1,0)$ it wins both. These
two one-decision cases use uncalibrated $U$ values. Their checked
inequalities isolate a local mechanism. The generated-history result in
Section~\ref{si:conditional-budget-histories} supplies a separate conditional
comparison with its own task inputs and reporting window.

The finite-budget rationale combines conditional reuse information with
the complete reward-producing bill. Policy-level net improvement additionally
requires a quality gain that covers any extra retained charge. Theorem~\ref{thm:main}
preserves long-run quality exploration under persistent cold costs; the
conditional histories compare complete pricing with NextC. Advantage over
UCT at a retained-search-cost ceiling remains an empirical question, with
the full optimization bill reported separately.

\subsection{Affordable coverage and value saturation}
\label{si:affordable-coverage}
Let $W$ be the fixed physical weight of the finite resource universe missing
from the initial cache. Let $H(t)$ be actual recurring work through evaluation
$t$, and let $T_{\rm cache}$ be the coverage threshold in
Equation~\eqref{eq:coverageconstant}. These quantities give a sufficient
certificate for the reported training endpoint.

\begin{proposition}[Completed affordable coverage]
Under Theorem~\ref{thm:main}, consider a completed history containing a prefix
of length $t\ge T_{\rm cache}$. Suppose physical resource weights and recurring
charges are nonnegative, successful publication persists, and
$W+H(t)\le B$. The largest complete affordable prefix contains a workflow
with training reward $V^*$. Its best observed training reward therefore
equals $V^*$.
\end{proposition}
\begin{proof}
Resource conservation bounds the prefix's cold charge by $W$, so its full
charge is at most $W+H(t)\le B$. Retrospective reporting retains at least this
prefix. Coverage has evaluated every terminal, including a maximizer of the
fixed reward table. All retained rewards are at most $V^*$.
\end{proof}
The certificate requires the completed coverage-length prefix as well as
the resource bound. A charging-target or evaluation-cap stop can occur before
that prefix exists. Its potentially large threshold also limits the
certificate's practical budget window.

\paragraph{Saturation and recurring work.}
For retained histories $A,B$ with the same best training reward $V^*$ and
net scale $\alpha>0$, their training-net difference is
$-\alpha[C_A-C_B]$. Additional charged work lowers net value after quality
saturates. More generally, a favorable quality difference $g>0$ is erased
whenever the additional retained search charge exceeds $g/\alpha$. This separates
the task reward gap from its resource and recurring-work requirements.

\paragraph{Forecast margins.}
For a fixed decision, write the true priced index as $U_a-\gamma m_a$ and
its predicted version as $U_a-\gamma\widehat m_a$. If every candidate cost
error is at most $\epsilon$ and the true winner's index margin exceeds
$2\gamma\epsilon$, the predicted index has the same winner. Each pairwise
index difference changes by at most $2\gamma\epsilon$. This local certificate
requires costs for the specified continuation law and fixed quality indices;
subsequent choices generate their own cache and return histories.

\section{Shared Caches, Actual Costs and Recommendation}\label{app:shared-cache}
\subsection{Shared Cache across Fixed Panels}
\label{sec:shared-theorem}
Let a fixed finite family of panels maintain separate trees and deterministic
reward tables. Each tree backs up complete returns on its own panel, while
all trees publish successful qid--prefix executions into one persistent cache.
External publications can reduce a panel's cold forecast. They cannot remove
its own previously saved resources.

\begin{theorem}[Shared-cache quality regret]
\label{thm:shared}
Under the hypotheses of Theorem~\ref{thm:main} for each panel, adding arbitrary
external cache publications preserves its instance-dependent bound
$R_j(T_j)\le A_j\log(T_j+1)+B_j$.
For a fixed finite panel family scheduled for $T$ evaluations in total,
\begin{equation}
 \sum_j R_j(T_j)\le
 \left(\sum_j A_j\right)\log(T+1)+\sum_j B_j.
 \label{eq:shared-log}
\end{equation}
\end{theorem}
The penalty remains nonnegative and bounded by the panel's prediction envelope $H_j=\gamma_j\sum_r M_{j,r}$. The original exploration argument eventually covers every local
terminal, after which the panel's own publications already zero all its cold
penalties. Adding shared entries preserves that fact. The same subtree
induction proves the local bound; $T_j\le T$ gives~\eqref{eq:shared-log}.
The same rate holds for a nonnegative cold-cost approximation dominated by
the complete forecast under its own continuation law. Such a penalty has the
same finite bound and vanishes after coverage. The first unpaid estimator
has this property; its quality-regret bound follows from coverage and the
tree induction, without requiring equality with UCT.
The checkpoint-conditioned model~\eqref{eq:state-price} also satisfies this
domination condition. At each history, finite-sum monotonicity bounds its cold
bill by the fixed envelope $M$; averaging under that history's continuation
law preserves the inequality. The prediction and law can both change while
this envelope remains fixed. The three envelope lemmas in
\texttt{DynamicCompletion} discharge the domination step.
For~\eqref{eq:online-mean}, the calibration pseudo-observation and clipped
physical observations lie in a fixed envelope, so their mean does too.
The sample indices induce a nonnegative terminal law of total mass one;
repeated terminals receive their accumulated mass. Lean's
\texttt{online\_mean\_shared\_root\_log\_bound} in \texttt{OnlineMean}
connects these online prices to the shared-cache quality theorem, and its
sampled-law lemmas identify~\eqref{eq:sample-cost} with a reference expectation.
Next-unpaid is dominated by that same complete forecast. The argument uses
coverage followed by exactly zero cold penalties; boundedness alone does not
imply logarithmic quality regret. The bound is in completed evaluations $T$,
and asserts neither logarithmic regret in expenditure nor a finite-budget F1 advantage.
The number and definitions of panels are fixed in this result. Merging rewards
from changing query subsets into one node mean requires a different analysis.

\subsection{Actual Search Costs and Recommendation}
We first consider one fixed reward panel with no external publications.
Let $W=\sum_{r\in\mathcal D(\mathcal X)\setminus K_0}w_r$, where $\mathcal D(\mathcal X)$ is the union of all terminal demands. The execution recurrence gives the pathwise conservation identity
\begin{equation}
 C_T^{\mathrm{cold}}=\sum_{r\in K_T\setminus K_0}w_r\le W.
 \label{eq:conservation}
\end{equation}
This bound holds for adaptive choices. With several publishing panels, the
equality applies to the family-wide cache and the sum of its physical charges.
A local panel's bill alone does not equal growth of that shared cache, because
other panels may have paid for the new entries.
For another length-$T$ policy $\pi$ on the same fixed reward table, executing against its own cache, define the cumulative net comparison
\begin{equation}
 R_{\mathrm{net}}^\pi(T)=\sum_{t=1}^T[f(x_t^\pi)-f(x_t)]
             +\eta(C_T-C_T^\pi),\quad\eta\ge0.
 \label{eq:netregret}
\end{equation}
Nonnegative comparator costs imply
\begin{equation}
 R_{\mathrm{net}}^\pi(T)\le R_Q(T)+\eta\left(W+\sum_{t=1}^Th_t(x_t)\right).
 \label{eq:netbound}
\end{equation}
If the two trajectories have equal total recurring overhead, that overhead cancels and the additive cost term is $\eta W$. Thus cold-only own-cache cumulative net regret is also $O(\log T)$. Arbitrary repeated overhead can contribute a linear term.

For $T\ge1$, the best observed training workflow $\widehat x_T$ satisfies $V^*-f(\widehat x_T)\le R_Q(T)/T$ and becomes exactly optimal once all terminals have been covered. The single-panel endpoint sorts observed terminals by that panel's training score, takes a nonempty top-$k$ shortlist, and chooses its validation maximizer. For validation scores $g$, the condition $\sup_{x\in\mathcal X}|g(x)-f(x)|\le\epsilon$ gives
\begin{equation}
 \max_x g(x)-g(x_{\mathrm{end}})\le R_Q(T)/T+2\epsilon.
 \label{eq:validation}
\end{equation}
This is a deterministic table-transfer bound; the experiment does not estimate a valid uniform $\epsilon$. Without such a condition, a training shortlist can omit the validation optimum even after full coverage.
The main experiment uses $k=5$, ranking distinct completed workflows by full-training F1 and breaking ties by first evaluation. Validation ranks only these five candidates and never changes the search statistics. Historical multi-panel nominations are excluded from this protocol.

Finally, endpoint utility $J=q_{\mathrm{rep}}(x_{\mathrm{end}})-\lambda_J C_{\mathrm{opt}}/\cref$ uses one declared reporting quality and the actual optimization bill. The formal reporting quality is held-out F1, excluding the tiny development sentinel; validation $g$ selects the workflow. The report presents F1 at common search budgets together with $J$ from the full physical optimization bill, including crossing and validation work. Retained-snapshot, validation, preparation and test bills are also listed separately. Its utility gap contains the cost difference even when reporting qualities are equal. $C_{\mathrm{opt}}$ includes the declared optimization phases, whereas~\eqref{eq:conservation} concerns search execution. Neither~\eqref{eq:mainbound} nor~\eqref{eq:validation} proves endpoint-$J$ optimality. The selection weight $\beta$ and reporting weight $\lambda_J$ have separate roles.

\section{Generated Workflows: Cost before and after an Edit}
\label{sec:dynamic-completion}
AFlow generates a child program after choosing a parent. Its edit tree records
program ancestry. Reuse requires a matching actual invocation prefix: the
query, ordered call inputs, frozen model and decoding settings. A shared parent
establishes ancestry; cache reuse is determined by invocation compatibility.

Let $G$ denote a proposal outcome containing the child program and its complete
proposal/repair transcript, or a final rejection; let $Z$ be its subsequent
execution trace and $\mathcal H_t$ the observed history. The pre-proposal cost target is
\begin{equation}
\begin{split}
 m_t(a)=\mathbb E_{G\mid\mathcal H_t,a}\bigl[&c_{\rm proposal}(G)
 +\mathbb E_{Z\mid G,\mathcal H_t,a}\bigl[
 c_w(K_t,D(G,Z))+h(G,Z)\bigr]\bigr].
\end{split}
\label{eq:generated-cost}
\end{equation}
The transcript makes the outer cost include all charged proposal and repair attempts,
including different costs for identical child programs reached through different retries. Rejected
proposals form failure outcomes with empty execution demand. The inner law is
conditional on the generated program; it permits dependent call counts,
inputs, cache misses and prices. After generation, conditioning on the
observed $G$ removes proposal uncertainty. Concrete invocation keys determine
hits as execution reveals them. Known source code alone need not determine a
future call's prompt, which may depend on earlier model outputs.

Equation~\eqref{eq:generated-cost} is the full editing-and-evaluation
target for the current counted-v3 study. Its implementation is the uncapped
prequential point predictor in Appendix~\ref{sec:gsm-current-study}:
parent source and prompt lengths predict the first proposal, observed
proposal/repair ratios scale that estimate, and parent execution bills predict
the complete training-panel evaluation. Updates use completed training
edits, including scored failures; an abandoned edit remains billed without
supplying a fitted observation. Predictor fitting uses training receipts only.
A calibrated conditional law and confidence bounds remain unavailable for
this point predictor. Unlike the finite RAG cold estimator,
its proposal/repair component can recur after execution resources are saved.
Historical pilot and finite-catalogue settings are separated from the current
cohort in Appendix~\ref{sec:gsm-history}.

\begin{proposition}[Accounting with unknown future resources]
\label{prop:dynamic-ledger}
For any finite realized sequence of finite demands $D_1,\ldots,D_T$ and
persistent updates $K_{t+1}=K_t\cup D_t$,
\begin{equation}
 \sum_{t=1}^T c_w(K_t,D_t)
 =c_w\!\left(K_1,\bigcup_{t=1}^T D_t\right).
 \label{eq:dynamic-ledger}
\end{equation}
Recurring work contributes the separate sum $\sum_t h_t$. The resource
universe and future programs need not be enumerated in advance.
\end{proposition}
Induction applies Equation~\eqref{eq:sequentialcache}. For finite conditional
scenario laws, distributing the outer and inner sums proves
Equation~\eqref{eq:generated-cost}; no independence assumption is needed.
\texttt{DynamicCompletion.lean} checks both identities.

\begin{proposition}[Conditional guarantee for a dynamic net-value rule]
\label{prop:dynamic-optimism}
At each actual history let $v_t(a),m_t(a)$ be the conditional reward and full cost
of one bounded edit/evaluation attempt, assigning reward zero to an exhausted
attempt with no evaluation.
Suppose $U_t(a)\ge v_t(a)$ and $L_t(a)\le m_t(a)$ for every available action.
Choose $a_t\in\operatorname*{arg\,max}_a\{U_t(a)-\gamma L_t(a)\}$, $\gamma\ge0$, and let $a_t^*$
maximize $v_t(a)-\gamma m_t(a)$ at that same history. If the chosen action's
widths satisfy $U_t(a_t)-v_t(a_t)\le s_t$ and
$m_t(a_t)-L_t(a_t)\le e_t$, then
\begin{equation}
 \sum_{t=1}^T \bigl[v_t(a_t^*)-\gamma m_t(a_t^*)
                 -v_t(a_t)+\gamma m_t(a_t)\bigr]
 \le \sum_{t=1}^T s_t+\gamma\sum_{t=1}^T e_t.
 \label{eq:dynamic-width}
\end{equation}
\end{proposition}
Optimism bounds the comparator by its index; selection bounds that index by
the chosen index. Subtracting the chosen true value leaves its two widths.
The one-step inequality and its sum are Lean-checked. Logarithmic loss follows
when the combined width sum is $O(\log T)$. This is a conditional same-history
oracle comparison. A whole-policy comparator owns a different cache history.
For the experimental raw-backup bonus and point cost model, certified
intervals and a logarithmic width-sum bound remain unestablished.

The finite-tree theorem uses a different route: finitely many cold resources
are eventually all materialized. An unbounded edit process can demand a fresh
unit-cost suffix at every step. Lean verifies that its total cold cost is
exactly $T$, despite persistence. The open-ended setting therefore has the
accounting identity and conditional width guarantee above. The RAG study evaluates finite workflows; counted-v3 GSM8K evaluates
generated programs. The latter uses the accounting identity but does not
satisfy the fixed-table, bounded-cold-prediction assumptions of the main
quality theorem. Its point predictor also supplies no certified intervals
or logarithmic width sum for this conditional dynamic guarantee.

\paragraph{What WTB restores.}
RAG serializes the query, evidence pool, ordered answer evidence, progress and
token state at each action prefix; execution forks from the deepest compatible
checkpoint. The AFlow implementation starts the child Python program in a fresh
workspace and restores WTB results for matching deterministic invocation
prefixes. Editing a call invalidates that call and its dependent suffix;
unchanged earlier calls remain reusable. Restoration covers matching WTB
invocation results; Python stack frames and arbitrary program files remain
outside that contract.

\paragraph{Stopped RAG loops.}
Once a saved evidence assessment halts retrieval, remaining requested
retrieval slots advance the finite grammar without issuing model calls.
They have zero physical resource weight. Lean proves that adding such
zero-weight demands preserves the bill. This connects the runtime guard to
the fixed terminal table under its stated execution assumptions: multiple
syntactic workflows may execute the same productive calls. Zero-work slots
alone do not establish fixed returns across independent fresh executions.

\fi

\section{Other Objectives and Estimator Proofs}
\label{app:other}
\subsection{Independent Cache Trajectories and Endpoints}
Each comparator reward is at most $V^*$, so its cumulative reward advantage is at most $R_Q(T)$. The actual cold bill is at most $W$ and the comparator cold bill is nonnegative, even though they use different caches. This proves~\eqref{eq:netbound}. Equal recurring-overhead totals cancel before applying the cold-budget bound.

The best observed training reward is at least the average observed reward. Subtracting from $V^*$ gives the simple-gap bound. After terminal coverage, an observed training maximizer is a global table maximizer. For~\eqref{eq:validation}, let $x_g$ maximize validation and let $x_f$ be the training-ranked head. The head belongs to every nonempty top-$k$ shortlist, so $g(x_{\mathrm{end}})\ge g(x_f)$. Therefore
\begin{align*}
 g(x_g)-g(x_{\mathrm{end}})
 &\le f(x_g)+\epsilon-f(x_f)+\epsilon\\
 &\le R_Q(T)/T+2\epsilon.
\end{align*}
The uniform table difference is a hypothesis. For a counterexample without it, two fully observed workflows with training scores $(1,0)$ and validation scores $(0,1)$ give validation regret one after training top-1 selection.

For endpoint utility, direct subtraction gives
\[
 J^\pi-J=\bigl(q_{\mathrm{rep}}(x_{\mathrm{end}}^\pi)-q_{\mathrm{rep}}(x_{\mathrm{end}})\bigr)
       +\frac{\lambda_J}{\cref}(C_{\mathrm{opt}}-C_{\mathrm{opt}}^\pi).
\]
Equal quality and a strictly larger actual bill give a strictly positive gap when $\lambda_J,\cref>0$. Cumulative regret and endpoint utility thus require distinct interpretations.

\subsection{Cost Targets and Error Bounds}
For the sampled resource formula~\eqref{eq:resourceform}, exchange the finite sample and resource sums in~\eqref{eq:sample-cost}. The coefficient of each forecast weight is the empirical frequency that the sampled paths demand this resource while it is absent from the current graph. Exchanging the finite sums in~\eqref{eq:reference} instead gives the exact reference expectation, with reference inclusion probabilities as coefficients. Shared-resource dependence never enters either operation.

To isolate the effect of publication, partition $\mathcal D(x)\setminus K$ into the disjoint sets $\mathcal D(x)\setminus(K\cup A)$ and $(\mathcal D(x)\cap A)\setminus K$. Subtracting the first bill leaves exactly the second. Averaging with the same law and weights proves~\eqref{eq:cacheforecast}. This argument formalizes sequential cache accumulation and its effect on prediction without treating a prediction as physical execution; see \leanname{coldWork_sequential_union}, \leanname{referenceColdCost_publication_savings}, and \leanname{referenceColdCost_cache_antitone}.

For~\eqref{eq:mse}, expand the square around the full conditional mean $m_t$. The cross term is zero because $\E[b_t(X)-m_t]=0$. The remaining excess is $(z-m_t)^2$, which vanishes exactly at $z=m_t$. The formal finite-sum theorem needs normalization; nonnegativity makes the normalized weights a probability law.

To prove Proposition~\ref{prop:error}, insert $\sum_x\rho_C(x)b_t(x)$ between the forecast mean and the actual mean. The first difference is bounded by the reference-weighted bill errors. Apply the triangle inequality within each cold resource sum and to the overhead. Bound the second difference by $B_{\max}\sum_x|\rho_C(x)-\rho_{\mathrm{exec}}(x)|$. The estimate explicitly includes both sources of error.

Let $|\widehat b-b|\le e$ and $b\ge0$. Then $0\le\underline b=\max\{0,\widehat b-e\}\le b$ and $b-\underline b\le2e$. These facts prove~\eqref{eq:optimism}. If $a$ maximizes that optimistic index and $a^*$ maximizes true net value, then
\begin{align*}
 \mu_{a^*}-\gamma b_{a^*}
 &\le Q_{a^*}+r_{a^*}-\gamma\underline b_{a^*}\\
 &\le Q_a+r_a-\gamma\underline b_a\\
 &\le\mu_a+2r_a-\gamma b_a+2\gamma e_a.
\end{align*}
The last step uses the selected arm's reward upper-error bound. This proves the stated conditional net-gap inequality without asserting that tree means satisfy those reward confidence conditions.

\subsection{Reduction and Recurring-Cost Counterexample}
A positive child count witnesses an earlier complete event whose expanded path passes through that child. Its terminal completion extends the child and has materialized the child's entire panel demand. Persistent publication retains that demand, so its current cold bill is zero for every nonnegative predictor. The same holds for its parent. Equation~\eqref{eq:next} is zero, and the actual initialized-child index comparisons become exactly ordinary UCT comparisons. Unvisited-first clauses are identical as well. This proves equivalence of the whole allocation predicates on a given history.

For trajectory equality, let $\mathcal E_{\mathrm{next}}(h)$ and $\mathcal E_{\mathrm{UCT}}(h)$ be the complete events that can extend history $h$ while satisfying the same fixed reward table, legal expansion protocol, and respective allocation rule. The predicate equivalence gives $\mathcal E_{\mathrm{next}}(h)=\mathcal E_{\mathrm{UCT}}(h)$. Use the same initial tree and cache and a shared resolver $\Phi(h,\mathcal E)$ that returns an admissible event or stops. The resolver fixes expansion, rollout completion, tie-breaking, evaluation outcomes including receipts, and stopping; a common random tape can be fixed as part of it. Equal histories give equal event sets and the same output of $\Phi$. Induction therefore gives equal event histories at every step, hence equal reward backups, expanded trees, and cache graphs. The Lean development proves both resolver validity and this coupled equality. Rollout coupling is necessary: expanding the same node can lead to different leaves and rewards when its completion choices differ.

For historical continuation costs, each terminal extending $p$ demands a superset of $p$'s resources. Subtracting the same cached set preserves inclusion, and nonnegative weights give $\widehat d_t(x)\ge\widehat d_t(p)$. Remove the inactive clipping and use $\sum_x\rho_C(x)=1$ to pull the parent term outside the expectation. This proves the common shift. With a common omitted tail the same algebra applies; action-dependent tails introduce the exact pairwise difference given in the main text.

For~\eqref{eq:nextnet}, apply the ordinary-UCT quality bound to the next-edge allocation predicate. The comparator's reward at every step is at most $V^*$. Each trajectory is charged against its own cache, and cancellation of equal recurring-overhead totals leaves a cost difference at most $W$. Multiplying by $\eta\ge0$ proves the bound. The complete composed statement is \leanname{next_edge_own_cache_log_bound}; it quantifies the gap and constants before all actual and comparator histories. For the local approximation relation, full net gap equals next-edge net gap plus $\eta$ times chosen-tail minus comparator-tail at each step. Sum these identities and use the assumed tail-variation bounds, as in \leanname{next_edge_tail_regret_transfer}. A logarithmic local proxy bound remains a separate premise.

For the recurring-cost example, initialize arm A then arm B. Their immediate-cost net values are $0.4$ and $0.6$, while their full-cost net values are $0.4$ and $0.3$. The next-edge UCB rule therefore treats A as the inferior proxy arm with gap $0.2$. At any subsequent selection of A, exact deterministic reward means and index maximality imply $0.2\le\sqrt{2\log(t+1)/N_A(t)}$. Counting such selections gives a logarithmic upper bound on $N_A(T)$. In particular the conservative bound $N_A(T)\le400\log T+4$ holds for $T\ge1$; for $T\ge2$ the direct count bound is already stronger. Since full-cost regret is $0.1(T-N_A(T))$, the claimed lower bound follows. The preserved Lean proof obtains the same conservative constants via its general local-regret counting lemma.

\subsection{Cache Compression and Information Loss}
For Proposition~\ref{prop:graph-compression}, exchange the finite sums over qids and demanded prefixes. This gives the residual bill in~\eqref{eq:residual-prefix}. Full-panel publication adds every qid incidence for each demanded prefix, so exactly those residual bills become zero. The product identity $(Q\times S)\cup(Q\times P)=Q\times(S\cup P)$ proves rectangularity by induction over the publication history. Lean checks these statements in \leanname{full_panel_residual_quotient}, \leanname{full_panel_residual_update}, and \leanname{full_panel_history_rectangular}.

For Proposition~\ref{prop:graph-separation}, enumerate the two resource sets in the stated witness. Both hit-count vectors and both full-panel residual-bill vectors coincide. The partial-panel bills are $(0,1)$ and $(1,0)$. A deterministic function of the identical summary chooses the same action in both states, and therefore loses one unit in one of them. If it chooses $u$ with probability $p$, expected excess costs are $1-p$ and $p$, whose maximum is at least $1/2$. A common estimate $z$ of $u$'s bill has errors $|z|$ and $|z-1|$, again with maximum at least $1/2$. Finally, the unit recurring bill makes total costs $(1,2)$ and $(2,1)$, proving the budget claim. The 12 checked declarations and assumption review are recorded in \path{theory/evidence/graph-information/REVIEW.txt}. These are information and one-decision results; no multi-round policy-regret lower bound is inferred.

\subsection{Completion Decisions and Shared Panels}
For Theorem~\ref{thm:demand-quotient}, each allowed panel is a union of profile
classes. Exchange the finite sums over queries, groups, and prefixes to obtain
Equation~\eqref{eq:group-residual}. Publication includes either every member of
a class at a prefix or none, so the displayed zeroing update is closed on the
summary. For necessity within partition-based residual sums, let a merged
class contain $q\in\mathcal Q_j$ and $r\notin\mathcal Q_j$. With unit prices,
compare a residual vector missing only $q$ to one missing only $r$. Every
group total is identical; the panel bills are one and zero. No decoder of
those totals recovers both. One-step prefixes make both caches prefix-complete.
The Lean module \texttt{TreeDemandQuotient} proves the finite-sum identity,
publication rule, profile criterion, and decoder impossibility separately.
The necessity statement characterizes partition-based residual sums. Lossless
encodings preserve the same distinctions; restricting reachable states can
permit additional compression.

For Proposition~\ref{prop:priced-tradeoff}, unpriced maximality gives
$U_b-U_a\ge0$ and priced maximality gives
$U_b-U_a\le\gamma(\widehat b_b-\widehat b_a)$.
Positive $\gamma$ implies $\widehat b_a\le\widehat b_b$.
The triangle inequality gives
$b_b-b_a\ge\widehat b_b-\widehat b_a-\epsilon_a-\epsilon_b$.
Applying the reward-error bounds as well proves~\eqref{eq:net-certificate}.
For~\eqref{eq:exclusive-cost}, apply sequential cache accounting to $A\cup B$
in both orders and subtract the two expressions. All these arguments are
checked in \texttt{TreeDecisionValue}. Its numerical example checks only the
two displayed one-decision cases, with no hidden claim of a legal full-tree
counterexample or of calibrated UCT confidence bounds.

Equation~\eqref{eq:completion-threshold} is a rearrangement of the two indices.
For Proposition~\ref{prop:allocation-error}, insert $\widehat b_a$ and
$\widehat b_{a^*}$. Estimated-index maximality removes the middle difference;
the remaining terms are bounded by $\gamma\epsilon_a$ and
$\gamma\epsilon_{a^*}$. Lean also checks the finite-sum version in
\leanname{completion_index_cumulative_error}. The next-edge reversal follows
by subtracting the common edge bill. The scaled information loss follows from
$\max\{H(1-p),Hp\}\ge H/2$. Proposition~\ref{prop:budget-coverage} sums each
policy's own costs, with nonnegative prefix costs ensuring every earlier
prefix also fits. These declarations are in \texttt{TreeFiniteBudget}.

For Theorem~\ref{thm:shared}, write a local cache as $K^{\rm own}_t\cup E_t$,
where $E_t$ contains publications made outside the local tree. The cold penalty
is nonnegative and at most $H_j$ because the completion law is normalized and
all resource predictions are nonnegative. Construct the same bounded tree run
as in the single-panel proof. At its finite coverage threshold, local
publications alone contain every required resource. Their inclusion in
$K^{\rm own}_t\cup E_t$ zeros the cold penalty. The original subtree induction
then applies without alteration. The external set may depend on the local
history; no independence assumption is used. Summing the resulting local
bounds and using $T_j\le T$ gives the family bound, with nonnegative $A_j$.
\texttt{TreeSharedCache} checks both statements.

The first unpaid estimator uses an ordered resource list for each query.
\texttt{nextUnpaidBill} skips cached resources and nonpositive predicted
prices; otherwise it returns the current price. Induction gives a nonnegative
value bounded by \texttt{orderedColdBill} for nonnegative predictions. Summing
over queries and averaging under the same nonnegative completion probabilities
preserves this inequality. Full coverage zeros the estimate. The theorem
\leanname{dominated_cold_shared_root_log_bound} proves the whole-tree bound
for any nonnegative penalty bounded above by the complete cold forecast.
Its proof replaces the full-forecast bound by transitivity and uses that
domination to zero the penalty at coverage. The immediate-child reduction
is not used. \leanname{nextUnpaidBill_cached_edge_counterexample} checks the
saved-edge example in the cost-estimator analysis.

\subsection{Reachable Separation and Generated Workflows}
For Proposition~\ref{prop:reachable-separation}, both demand lists contain four
unit-price resources and start from the empty set. Direct set subtraction gives
one residual unit at each prefix over the three-query universe. Restricting
the next demand to $A$ gives $(0,1)$ and $(1,0)$. Adding the common unit cost
proves the budget statement. \texttt{ReachableDemandSeparation} checks
reachability, warm-up bills, residual equality, ranking and affordability.

For Proposition~\ref{prop:dynamic-ledger}, induct on the realized demand list.
The induction step uses $c_w(K,A)+c_w(K\cup A,B)=c_w(K,A\cup B)$.
Adding the recurring-work sum yields the complete accounting identity.
Expanding both sides of Equation~\eqref{eq:generated-cost} and normalizing the
conditional execution law proves the two-stage expression, with arbitrary
dependence on the generated child.

For Proposition~\ref{prop:dynamic-optimism}, optimistic bounds give
$v_t(a_t^*)-\gamma m_t(a_t^*)\le U_t(a_t^*)-\gamma L_t(a_t^*)$.
Maximality bounds this by the chosen index. Subtracting the chosen true net
value leaves at most $s_t+\gamma e_t$; finite summation proves the result.
\texttt{DynamicCompletion} checks the identities, this bound, zero-weight
padding, and the fresh-unit-resource sequence whose total bill is $T$.
Its statements quantify over the realized cache state, allowing future
generated programs to remain unspecified.

For checkpoint-conditioned predictions, finite summation preserves
$0\le\widehat w_{t,r}\le\bar w_r$. Averaging under any nonnegative continuation
law preserves the same domination. Apply the shared-cache theorem with the
fixed envelope $\bar w$: its coverage threshold and constants still apply,
and coverage zeros every dominated live penalty. The declaration
\leanname{checkpoint_conditioned_shared_root_log_bound} checks this complete
composition, including history-dependent prices and continuation laws.

\ifdefined\AgentUCTProofOnly\else
\section{Current GSM8K Generated-Program Study}
\label{sec:gsm-current-study}
\subsection{Frozen counted-v3 protocol}
\label{sec:aflow-system-study}
The formal counted-v3 cohort compares allocators inside one AFlow system: three methods, five paired seeds, and fifteen completed searches. It is
distinct from AFlow--RAG, which optimizes a fixed component grammar.
Its identity is \texttt{study\_protocol=counted-v3}. The inherited
\texttt{formal-v1} data/calibration filenames and \texttt{fast-v2} worker
name identify source components, not additional trials pooled into this
cohort.
All three primary arms---AFlow's parent selector, MCTS-UCT and AgentUCT---
share the initial program, Qwen3.5-4B editor and executor, prompts,
operators, execution feedback, code sandbox, scorer, repair limits, duplicate
handling and held-out selection protocol. MCTS-UCT and AgentUCT share progressive
widening $\min(8,\max(1,\lceil\sqrt{N}\rceil))$, depth limit 12 and no rollout;
their difference is completion pricing. AFlow samples among the four
highest-scoring eligible parents using its original probability mixture,
with the same depth and eight-child caps but without visit-based widening.
Its comparison therefore concerns the allocation policy as a whole. The
shared execution system isolates those policy differences from changes in
the editor or operators. The comparison applies to this bounded program-generation system
with its shared runtime and panels.

The GSM8K panels contain 150 training, 300 validation and 300 primary test
questions, with separate calibration and sentinel questions and exclusions
for the development pilot. Training and validation use the official training
split; primary testing uses the official test split. Five seeds give fifteen
searches. Independent cost calibration evaluated four fixed programs on
twelve separate questions: Custom, Programmer, two-step refinement, and
three-answer generation followed by ScEnsemble. Four predetermined editing
episodes measured proposal and repair cost without executing their generated
programs. The calibration cost was 109,983 raw tokens.

The measured costs set $C_{\rm ref}=150{,}954.5$ raw tokens and reporting
budgets of 6.423M, 12.772M and 19.122M raw tokens. The reference is the median
projected 150-question fixed-program bill plus the median editing-episode
bill. The budget rule adds 8, 16 or 24 times the largest projected program
bill plus the largest editing bill to the projected initial-program cost,
rounding upward to 1,000 tokens. These are cost-based planning equivalents;
actual completed evaluations depend on the programs proposed. No test quality
enters calibration or budget choice. Every evaluation completes before its
actual bill determines the affordable snapshot. The formal-config tiny changes
only loaded data, retaining these algorithm, model, budget and cost settings.

\paragraph{The formal edit-cost forecast.}
\label{sec:gsm-main-forecast}
The replacement GSM8K cohort uses \texttt{prequential-edit-token-v1} to estimate
the bill for proposing an edit of a known parent program and obtaining its
complete training feedback.  Let $H_t$ contain the completed training edits
retained in the current search checkpoint.  For parent $p$, the request-size
proxy is $s_p=|\mathrm{graph}_p|+|\mathrm{prompt}_p|+|\mathrm{operator\ description}|+1800$
characters.  From the first proposer call of edit $h$, let $I_h$ and $O_h$ be its
input and output tokens, and $F_h$ its actual token bill.  Let $P_h$ be the actual
bill of all proposer and repair calls within that edit, $E_h$ its newly charged
training-execution bill, and $D_h$ the saved parent-cold bill used when making
its forecast.  The empirical factors are
\begin{align}
 r_t&=\frac{1}{|H_t|}\sum_{h\in H_t}\frac{I_h}{s_h},&
 o_t&=\frac{1}{|H_t|}\sum_{h\in H_t}O_h,\nonumber\\
 a_t&=\frac{1}{|H_t|}\sum_{h\in H_t}\frac{P_h}{\max(F_h,1)},&
 e_t&=\frac{1}{|H_t|}\sum_{h\in H_t}\frac{E_h}{\max(D_h,1)}.
 \label{eq:gsm-formal-edit-factors}
\end{align}
For an empty history, $(r_t,o_t,a_t,e_t)=(0.25,884,1,1)$.  The $1768$-token
proposal prior is the median input-plus-output bill of the first physical call
in four independent proposal/repair episodes, one from each fixed calibration
parent.  The output initialization $884$ is half that total-token prior, not
a separately measured mean output length.  For each
training question $q$, $d_{p,q}$ sums the actual bills referenced by the parent's
latest saved training-execution row, including charged failed attempts and
format retries.  If that row has no call IDs, it uses
$d_{p,q}=0.25\ell_q+531$, where $\ell_q$ is the input-question plus supplied-context
character count and $531$ is the median actual bill of physical executor calls
across the four predeclared fixed calibration workflows.  Generated calibration
edits were not evaluated, and task scores did not set either prior.  The
question-indexed vector gives
\begin{align}
 \widehat C_t^{\rm first}(p)&=\max\{1,s_p r_t+o_t\},\nonumber\\
 \widehat C_t(p)&=a_t\widehat C_t^{\rm first}(p)
 +e_t\sum_{q\in\mathcal Q_{\rm train}}d_{p,q}.
 \label{eq:gsm-formal-edit-cost}
\end{align}
The factor $e_t$ reflects physical execution relative to the parent-cold bill:
cache reuse can reduce it, while added calls and retries can make it exceed one.
These empirical factors have no imposed upper cap or calibrated confidence
radius.  AgentUCT uses the complete forecast in its normalized cost penalty
with $\beta=2$ and $C_{\rm ref}=150{,}954.5$; MCTS-UCT sets that penalty to zero.
The forecast is computed before generating the new program.  Its observations
are committed only after the full $150$-question training evaluation, with raw
quality backpropagated at that same boundary.  A complete evaluation with
execution failures contributes its actual bills and scored feedback; an edit
that never reaches complete feedback retains its bills without fitting these
factors.  Estimator updates use completed training receipts only.

\paragraph{Initial program, prompts, and history.}
All three policies start from the same \texttt{Workflow} class.
Its constructor stores
\begin{quote}
\begin{verbatim}
operator.Custom(llm=None)
\end{verbatim}
\end{quote}
and its asynchronous call executes
\begin{quote}
\begin{verbatim}
self.custom(
    input=problem,
    instruction=prompt_custom.ANSWER_PROMPT
)
\end{verbatim}
\end{quote}
and returns \texttt{(response["response"], 0)}.  The exact initial instruction is
\begin{quote}\ttfamily
Solve the problem. End with the final answer in \textbackslash boxed\{...\}.
\end{quote}
The executed instruction ends with a newline; its saved Python prompt assignment
escapes that newline and the backslash before \texttt{boxed}.
The shared proposer assembles \texttt{WORKFLOW\_INPUT},
\texttt{WORKFLOW\_CUSTOM\_USE}, and \texttt{WORKFLOW\_OPTIMIZE\_PROMPT} from
\path{ragspace_experiments/vendor/aflow_prompts.py}, together with the parent's
program, prompt, training score, sibling experience, and the executable
operator descriptions.  Feedback supplies at most the first two saved training
predictions, scores, and failure strings; it excludes gold solutions and uses
the question directly up to $1600$ characters, otherwise its first and last
$800$ characters with an omission marker.  Operators \texttt{Custom},
\texttt{ScEnsemble}, and \texttt{Programmer} are shared upstream definitions in
\path{sources/reference/AFlow/scripts/operators.py}; their fixed ensemble and
Python-verifier prompts are in \path{scripts/prompts/prompt.py} within that
upstream source.  The appended XML contract requires complete
\texttt{modification}, \texttt{graph}, and \texttt{prompt} fields.

The \texttt{counted-errors-v1} serializer preserves each evaluated sibling,
including a zero-score failed evaluation, as a separate history row.  Repeated
identical execution-error strings are represented by their counts, with
distinct strings in first-seen order.  Only consecutive identical
\emph{unevaluated} feedback rows are combined, retaining their occurrence
counts.  The saved search history, failed attempts, and their bills remain
complete.  A rejected proposal receives the exact rejection reason and source
in its repair prompt; a proposal episode permits at most three proposer/repair
calls.  The campaign separately caps proposal admissions at $256$ and executed
workflow evaluations at $256$, including initialization and runtime failures.
Identical parent or earlier accepted sibling programs are rejected by the
common contract; operator format failures also retain all attempts under the
shared three-attempt XML and Pydantic-type contract.

\paragraph{Frozen data and generation runtime.}
The five configurations \texttt{formal-s11/s23/s37/s53/s71.json} in
\texttt{dynamic-aflow-gsm8k-counted-v3} all reference the same
\texttt{dynamic-aflow-gsm8k-formal-v1/data.json}; the data-preparation manifest
records original split and row index for every question.  Training and
validation contain $150$ and $300$ disjoint official-train questions; the
separate $12$ calibration questions and one diagnostic sentinel also come from
that split.  The test array places that sentinel first, followed by $300$
official-test questions; \texttt{primary\_test\_start\_offset=1} excludes it from
reported accuracy.  The preparation fixes sampling seed $20261004$ and excludes
$81$ question texts appearing in the enumerated earlier development inputs.
Its ordered source IDs and complete question rows are retained with the frozen
configurations.

All proposal, repair, and execution calls use the pinned Qwen3.5-4B checkpoint
identified by \path{experiments/rag_component_selection/model_assets.py}.
The \texttt{aflow-fixed-group-fast-v2} worker uses fixed groups of eight
questions, BF16, SDPA, left padding, and a chat template with
\texttt{enable\_thinking=False}.  Execution is greedy; proposal and repair are
sampled as singleton requests with temperature $0.7$ and top-$p=0.9$, using the
frozen prompt-derived seed rule.  The context bound is $8192$ tokens, with
output bounds $1024$ for execution and $2048$ for proposal/repair.  Each question
has a $42$-call and $600$-second workflow limit.  Saved generator receipts for
all $15$ units record Python from \texttt{.venv-rag-batch-v2}, PyTorch $2.14.0$,
Transformers $5.17.0$, \texttt{causal-conv1d} $1.7.0$, and
\texttt{flash-linear-attention} $0.5.2$; the worker source and dependency
metadata specify the remaining runtime.  The control runner uses
\texttt{.venv-p0p1-v22}, with the counted-history overlay preceding the shared
XML-contract overlay.  Thus the formal cohort has one shared proposer,
executor, scorer, retry contract, and data binding across the three policies.

The search bill includes proposal, repair, execution and the whole crossing
evaluation. A completed full training panel supplies one reward; no partial
panel updates tree statistics or prices. The results below report complete-panel accuracy, actual phase bills, physical failures, scored candidate errors and saved allocation counts. Useful-edit, duplicate, validity and cache-hit rates are not aggregated here. Historical pilots and retired diagnostic runs are reported separately in Appendix~\ref{sec:gsm-history}.

\subsection{Complete five-seed generated-workflow comparison}
\label{sec:gsm-five-seed-complete}
Completion pricing produced a modest mean quality advantage over MCTS-UCT
in the frozen GSM8K AFlow system (Table~\ref{tab:gsm-counted-all-five-seed}).
At 19.122M raw search tokens, AgentUCT reached $93.07\pm1.48\%$,
compared with $92.67\pm0.78\%$ for MCTS-UCT and
$93.07\pm1.40\%$ for AFlow. AgentUCT exceeded MCTS-UCT
in seeds 11, 23 and 37, tied it in seed 53, and scored lower in seed 71.
AgentUCT and AFlow had the same largest-budget mean, despite different
seed outcomes. All three frozen budgets and all five seeds are retained. The comparison
covers these allocation policies in the frozen execution system.
AgentUCT seed 71 selected its initial node 0 at all three budgets, reusing
one test measurement. Recursive allocation traces are decision diagnostics;
this repeated endpoint retains its unchanged selection.
The primary denominator is 300 test questions per endpoint; the diagnostic
sentinel is excluded. The shared Qwen3.5-4B, fixed-eight execution groups,
counted-error proposal history, complete 150-question training joins and
up-to-five-to-validation selection are unchanged.

\begin{table}[t]
\centering\small
\caption{Complete GSM8K comparison across 5 paired search seeds. Test accuracy (\%) is reported directly for every seed and all frozen raw-token budgets. Mean $\pm$ sample SD equally weights seeds; SD is not a confidence interval. The same test questions and repeated selected endpoints make budget columns correlated.}
\label{tab:gsm-counted-all-five-seed}
\begin{tabular}{@{}llrrr@{}}
\toprule
Method & Seed & 6.423M & 12.772M & 19.122M\\
\midrule
AgentUCT & 11 & 92.67 & 92.67 & 93.00\\
AgentUCT & 23 & 95.33 & 95.33 & 95.33\\
AgentUCT & 37 & 93.67 & 93.33 & 93.33\\
AgentUCT & 53 & 92.33 & 92.33 & 92.33\\
AgentUCT & 71 & 91.33 & 91.33 & 91.33\\
AgentUCT & Mean $\pm$ SD & $93.07\pm1.52$ & $93.00\pm1.49$ & $93.07\pm1.48$\\
\midrule
MCTS-UCT & 11 & 93.00 & 92.67 & 92.67\\
MCTS-UCT & 23 & 94.00 & 94.00 & 94.00\\
MCTS-UCT & 37 & 91.33 & 92.33 & 92.33\\
MCTS-UCT & 53 & 93.00 & 92.33 & 92.33\\
MCTS-UCT & 71 & 91.67 & 91.67 & 92.00\\
MCTS-UCT & Mean $\pm$ SD & $92.60\pm1.09$ & $92.60\pm0.86$ & $92.67\pm0.78$\\
\midrule
AFlow & 11 & 93.00 & 95.00 & 95.00\\
AFlow & 23 & 88.33 & 92.67 & 92.67\\
AFlow & 37 & 92.33 & 93.00 & 94.00\\
AFlow & 53 & 93.00 & 92.33 & 92.00\\
AFlow & 71 & 91.33 & 91.67 & 91.67\\
AFlow & Mean $\pm$ SD & $91.60\pm1.95$ & $92.93\pm1.26$ & $93.07\pm1.40$\\
\bottomrule
\end{tabular}
\end{table}
\begin{table}[t]
\centering\footnotesize
\setlength{\tabcolsep}{3pt}
\caption{Complete actual phase bills for all 15 GSM8K counted-v3 units. Costs are M raw tokens. Search includes executor, proposer and repair calls; Formal is Search $+$ Validation $+$ Test. All physical calls settled, including the failed calls shown. These full-run charges retain crossing evaluations, failures and retries. Overshoot is contained in Search and is not charged again. Shared preparation (0.109983M raw tokens) is separate and counted once.}
\label{tab:gsm-all15-phase-bills}
\begin{tabular}{@{}lrrrrrrr@{}}
\toprule
Method & Seed & Search & Valid. & Test & Formal & Calls & Failed\\
\midrule
AgentUCT & 11 & 19.432714 & 3.216239 & 0.998435 & 23.647388 & 27213 & 0\\
AgentUCT & 23 & 19.231072 & 4.033069 & 0.630257 & 23.894398 & 32151 & 0\\
AgentUCT & 37 & 19.216682 & 0.905446 & 0.384589 & 20.506717 & 22142 & 0\\
AgentUCT & 53 & 19.252406 & 4.307082 & 0.369586 & 23.929074 & 39624 & 0\\
AgentUCT & 71 & 19.358143 & 7.312719 & 0.149341 & 26.820203 & 25794 & 0\\
\midrule
MCTS-UCT & 11 & 20.704149 & 3.772249 & 0.998435 & 25.474833 & 28341 & 0\\
MCTS-UCT & 23 & 19.348587 & 2.195358 & 0.863685 & 22.407630 & 24844 & 3\\
MCTS-UCT & 37 & 20.092556 & 1.927642 & 0.403501 & 22.423699 & 27746 & 0\\
MCTS-UCT & 53 & 19.415722 & 2.541933 & 1.134356 & 23.092011 & 36146 & 0\\
MCTS-UCT & 71 & 19.168017 & 2.252405 & 0.501162 & 21.921584 & 25427 & 0\\
\midrule
AFlow & 11 & 19.297776 & 3.333025 & 1.797275 & 24.428076 & 29457 & 150\\
AFlow & 23 & 19.610508 & 3.514002 & 1.678639 & 24.803149 & 25171 & 5\\
AFlow & 37 & 19.123214 & 0.701023 & 0.483195 & 20.307432 & 26204 & 4\\
AFlow & 53 & 19.799962 & 5.391294 & 2.272943 & 27.464199 & 29115 & 3\\
AFlow & 71 & 19.515219 & 5.260455 & 0.600111 & 25.375785 & 26705 & 129\\
\bottomrule
\end{tabular}
\end{table}

For these four GSM8K units, the complete training, distinct validation
and test panel counts were AgentUCT seed 71 (52/11/1), MCTS-UCT seed 37 (98/8/2), MCTS-UCT seed 53 (93/9/2), MCTS-UCT seed 71 (72/7/2).
Every physical call settled. Workflow errors remained scored and billed:
AgentUCT seed 71 (1,371 question executions), MCTS-UCT seed 37 (3,181 question executions), MCTS-UCT seed 53 (2,158 question executions), MCTS-UCT seed 71 (978 question executions).
The largest-budget complete crossing and overshoot amounts, respectively,
were AgentUCT seed 71 (0.267421M/0.236143M), MCTS-UCT seed 37 (1.227992M/0.970556M), MCTS-UCT seed 53 (0.777511M/0.293722M), MCTS-UCT seed 71 (0.100444M/0.046017M).
The overshoot is a subset of the billed crossing work. Last affordable
complete snapshots exclude its future rewards and states. Shared preparation
remains 0.109983M raw tokens.

\subsection{Saved checkpoints, allocation traces and complete phase bills}
The following seed-specific tables retain checkpoint counts and affordable
search costs in addition to the complete quality table above. All belong to
the same counted-v3 cohort. Repeated test endpoints are reused observations;
differences in allocation traces are descriptive, without a decision-level
quality attribution.
The counted-v3 seed-11 run recorded recursive allocation through deeper
selection states. The full AgentUCT search, including its crossing evaluation,
completed 92 training panels and recorded 210 traversal decisions, including
127 below the root, reaching selection-state depth four. Completion pricing
changed 185 decisions, including 107 below the root. The retained 19.122M
snapshot ended at iteration 90: its 91 panels contained 207 traversal
decisions, 125 below the root, with 182 cost-induced changes, 105 below the
root. The full MCTS-UCT search completed 60 panels and recorded 79 traversal
decisions, 28 below the root, reaching selection-state depth one. Its 19.122M
snapshot ended at iteration 58 and contained 77 traversal decisions,
27 below the root. All MCTS-UCT choices matched the cost-free index.
AFlow completed 89 panels. Generated-program errors remained scored and billed.

At the 19.122M budget, AgentUCT answered one more test question correctly
than MCTS-UCT in seed 11 (279 versus 278 of 300), a difference of one-third
of a percentage point (Table~\ref{tab:gsm-counted-seed11}). AFlow scored
95\%, two percentage points above AgentUCT's 93\%, and was highest in this
seed. This paragraph describes one paired seed; the five-seed table above
reports across-seed performance. A quality benefit from deeper allocation
remains unmeasured.
AgentUCT, MCTS-UCT and AFlow used nine, seven and eight distinct validation
measurements, respectively, and two distinct test measurements each.
AgentUCT's first two budgets selected node 11 and reused the same test.
The last two budgets for MCTS-UCT and AFlow likewise reused tests of
nodes 34 and 36, respectively. These budgets share the same test
measurements, with statistical replication at the search-seed level.

\begin{table}[t]
\centering\small
\caption{Completed GSM8K results for seed 11 in the counted-v3 cohort.
Each row retains the last complete search checkpoint within budget; $N$
counts 150-question training panels. Correct answers and test accuracy use
300 primary questions after Top5$\to$validation selection. Costs are millions
of raw tokens. These measurements cover one paired seed.}
\label{tab:gsm-counted-seed11}
\setlength{\tabcolsep}{4pt}
\begin{tabular}{@{}lrrrrr@{}}
\toprule
Method & Budget (M) & $N$ & Retained (M) & Correct / 300 & Test acc. (\%)\\
\midrule
AgentUCT & 6.423 & 33 & 6.378792 & 278 & 92.67\\
AgentUCT & 12.772 & 60 & 12.402108 & 278 & 92.67\\
AgentUCT & 19.122 & 91 & 18.863808 & 279 & 93.00\\
\midrule
MCTS-UCT & 6.423 & 30 & 5.982823 & 279 & 93.00\\
MCTS-UCT & 12.772 & 51 & 12.112399 & 278 & 92.67\\
MCTS-UCT & 19.122 & 59 & 18.513759 & 278 & 92.67\\
\midrule
AFlow & 6.423 & 24 & 5.994099 & 279 & 93.00\\
AFlow & 12.772 & 56 & 12.550687 & 285 & 95.00\\
AFlow & 19.122 & 88 & 19.120109 & 285 & 95.00\\
\bottomrule
\end{tabular}
\end{table}

The complete MCTS-UCT bill was 25.474833M raw tokens: 20.704149M for search,
3.772249M for validation and 0.998435M for testing. Its search bill included
0.208821M for proposals and 0.131491M for repairs.
The complete AgentUCT bill was 23.647388M: 19.432714M for search,
3.216239M for validation and 0.998435M for testing. Its search bill included
0.305519M for proposals and 0.240910M for repairs. The complete AFlow bill
was 24.428076M: 19.297776M for search, 3.333025M for validation and 1.797275M
for testing, with 0.317823M for proposals and 0.212644M for repairs included
in search. At the largest budget, the final search increments through
complete evaluations cost 2.190390M for MCTS-UCT, 0.568906M for AgentUCT
and 0.177667M for AFlow. Their respective search-budget overshoots were
1.582149M, 0.310714M and 0.175776M.
These full increments remain billed; their rewards and states are excluded
from the affordable snapshots.

For seed 53, AgentUCT selected node 17 at all three reporting budgets
and obtained 277/300 correct answers (Table~\ref{tab:gsm-counted-aflow-seed53}).
The repeated endpoint reuses one complete test measurement. AFlow selected
nodes 5, 10 and 43 with separate test measurements; its accuracy decreased
from 93\% to 92\% as the budget increased. The complete matched MCTS-UCT and five-seed results are reported in
Table~\ref{tab:gsm-counted-all-five-seed}.

\begin{table}[t]
\centering\small
\caption{Completed AgentUCT and AFlow GSM8K results for seed 53 in the
counted-v3 cohort. Each budget uses its last affordable complete checkpoint
and Top5$\to$validation selection. $N$ counts 150-question training panels;
accuracy uses the same 300 primary test questions. Costs are M raw tokens.
The three AgentUCT rows reuse one test measurement.}
\label{tab:gsm-counted-aflow-seed53}
\setlength{\tabcolsep}{4pt}
\begin{tabular}{@{}lrrrrr@{}}
\toprule
Method & Budget (M) & $N$ & Retained (M) & Correct / 300 & Test acc. (\%)\\
\midrule
AgentUCT & 6.423 & 24 & 6.274982 & 277 & 92.33\\
AgentUCT & 12.772 & 48 & 12.674662 & 277 & 92.33\\
AgentUCT & 19.122 & 99 & 18.640472 & 277 & 92.33\\
AFlow & 6.423 & 18 & 6.155695 & 279 & 93.00\\
AFlow & 12.772 & 35 & 12.543798 & 277 & 92.33\\
AFlow & 19.122 & 48 & 19.011865 & 276 & 92.00\\
\bottomrule
\end{tabular}
\end{table}

The complete seed-53 AgentUCT search evaluated 100 training panels and used
10 distinct validation panels and one test measurement. Its 23.929074M
raw-token bill comprised 19.252406M for search, 4.307082M for validation
and 0.369586M for testing. All 39,624 physical calls were settled, with no
physical call failures; 2,461 question executions with candidate-program
errors remained scored and billed. The last complete crossing evaluation
cost 0.611934M, whereas the search-budget overshoot was 0.130406M.
The crossing charges remain in the full bill, and future rewards and states
are excluded from earlier affordable snapshots.

The complete seed-53 AFlow search evaluated 49 training panels and used
11 distinct validation and three test measurements. Its 27.464199M raw-token
bill comprised 19.799962M for search, 5.391294M for validation and 2.272943M
for testing. Search included 0.232454M for proposals and 0.316211M for
repairs. All 29,115 physical calls were settled, including three failed
calls; 877 question executions with candidate-program errors remained
scored and billed. The last complete crossing evaluation cost 0.788097M,
whereas the amount above the 19.122M budget was 0.677962M. The former
remains in the full bill and is excluded from the retained snapshot.

The AFlow runs for seeds 23 and 71 improved from the first
to the second reporting budget, then reused their selected workflows at
the largest budget (Table~\ref{tab:gsm-counted-aflow-seeds23-71}).
Seed 23 selected nodes 18, 67 and 67; seed 71 selected nodes 0, 40 and 40.
For each seed, the last two budgets therefore share one complete test
measurement. All budgets use the same 300 primary questions, so these
measurements provide correlated endpoints. The complete AgentUCT and MCTS-UCT cohorts are included in
Table~\ref{tab:gsm-counted-all-five-seed}; the seed-23 comparison follows below.

\begin{table}[t]
\centering\small
\caption{Completed AFlow GSM8K results for seeds 23 and 71 in the
counted-v3 cohort. Budgets retain the last affordable complete checkpoint
and use Top5$\to$validation selection. $N$ counts complete 150-question
training panels; accuracy uses 300 primary test questions. Costs are M raw
tokens. Each seed's last two rows reuse one test measurement.}
\label{tab:gsm-counted-aflow-seeds23-71}
\setlength{\tabcolsep}{4pt}
\begin{tabular}{@{}rrrrrr@{}}
\toprule
Seed & Budget (M) & $N$ & Retained (M) & Correct / 300 & Test acc. (\%)\\
\midrule
23 & 6.423 & 43 & 6.366318 & 265 & 88.33\\
23 & 12.772 & 77 & 12.478360 & 278 & 92.67\\
23 & 19.122 & 104 & 18.579361 & 278 & 92.67\\
71 & 6.423 & 24 & 6.314414 & 274 & 91.33\\
71 & 12.772 & 41 & 12.520418 & 275 & 91.67\\
71 & 19.122 & 66 & 18.914536 & 275 & 91.67\\
\bottomrule
\end{tabular}
\end{table}

The full seed-23 bill was 24.803149M raw tokens: 19.610508M for search,
3.514002M for validation and 1.678639M for testing. Search included
0.555961M for proposals and 0.557993M for repairs. Its 105 training panels,
10 distinct validation panels and two test measurements contained 19,352
official question scores. All 25,171 physical calls were settled, including
five failed calls; 1,577 question executions with candidate-program errors
remained scored and billed. The last complete crossing evaluation cost
1.031147M, with a search-budget overshoot of 0.488508M.

For seed 71, the corresponding full bill was 25.375785M: 19.515219M for
search, 5.260455M for validation and 0.600111M for testing. Search included
0.283484M for proposals and 0.212275M for repairs. Its 67 training panels,
12 distinct validation panels and two test measurements contained 14,252
official question scores. All 26,705 physical calls were settled, including
129 failed calls; 1,047 question executions with candidate-program errors
remained scored and billed. The final complete crossing cost 0.600683M,
with a search-budget overshoot of 0.393219M. Both runs retain these crossing
charges in their full bills and exclude future rewards and states from
earlier affordable snapshots.

AgentUCT and MCTS-UCT seed 23 each reused one selected workflow across
all three budgets (Table~\ref{tab:gsm-counted-agentuct-seed23}). AgentUCT
selected node 18 and answered 286/300 questions correctly (95.33\%);
MCTS-UCT selected node 13 and answered 282/300 (94.00\%). The completed
AFlow seed-23 accuracies were 88.33\%, 92.67\% and 92.67\% at the respective
budgets. These results describe one matched three-method seed; repeated
budget endpoints are correlated. The full five-seed comparison is reported
in Table~\ref{tab:gsm-counted-all-five-seed}.

\begin{table}[t]
\centering\small
\caption{Completed AgentUCT and MCTS-UCT GSM8K seed 23. Each budget retains
the last affordable complete checkpoint and uses Top5$\to$validation selection.
$N$ counts 150-question training panels; accuracy uses the same 300 primary
test questions. AgentUCT reuses node 18's one test measurement, and MCTS-UCT
reuses node 13's one test measurement. Costs are M raw tokens.}
\label{tab:gsm-counted-agentuct-seed23}
\setlength{\tabcolsep}{4pt}
\begin{tabular}{@{}lrrrrr@{}}
\toprule
Method & Budget (M) & $N$ & Retained (M) & Correct / 300 & Test acc. (\%)\\
\midrule
AgentUCT & 6.423 & 34 & 6.303265 & 286 & 95.33\\
AgentUCT & 12.772 & 78 & 12.752482 & 286 & 95.33\\
AgentUCT & 19.122 & 123 & 18.919751 & 286 & 95.33\\
MCTS-UCT & 6.423 & 36 & 6.401393 & 282 & 94.00\\
MCTS-UCT & 12.772 & 72 & 11.131243 & 282 & 94.00\\
MCTS-UCT & 19.122 & 95 & 18.979224 & 282 & 94.00\\
\bottomrule
\end{tabular}
\end{table}

The full seed-23 AgentUCT bill was 23.894398M raw tokens: 19.231072M
for search, 4.033069M for validation and 0.630257M for testing. Search
included 0.452297M for proposals and 0.417624M for repairs. Its 124
training panels, eight distinct validation panels and one test measurement
retained all 32,151 settled physical calls, with no physical call failures;
2,190 question executions with candidate-program errors remained scored
and billed. The last complete crossing evaluation cost 0.311321M,
whereas the search-budget overshoot was 0.109072M. The full bill retains
crossing charges, and earlier snapshots exclude future rewards and states.
The original preparation remains separate.

The full seed-23 MCTS-UCT bill was 22.407630M raw tokens: 19.348587M
for search, 2.195358M for validation and 0.863685M for testing. Search
included 18.776162M for workflow execution, 0.313854M for proposals and
0.258571M for repairs. Its 96 training panels, five distinct validation
panels and one test measurement covered 16,201 question executions,
including one diagnostic test sentinel excluded from primary accuracy.
All 24,844 physical calls settled, including three failed calls; 2,083
question executions with workflow errors remained scored and billed.
The last complete crossing evaluation cost 0.369363M, whereas the
search-budget overshoot was 0.226587M. The complete crossing charge
remains in the full bill; earlier affordable snapshots exclude future
rewards and states. The
shared preparation of 0.109983M raw tokens remains separate.

For seed 37, AgentUCT obtained 281, 280 and 280 correct answers across
the three budgets, whereas AFlow obtained 277, 279 and 282
(Table~\ref{tab:gsm-counted-seed37}). AgentUCT reused node 75 at the last
two budgets; AFlow selected nodes 50, 96 and 125 with distinct test
measurements. The ranking therefore changed with budget in this seed.
The matched MCTS-UCT results appear in the complete five-seed comparison
(Table~\ref{tab:gsm-counted-all-five-seed}).

\begin{table}[t]
\centering\small
\caption{Completed AgentUCT and AFlow GSM8K results for seed 37 in the
counted-v3 cohort. Each budget uses the last affordable complete checkpoint
and Top5$\to$validation selection. $N$ counts 150-question training panels;
accuracy uses the same 300 primary test questions. Costs are M raw tokens.
The last two AgentUCT rows reuse one test measurement.}
\label{tab:gsm-counted-seed37}
\setlength{\tabcolsep}{4pt}
\begin{tabular}{@{}lrrrrr@{}}
\toprule
Method & Budget (M) & $N$ & Retained (M) & Correct / 300 & Test acc. (\%)\\
\midrule
AgentUCT & 6.423 & 50 & 6.388173 & 281 & 93.67\\
AgentUCT & 12.772 & 76 & 12.763173 & 280 & 93.33\\
AgentUCT & 19.122 & 116 & 18.959821 & 280 & 93.33\\
AFlow & 6.423 & 52 & 6.362538 & 277 & 92.33\\
AFlow & 12.772 & 98 & 12.732002 & 279 & 93.00\\
AFlow & 19.122 & 159 & 19.102452 & 282 & 94.00\\
\bottomrule
\end{tabular}
\end{table}

The full seed-37 AgentUCT bill was 20.506717M raw tokens: 19.216682M
for search, 0.905446M for validation and 0.384589M for testing. Search
included 0.440943M for proposals and 0.399368M for repairs. Its 117
training panels, seven distinct validation panels and two test measurements
retained all 22,142 settled physical calls, with no physical call failures;
3,019 question executions with candidate-program errors remained scored
and billed. The last complete crossing evaluation cost 0.256861M,
whereas the search-budget overshoot was 0.094682M.

For AFlow seed 37, the full bill was 20.307432M: 19.123214M for search,
0.701023M for validation and 0.483195M for testing. Search included
0.804010M for proposals and 0.629095M for repairs. Its 160 training panels,
ten distinct validation panels and three test measurements retained all
26,204 settled physical calls, including four failed calls; 2,312 question
executions with candidate-program errors remained scored and billed.
The last complete crossing evaluation cost 0.020762M, whereas the
overshoot was 0.001214M. Both runs preserve these crossing charges in
their full bills and exclude future rewards and states from earlier
affordable snapshots. The original 109,983-token preparation remains
separate; neither run required new preparation or calibration.

Saved AgentUCT seed-37 traversals contain 294 parent-state choices,
186 below the root; completion pricing changes 253 choices, including
148 below the root. At the largest affordable snapshot, the corresponding
counts are 290, 183, 250 and 146. The remaining four choices belong to
the crossing evaluation and are excluded from that snapshot. The highest
saved parent depth is four in both scopes. These are recorded allocation
decisions, not program depth or a causal estimate of their quality benefit.

\section{Detailed comparison protocols}
\label{sec:followup-design}
Eleven RAG methods have complete five-seed results: the four MCTS
allocators, AgentUCT-StaticFull and six common-space controls.
The three fixed pipeline references and fifty prespecified random workflows
also have complete test measurements. The complete five-seed GSM8K results are reported in the supplementary
experimental tables.

\subsection{Direct algorithm comparison in the RAG space}
\label{sec:rag-comparators}
This comparison tests whether completion pricing remains useful when the
alternative is model-guided editing, surrogate optimization or explicit
stage allocation. Every optimizer uses the same 61,344-workflow grammar,
150 training questions, 300 validation questions and 300 primary test
questions. Qwen3.5-4B serves all generative roles, and the same retrieval
components and persistent prefix execution service are available to every
method. Search seeds are $\{11,23,37,53,71\}$. The endpoint at each of
16M, 24M and 32M actual weighted-token units is selected from training Top5
by validation F1, then evaluated on the primary test panel.

The original twenty runs supply four completed rows in
Table~\ref{tab:direct-algorithm-coverage-complete}. The six additional algorithms
supply 30 completed method--seed searches and 90 budget-level quality cells.
Complete five-seed results are available for AFlow--RAG, uniform-terminal
search, FLASH*--RAG (TPE), EEIPU, LaMBO and MSBO. Their paired results
appear in the main text and supplement. The comparison is between optimization
algorithms within this RAG system. Each implementation preserves specified decisions
from its source algorithm while changing the search domain, execution service
and accounting unit. These boundaries are part of the experimental definition,
with implementation details in the Supplementary Information\ifdefined\AgentUCTWithAppendix
\space(Section~\ref{sec:rag-optimizer-adapters})\fi.

\begin{table*}[t]
\centering\small
\caption{Direct RAG comparisons and AgentUCT ablations. Test F1 (\%) is mean
$\pm$ sample SD over five paired seeds after training Top5 and validation
selection. All rows use the common RAG protocol.}
\label{tab:direct-algorithm-coverage-complete}
\begin{tabularx}{\textwidth}{@{}lcccX@{}}
\toprule
Method & 16M & 24M & 32M & Selection mechanism in the common grammar\\
\midrule
AgentUCT & $67.40\pm2.54$ & $69.57\pm0.28$ & $69.57\pm0.28$ & Adaptive full remaining-cost estimate\\
MCTS-UCT & $67.38\pm2.52$ & $68.01\pm1.12$ & $68.01\pm1.12$ & UCB allocation\\
AgentUCT-NextC & $67.40\pm2.54$ & $67.83\pm2.69$ & $68.22\pm1.96$ & First positive unpaid cost\\
AgentUCT-StaticFull & $67.40\pm2.54$ & $67.36\pm2.67$ & $67.36\pm2.67$ & AgentUCT ablation with a frozen complete-cost estimator; live execution reuse\\
MCTS-Random & $67.40\pm2.54$ & $68.44\pm0.89$ & $68.80\pm0.87$ & Uniform child allocation\\
AFlow--RAG & $65.93\pm1.91$ & $66.38\pm1.24$ & $67.25\pm1.57$ & Top-four parent mixture and Qwen edits with training feedback\\
FLASH*--RAG (TPE) & $66.49\pm1.70$ & $66.96\pm0.97$ & $67.29\pm0.85$ & D-optimal design, linear EIPS and categorical TPE\\
EEIPU--RAG & $67.15\pm2.17$ & $68.19\pm0.67$ & $67.29\pm0.78$ & Objective and stage-cost GPs with prefix-conditioned proposals\\
LaMBO--RAG & $64.78\pm0.96$ & $65.87\pm2.29$ & $66.11\pm2.62$ & Slowly moving region sampling and local GP-UCB\\
MSBO--RAG & $67.20\pm0.46$ & $67.93\pm2.57$ & $67.44\pm2.53$ & Cascade posterior with native start/advance execution\\
Uniform-terminal search & $67.08\pm0.76$ & $66.68\pm0.82$ & $67.18\pm1.59$ & Uniform complete workflows, training Top5 and validation selection\\
\bottomrule
\end{tabularx}
\end{table*}

\paragraph{AFlow--RAG}
AFlow's parent choice and model-directed expansion~\citep{zhang2024aflow}
operate on complete RAG tuples. Among the $K\leq4$ highest-scoring completed
parents, parent $i$ is sampled with probability
$0.3/K+0.7\exp(20F_i)/\sum_j\exp(20F_j)$, where $F_i$ is its own training
F1. The editor receives the selected tuple, that parent's modification
history and up to three training examples containing the question,
prediction, gold answer and F1. It proposes a legal tuple and an explanation
of the edit. Qwen proposals use temperature 0.7, top-$p=0.9$ and a
2,048-token output cap; execution remains greedy. Each round permits three
attempts in total. Illegal tuples, previously evaluated complete workflows
and repeated parent-specific modification text trigger repair. An exhausted
round supplies no quality observation, while every paid attempt remains in
search cost. The fixed initial BM25 workflow is also evaluated and charged.

\paragraph{FLASH*--RAG (TPE)}
The implementation follows FLASH's design, algorithm-selection and
parameter-tuning phases~\citep{zhang2016flash}. Greedy D-optimal design
occupies the first 15\% of the 32M search budget; additive ridge quality and
log-cost models guide EIPS selection until 30\%; categorical tree-structured
Parzen estimation (TPE) uses the remaining 70\%. These are transition
thresholds on completed actual bills, so a complete evaluation can cross a
phase boundary. The algorithm graph is pruned using its ten highest-ranked
structural signatures. Hyperopt's categorical density ratio then selects an
unseen legal configuration in that graph. The star identifies the TPE variant;
the acquisition's log-cost convention and the finite-grammar pruning rule
are specified in the appendix.

\paragraph{EEIPU--RAG}
EEIPU~\citep{essofi2024memoization} learns training F1 with one Gaussian
process (GP) and execution log-cost with three independent stage GPs. The
stages are query construction, evidence construction and answer generation.
After five complete warmup workflows, the current training Top5 supplies
cached prefixes for candidate generation, alongside an empty prefix for
exploration. Expected improvement is multiplied by expected inverse
remaining cost raised to $\max(0,1-C/32\mathrm{M})$, using 128 Monte Carlo
draws and an approximately 1,024-candidate pool. Cache discounts require a
materialized prefix for every training question. A one-token acquisition
floor stabilizes the surrogate while actual zero-cost reuse remains zero
in the physical bill.

\paragraph{LaMBO--RAG}
LaMBO~\citep{lin2021lambo} couples local GP-UCB choices to slowly moving
sampling over a modular partition tree. The first two of the same three
RAG modules are each split into two approximately equal-sized categorical regions,
giving fixed tree depths $(1,1)$. After five complete warmup workflows,
local acquisition uses $\mu+2\sigma$ and the multiplicative bandit update
uses learning rate 0.05. The lazy update retains the preceding workflow's
earlier modules when a later region changes. Finite exhausted regions are
masked, with an explicit return to the root when needed. This tests the
modular allocation mechanism with fixed partitions and the common persistent
cache. Its preceding-configuration switching model is distinct from the
physical cache bill reported here.

\paragraph{MSBO--RAG}
MSBO~\citep{torresi2026multistage} can start an evidence prefix or advance a
saved prefix to an answer. The first stage executes the first six actions,
through evidence reranking, on all 150 questions. Its two observations are
the mean grounded answerability judgment and the mean fraction of evidence
documents cited by the grounded assessment. Neither uses a gold answer or
a terminal F1. The second stage restores those native execution states,
chooses the retained evidence and context order, and generates all answers
before training F1 is released. Independent stage GPs propagate a cascade
posterior with 256 Monte Carlo draws after five complete warmup workflows.
The acquisition uses expected improvement, falling back to UCB when its
maximum is at most 0.005. UCB follows the source rule
$\beta_{\mathrm{MSBO}}(C)=1+100\max(0,1-C/32\mathrm{M})$.

The source algorithm's final full-path decision is retained by mapping its
$100+5$ cost epochs to the 32M reporting budget. When the remaining budget
is at most $32\mathrm{M}/105$, the next acquisition chooses a complete
cascade; the same choice is made every 1,000 acquisition decisions. Once
selected, both stages of that path finish even if the evidence stage crosses
the budget. Its suffix is fixed by that decision. Ordinary stage decisions
update only after all 150 questions have joined. This preserves
start/advance control while the tree-search arms, AFlow and the complete-workflow
surrogates retain a full-workflow feedback barrier.

\paragraph{Snapshots and interpretation.}
For every method, the primary budget result uses the last complete terminal
workflow snapshot whose cumulative search bill is at most the reporting
budget. Its shortlist, cache and optimizer state are frozen at that boundary.
MSBO additionally retains stage-boundary snapshots to measure paid evidence
work and unfinished samples; these cannot contribute a partial F1 to the
primary comparison. Physical bills include initialization, intermediate
stages, model-directed proposals, repairs and paid failures as applicable.
Crossing evaluations remain in the full bill, and all validation and test
charges are reported separately. Optimizer-selection elapsed time is instrumented separately from task execution; no timing comparison is reported here. Uniform-terminal search
draws distinct legal complete paths uniformly and uses the same training
Top5 and validation endpoint rule. The resulting comparisons isolate
allocation within a shared system; they do not claim reproduction of the
original papers' complete benchmark systems or native cache policies.

\subsection{Published pipelines, random composition and strong planners}
\label{sec:workflow-references}
Fixed pipelines test whether optimization improves on an established
combination of components. This study compares selected configurations with
three published combinations: Query2doc
query expansion followed by BM25 retrieval~\citep{wang2023query2doc}, HyDE
hypothetical-document generation followed by dense retrieval
\citep{gao2023hyde}, and RankGPT-style BM25 candidate retrieval followed by
listwise language-model reranking~\citep{sun2023rankgpt}. Each paper itself
specifies a combination represented by the current grammar.

All three share the candidate documents, Qwen3.5-4B executor, grounded
evidence assessment, retrieval width six, one retrieval round, four retained
documents and ranked context. The supplement lists their exact actions and
changes from the originals: Query2doc's prompt and query repetition,
HyDE's encoder and single generated passage, and RankGPT's model and ranking
window. They are shared-space pipeline methods, with no system-level
reproduction claim. Each runs directly on the same 300 primary test
questions; all three F1/EM values, paired effects and execution bills are
reported separately. No test outcome selects a reference.

\paragraph{Random complete workflows.}
Using seed 20261004, we sampled fifty distinct workflows before testing.
Uniform draws over the 61,344 legal terminals give each workflow equal
probability; uniform child choices can give different leaf probabilities.
The experiment reports the entire test-quality distribution, its mean,
median, spread and costs, retaining all fifty prespecified draws as
the reference distribution. This reference asks how much quality is available from
unoptimized composition. MCTS-Random allocation and uniform-terminal search
answer different algorithmic questions: both spend search tokens and select
an endpoint through the declared training and validation protocol.
The three fixed pipelines plus fifty random workflows require 15,900 primary
question--workflow executions before reuse; sentinels are separate.

\paragraph{Strong-model planning, deferred.}
DeepSeek-V4-Pro and GPT-5.5 were specified to propose five independently
recorded plans each from identical task instructions, grammar, component
descriptions and three frozen training examples. They receive no primary
test questions, answers, predictions or reference scores. Each plan is
frozen before Qwen3.5-4B executes it; no quality-feedback optimization or
validation selection is allowed. At most one syntax-only repair follows an
invalid response, and every attempt remains in the bill. An unrepaired
plan receives zero task quality in the intent-to-run mean, alongside a
separately reported valid-plan mean.

The requested identifiers are native \path{deepseek-v4-pro} and OpenRouter's
\path{openai/gpt-5.5}~\citep{deepseek2026v4pro,openai2026gpt55}, both with
high reasoning effort. Their first real requests were rejected for account
balance and regional availability, respectively. The references are now
deferred by the authors. The infrastructure failures produced no completed
planning outcome or measured planner quality. A later
campaign will retain all five plans per model, provider identifiers, prompts,
reasoning settings and actual usage, with API currency/raw tokens separate
from the local executor's weighted-token accounting.

\paragraph{Test-level inference.}
Algorithm comparisons retain five seed values, mean $\pm$ sample SD and
paired effect sizes, without algorithm-level significance testing.
For reference $r$ and question $q$, define
\[
 d_{r,q}=\frac{1}{5}\sum_s F_{s,q}^{\mathrm{AgentUCT},32\mathrm{M}}-F_{r,q},
\]
where $F_{r,q}$ is one fixed pipeline's score or the mean score across all
fifty random workflows. The primary effect is $300^{-1}\sum_q d_{r,q}$.
A paired bootstrap of question IDs supplies a conditional 95\% interval;
a paired sign-randomization test supplies a two-sided test of the declared
exchangeability null. Each question's full seed/workflow vector moves
together. Holm adjustment covers the four active contrasts: three fixed
pipelines and the random-composition mean. Planner contrasts are deferred.
The 32M endpoint is fixed; 16M and 24M remain descriptive sensitivity results.
Test outcomes never select a reference, budget or parameter value.

The intervals describe question variation for the observed workflows.
Variation across workflows and sensitivity to resampling both workflows
and questions are reported separately, as is search-seed variation.
The $5\times300$ score cells comprise 1,500 measurements on shared
questions, with dependent scores within each question's seed/workflow vector.
Repeated budgets reuse their test samples.
Because the primary panel has already been inspected during development,
these are retrospective comparisons on that panel.

\subsection{Mechanism tests and execution order}
The two completed five-seed single-factor controls separately freeze
calibrated action prices and replace historical continuation by uniform
legal-child sampling, preserving live cache conditioning. Their direct scores,
affordable complete-evaluation counts and full phase bills are reported in
Section~\ref{sec:single-factor-mechanism}. The completed AgentUCT-StaticFull
comparison (Section~\ref{sec:static-full-results}) remains a joint intervention
on prices, continuation demand and cache conditioning, with live execution
reuse.

Saved RAG traces place the observed allocation changes at the root after the
common initialization. Saved AgentUCT GSM8K seed-37 traversals reached
selection-state depth four and record cost-induced decisions below the root
(see the supplementary experimental design). Selection depth describes where
the allocator compares alternatives, separately from generated-program depth.
These decisions establish recursive allocation changes without isolating
their contribution to endpoint quality.

The complete five-seed GSM8K comparison measures transfer to generated
programs. AgentUCT had a modest largest-budget mean advantage over MCTS-UCT
and the same mean as AFlow. The supplementary tables retain every
seed and frozen budget, including opposite seed-level outcomes and reused
test endpoints. The observed quality ordering varies with task and seed. The completed five-seed AFlow--RAG comparison remains separate
competitive evidence within the shared discrete space. Additional synthetic
matrices and cache-disabled historical replays are deferred.

\section{Original RAG Results and Component Catalogue}\label{sec:rag-full-results}
\label{app:experiments}
The completed RAG studies optimize component workflows. AFlow--RAG generates
and charges legal edits within the same grammar. The separate GSM8K study
optimizes generated Python workflows; earlier fixed-library and bounded
generated-workflow pilots remain diagnostic and are excluded from its formal cohort.
Every generative execution role uses the same Qwen3.5-4B checkpoint: query refinement,
progress assessment, listwise reranking, answer generation, workflow review
and repair. The dynamic AFlow study generates and charges Python-workflow edits.
Dedicated embedding models remain retrieval components. All inference runs on
the experiment server. Up to seven GPUs run Qwen workers; the eighth hosts the shared E5, BGE and Qwen
embedding service. Each trial owns its document indices and online caches;
sharing model residency creates no cross-trial reuse credit. Direct small-model component selection is
outside this study.
The deferred planning references specify DeepSeek-V4-Pro and GPT-5.5 as
offline planners, with proposed workflows executed by Qwen3.5-4B. These
references currently have no completed measurements; their protocol separates
planner and execution charges.

The paired RAG evaluator fixes bfloat16 inference, greedy generation in fixed
query groups with model batches capped at four, encoder batches capped at eight for both index construction and
query encoding, model context limits and deterministic ranking tie rules.
Document indices are built under this common configuration before search.
Changing these evaluator settings defines a separate experimental cohort.
The completed batch cohort started all four methods and five seeds afresh,
with new search trees and responses. Earlier single-request runs remain
a separate cohort.

\subsection{Conceptual comparison with related optimizers}
Table~\ref{tab:related} summarizes the design models discussed in Related Work.
The completed five-seed RAG comparisons contain AgentUCT, MCTS-UCT,
AgentUCT-NextC, AgentUCT-StaticFull, MCTS-Random, AFlow,
uniform-terminal search, FLASH*--RAG (TPE), EEIPU--RAG, LaMBO--RAG and
MSBO--RAG. The separate formal GSM8K study
compares AFlow, MCTS-UCT and AgentUCT within the same dynamic execution system;
its earlier pilot supplies execution evidence only.
\begin{table*}[t]
\centering\small
\caption{Conceptual comparison of cost and execution-state models, not an
empirical benchmark. Rows describe the cited methods' allocation mechanisms
and analysis targets; Related Work supplies their references.}
\label{tab:related}
\begin{tabularx}{\textwidth}{@{}l XXX@{}}
\toprule
Method & Cost and reusable state & Allocation statistic & Analysis target\\
\midrule
FLASH & Pipeline-prefix intermediates; LRU eviction & Cost-sensitive BO acquisition & Efficient pipeline optimization\\
EEIPU & Selected memoized prefixes; stage-cost GPs & EI and inverse remaining cost & Budgeted BO with memoization\\
LaMBO & Modular change from the preceding configuration & Slowly moving bandit and local BO & Switching-cost-regularized regret\\
MSBO & Inventory of partial cascades and observations & Stage models; start or advance a sample & Multistage evaluation efficiency\\
AFlow & Generated program and execution feedback & Parent score/uniform mixture & Workflow optimization\\
AgentUCT & Persistent qid--prefix demands shared by panels & Raw ancestor returns and completion-cost price & Demand-information boundary; same-state tradeoff; fixed-table quality regret\\
\bottomrule
\end{tabularx}
\end{table*}

\subsection{Direct AFlow comparison in the common RAG space}
The AFlow implementation proposes edits within the same 61,344-workflow grammar,
using Qwen3.5-4B and completed training feedback. Its parent selection retains
the score/uniform mixture. Proposal and repair calls enter the actual search
bill, and every accepted workflow completes the full 150-question panel
before it can supply feedback for another proposal. All methods use the same
three actual-cost cutoffs and Top5--validation--test selection. This comparison
tests allocation within a common component space; the generated Python-workflow
experiment on GSM8K tests transfer within the AFlow execution system.

AgentUCT exceeds the AFlow policy at 24M and 32M in all five paired
seeds (Table~\ref{tab:completed-aflow-paired}). At 32M, the differences are
1.24, 3.71, 3.25, 3.22 and 0.18 percentage points for seeds
11, 23, 37, 53 and 71, respectively, with mean $2.32\pm1.53$ points.
AFlow's five-seed mean F1 is $65.93\pm1.91\%$, $66.38\pm1.24\%$
and $67.25\pm1.57\%$ at 16M, 24M and 32M.
For seeds 11 and 23, AFlow retains 19 and 21 complete evaluations, respectively,
compared with AgentUCT's 18 and 19: the quality advantage in these runs does
not require more completed evaluations. At 16M, AFlow leads by 2.61 points
in seed 53, where the cutoff primarily covers common initialization for
AgentUCT and MCTS-UCT. At 16M, AgentUCT leads in three seeds and AFlow
in two. The ordering varies with budget and seed.

\begin{table*}[t]
\centering\small
\caption{Completed paired AFlow--RAG comparisons. Entries are primary test
F1 in percent under the common actual search-budget ceiling and Top5
validation protocol. The methods are compared directly by their test F1.
All five matched seeds are shown; repeated budget endpoints are not
independent measurements.}
\label{tab:completed-aflow-paired}
\begin{tabular}{@{}rrrrr@{}}
\toprule
Seed & Budget (M) & AgentUCT & MCTS-UCT & AFlow\\
\midrule
11 & 16 & 69.51 & 69.44 & 66.25\\
11 & 24 & 69.51 & 69.44 & 65.76\\
11 & 32 & 69.51 & 69.44 & 68.27\\
23 & 16 & 67.05 & 67.05 & 65.76\\
23 & 24 & 69.52 & 67.32 & 65.76\\
23 & 32 & 69.52 & 67.32 & 65.81\\
37 & 16 & 69.20 & 69.20 & 63.22\\
37 & 24 & 69.20 & 68.69 & 65.76\\
37 & 32 & 69.20 & 68.69 & 65.95\\
53 & 16 & 63.21 & 63.21 & 65.81\\
53 & 24 & 69.97 & 66.58 & 66.02\\
53 & 32 & 69.97 & 66.58 & 66.75\\
71 & 16 & 68.01 & 68.01 & 68.59\\
71 & 24 & 69.63 & 68.01 & 68.59\\
71 & 32 & 69.63 & 68.01 & 69.45\\
\bottomrule
\end{tabular}

\end{table*}

Complete search bills, including proposal and repair, are 32.187M and
33.429M weighted tokens for seeds 11 and 23. Their last affordable 32M
snapshots cost 30.195M and 31.881M. The complete crossing steps cost 1.991M
and 1.549M, whereas the amounts above the ceiling are 0.187M and 1.429M.
Validation costs a further 29.319M and 20.773M; test costs 9.698M and 5.552M.
All 42 and 39 rejected proposal rounds remain charged. Previously paid
input-index preparation is cited separately, including the reused indices
in seed 23.
For seeds 37 and 53, complete search costs 32.739M and 33.184M; the 32M
snapshots retain 31.013M and 31.542M. Their full crossing evaluations cost
1.726M and 1.642M, with 0.739M and 1.184M above the ceiling.
Validation costs 31.315M and 21.950M, and test costs 8.992M and 8.289M.
All 26 and 42 rejected proposal rounds remain charged; previously paid
input preparation is accounted for separately.
For seed 71, complete search costs 33.769M and the 32M snapshot retains
31.814M. The full crossing evaluation costs 1.955M, of which 1.769M is
above the ceiling. Validation and test cost 30.119M and 7.741M,
respectively. All twelve rejected proposal rounds remain charged.
The five completed runs retain 21.0 evaluations on average at 32M,
compared with AgentUCT's 17.6. These counts concern affordable complete
evaluations; all crossing charges remain in the full bills.

\subsection{Completed common-space extension measurements}
Table~\ref{tab:completed-seed11-extensions} reports each completed optimizer
trial alongside the 32M test F1 of AgentUCT and MCTS-UCT with the same seed.
The shared grammar, evaluator, actual-cost cutoffs and held-out selection
protocol are fixed; each optimizer retains its declared initialization and
allocation policy. The table reports all 25 individual trials for these five controls. The original four-method comparison, AFlow and
AgentUCT-StaticFull have separate completed five-seed comparisons.
Uniform-terminal, FLASH*, EEIPU, LaMBO and MSBO have complete five-seed cohorts
(Table~\ref{tab:uniform-flash-complete}).
MSBO--RAG reaches 69.93\% at 32M in seed 23, 0.41 percentage points above
AgentUCT's 69.52\% with the same seed. Across all five seeds, AgentUCT leads
MSBO by $2.13\pm2.51$ points at 32M, with four wins and one loss.

\begin{table*}[t]
\centering\small
\caption{Completed common-space optimizer trials. F1 is primary test performance
in percent after Top5 validation selection. The AgentUCT and MCTS-UCT columns
show their same-seed 32M test F1 directly. $N_{32}$ and cost count complete evaluations and M weighted-token units
in the last affordable 32M snapshot. Each row is one trial; repeated endpoints
across budgets reuse measurements.}
\label{tab:completed-seed11-extensions}
\begin{tabular}{@{}lrrrrrrrr@{}}
\toprule
Method & Seed & 16M F1 & 24M F1 & 32M F1 & \shortstack{AgentUCT\\32M F1} & \shortstack{MCTS-UCT\\32M F1} & $N_{32}$ & Cost (M)\\
\midrule
FLASH*--RAG (TPE) & 11 & 67.79 & 66.55 & 66.55 & 69.51 & 69.44 & 19 & 31.761\\
FLASH*--RAG (TPE) & 23 & 64.15 & 65.67 & 66.28 & 69.52 & 67.32 & 24 & 31.556\\
FLASH*--RAG (TPE) & 37 & 65.38 & 67.47 & 67.47 & 69.20 & 68.69 & 17 & 31.744\\
FLASH*--RAG (TPE) & 53 & 66.86 & 66.86 & 67.91 & 69.97 & 66.58 & 13 & 30.497\\
FLASH*--RAG (TPE) & 71 & 68.25 & 68.25 & 68.25 & 69.63 & 68.01 & 22 & 31.980\\
\addlinespace
EEIPU--RAG & 11 & 67.95 & 67.95 & 67.26 & 69.51 & 69.44 & 17 & 29.469\\
EEIPU--RAG & 23 & 67.72 & 68.09 & 68.09 & 69.52 & 67.32 & 24 & 30.432\\
EEIPU--RAG & 37 & 69.13 & 68.05 & 66.01 & 69.20 & 68.69 & 19 & 31.914\\
EEIPU--RAG & 53 & 67.53 & 67.53 & 67.53 & 69.97 & 66.58 & 24 & 29.968\\
EEIPU--RAG & 71 & 63.43 & 69.31 & 67.59 & 69.63 & 68.01 & 22 & 31.227\\
\addlinespace
LaMBO--RAG & 11 & 65.18 & 64.60 & 65.63 & 69.51 & 69.44 & 37 & 31.112\\
LaMBO--RAG & 23 & 66.16 & 66.36 & 66.36 & 69.52 & 67.32 & 35 & 30.859\\
LaMBO--RAG & 37 & 64.72 & 69.59 & 67.29 & 69.20 & 68.69 & 19 & 29.481\\
LaMBO--RAG & 53 & 64.10 & 65.07 & 69.19 & 69.97 & 66.58 & 26 & 31.037\\
LaMBO--RAG & 71 & 63.72 & 63.72 & 62.07 & 69.63 & 68.01 & 34 & 31.491\\
\addlinespace
MSBO--RAG & 11 & 67.82 & 68.59 & 68.59 & 69.51 & 69.44 & 14 & 29.855\\
MSBO--RAG & 23 & 67.08 & 69.93 & 69.93 & 69.52 & 67.32 & 17 & 26.855\\
MSBO--RAG & 37 & 66.77 & 68.66 & 66.64 & 69.20 & 68.69 & 12 & 27.007\\
MSBO--RAG & 53 & 67.51 & 69.02 & 68.59 & 69.97 & 66.58 & 15 & 30.116\\
MSBO--RAG & 71 & 66.81 & 63.43 & 63.43 & 69.63 & 68.01 & 19 & 31.010\\
\addlinespace
Uniform-terminal search & 11 & 66.76 & 66.76 & 69.33 & 69.51 & 69.44 & 18 & 30.749\\
Uniform-terminal search & 23 & 66.28 & 66.28 & 66.28 & 69.52 & 67.32 & 16 & 31.269\\
Uniform-terminal search & 37 & 67.99 & 67.99 & 67.99 & 69.20 & 68.69 & 20 & 31.268\\
Uniform-terminal search & 53 & 67.78 & 65.78 & 65.17 & 69.97 & 66.58 & 16 & 31.550\\
Uniform-terminal search & 71 & 66.58 & 66.58 & 67.11 & 69.63 & 68.01 & 14 & 31.804\\
\bottomrule
\end{tabular}

\end{table*}

\begin{table*}[t]
\centering\small
\caption{Complete five-seed comparisons with five common-space sampling, surrogate and stage-allocation controls. Test F1 is directly compared with AgentUCT at each budget, as mean $\pm$ sample SD across five search seeds. Mean $N$ counts affordable complete evaluations for the listed method; budgets can reuse endpoints and are not independent replications.}
\label{tab:uniform-flash-complete}
\begin{tabular}{@{}llrrr@{}}
\toprule
Method & Budget & AgentUCT F1 (\%) & Method F1 (\%) & Mean $N$\\
\midrule
Uniform-terminal & 16M & $67.40\pm2.54$ & $67.08\pm0.76$ & 8.2\\
Uniform-terminal & 24M & $69.57\pm0.28$ & $66.68\pm0.82$ & 12.4\\
Uniform-terminal & 32M & $69.57\pm0.28$ & $67.18\pm1.59$ & 16.8\\
FLASH*--RAG (TPE) & 16M & $67.40\pm2.54$ & $66.49\pm1.70$ & 8.0\\
FLASH*--RAG (TPE) & 24M & $69.57\pm0.28$ & $66.96\pm0.97$ & 13.2\\
FLASH*--RAG (TPE) & 32M & $69.57\pm0.28$ & $67.29\pm0.85$ & 19.0\\
EEIPU--RAG & 16M & $67.40\pm2.54$ & $67.15\pm2.17$ & 12.6\\
EEIPU--RAG & 24M & $69.57\pm0.28$ & $68.19\pm0.67$ & 16.2\\
EEIPU--RAG & 32M & $69.57\pm0.28$ & $67.29\pm0.78$ & 21.2\\
LaMBO--RAG & 16M & $67.40\pm2.54$ & $64.78\pm0.96$ & 11.4\\
LaMBO--RAG & 24M & $69.57\pm0.28$ & $65.87\pm2.29$ & 20.6\\
LaMBO--RAG & 32M & $69.57\pm0.28$ & $66.11\pm2.62$ & 30.2\\
MSBO--RAG & 16M & $67.40\pm2.54$ & $67.20\pm0.46$ & 8.8\\
MSBO--RAG & 24M & $69.57\pm0.28$ & $67.93\pm2.57$ & 12.4\\
MSBO--RAG & 32M & $69.57\pm0.28$ & $67.44\pm2.53$ & 15.4\\
\bottomrule
\end{tabular}

\end{table*}

At 32M, the paired AgentUCT advantages over EEIPU and LaMBO are
$2.27\pm0.64$ and $3.46\pm2.58$ percentage points (mean $\pm$ sample SD),
positive in every seed. LaMBO nevertheless exceeds AgentUCT by 0.389 points
in seed 37 at 24M. The fifteen budget cells for EEIPU and LaMBO reference
twelve and thirteen distinct complete selected test measurements,
respectively; statistical replication remains at the five search seeds.
In seed 71, full search/validation/test bills are
33.186M/41.059M/15.392M for EEIPU and 33.987M/20.580M/7.764M for LaMBO.
Their complete crossing evaluations cost 1.959M and 2.497M, whereas their
32M overshoots are 1.186M and 1.987M. Paid input preparation is separate;
these bills cover all three cutoffs' measurement campaign. The original
32M-only optimization utility uses its own declared phase bills.

FLASH*--RAG uses two D-optimal initial evaluations, three linear-EIPS
evaluations and fifteen TPE evaluations in seed 11. The pruned tuning
domain contains 2,304 legal workflows. The asterisk denotes the declared
common search system. Complete search costs 33.716M weighted tokens;
validation and test cost 33.036M and 7.452M, respectively. The last complete
evaluation adds 1.955M, with 1.716M above the 32M ceiling. EEIPU--RAG performs
five warmup and thirteen acquisition evaluations in seed 11, with complete search,
validation and test bills of 33.310M, 41.727M and 15.611M. Its runtime
handoff preserves the completed search and all previously paid calls.
Uniform-terminal search samples legal complete workflows without replacement
and evaluates nineteen of them in seed 11. Its search, validation and test
bills are 32.526M, 41.589M and 9.105M; the last complete evaluation adds
1.778M, taking search 0.526M above the ceiling. It uses training evaluations
and the common Top5 protocol, whereas the separate fifty-workflow reference
directly measures a prespecified test distribution.
Shared input-index preparation is cited separately for these methods.
LaMBO--RAG completes 38 training panels in seed 11: five warmup evaluations
and 33 local GP-UCB evaluations under slowly moving region sampling.
Its 16M, 24M and 32M snapshots retain 14, 25 and 37 complete evaluations,
with test F1 of 65.18\%, 64.60\% and 65.63\%. At 32M the retained bill
is 31.112M; full search costs 33.058M, including a complete 1.946M
crossing evaluation and 1.058M above the ceiling. These bills describe seed 11.
The implementation uses fixed partition depths $(1,1)$, a finite categorical
domain and persistent WTB reuse. Physical execution bills provide the implementation's accounting; the original
numerical switching-cost objective and its helper remain unused. Saved region probabilities, conditional sampling,
lazy-prefix eligibility and completed-feedback updates are audited;
unrecorded GP acquisition rankings and SMB random symbols remain unavailable.
MSBO--RAG completes 26 stages in seed 11: twelve evidence stages and
fourteen terminal evaluations. Every stage joins all 150 training questions
before its feedback enters the optimizer. Its primary snapshots retain
9, 10 and 14 complete workflows at 16M, 24M and 32M; test F1 is
67.82\%, 68.59\% and 68.59\%. The corresponding terminal-boundary bills
are 13.527M, 19.058M and 29.855M. An evidence-only snapshot records
additional paid work; terminal quality becomes available after the full answer stage.
Full search costs 33.277M, including the final evidence-only stage of
3.422M; the amount above 32M is 1.277M. Validation and test cost 33.994M
and 12.352M across the three-cutoff measurement campaign. The stage audit
checks native evidence proxies, per-question reuse of the six-action prefix
and complete-panel updates. Unsaved GP candidate
rankings and Monte Carlo acquisition draws remain outside the audit evidence.
In seed 71, MSBO completes 34 stages: fifteen evidence stages and
nineteen terminal evaluations. Its 32M primary and auxiliary snapshots
both end at stage 33 with a bill of 31.010M. The final evidence stage
costs 1.019M, taking full search to 32.029M; the overshoot is 0.029M.
Validation and test cost 28.610M and 6.970M across the three cutoffs.
The 5.613M cost of all unadvanced evidence inventory spans the full search
and is distinct from this final crossing stage. Paid input preparation
remains separate. Across the five seeds, the fifteen MSBO budget cells
reference twelve complete selected test measurements; statistical
replication remains at the five search seeds.

In seed 23, MSBO completes thirteen evidence stages and seventeen terminal
evaluations. Its 32M primary snapshot ends at stage 28 with a bill of
26.855M. Stage 29 adds evidence and raises the bill to 30.284M without
another complete workflow evaluation; it belongs to the auxiliary snapshot.
Stage 30 costs 3.558M and crosses the ceiling, bringing full search to
33.842M. The 6.987M paid after the last terminal snapshot covers both
evidence stages, whereas the amount above the ceiling is 1.842M.
Validation and test cost 32.506M and 9.209M, respectively, across the
three-cutoff measurement campaign. All of these charges are retained.
These are individual-seed measurements.
For FLASH* and EEIPU, saved traces establish the training observations, selected paths, phase
boundaries and downstream selection. The complete acquisition rankings and
random-draw transcript were unsaved, leaving numerical argmax reconstruction
outside the audit evidence.

\subsection{The RAG execution contract}
The main grammar performs one retrieval round and chooses a query strategy,
retriever, evidence-selection rule and retrieval width. Query choices include the current query,
pseudo-document-only retrieval, Query2doc with one or three copies of the
current query followed by its pseudo-document, two-query decomposition,
a concise rewrite, two synonymous rewrites, and HyDE. Synonymous rewrites
preserve one information need; decomposition splits it into subquestions.
Multi-round retrieval is deferred to a separate lower-priority ablation.

Ordinary queries have 30 retriever options: BM25, E5, Qwen embedding,
the original four-way RRF, five BGE-M3 dense/sparse mixtures with dense weights
$0,0.25,0.5,0.75,1$, 15 RRF source-set/constant combinations,
and six weighted fusions.
The five RRF source sets are BM25+E5, BM25+BGE, BM25+Qwen,
E5+BGE, and BM25+E5+BGE+Qwen; each uses $k\in\{10,60,100\}$.
BGE denotes its dense representation in these five sets.
Weighted fusion combines BM25 with E5 or Qwen. Each source's scores are
min--max normalized over the candidate documents; constant scores map to zero.
The BM25 weight is $\alpha\in\{0.25,0.5,0.75\}$ and the dense weight is $1-\alpha$.
HyDE instead selects one of three passage encoders: E5, BGE dense or Qwen.
Generated passages guide retrieval and never enter the evidence pool as sources.

Evidence selection uses relevance or MMR with relevance weight
$0.3,0.6,0.9$; retrieval width is $2,4,6,10$, capped by the candidate count.
The workflow then chooses identity or Qwen reranking and selects the first
$2,4,6$ available documents, capped by the evidence count. It preserves their
full source text and applies one CPU-only ordering: ranked, reversed, or ends.
The ends order alternates relevance ranks between the front and back, giving
$1,3,5,6,4,2$ for six documents. All three orders preserve the selected set.
For two documents, ends and ranked coincide. Ordering performs no model call;
the selected ordered context is passed to one final answer-generation stage.
There is no separate model-based order evaluator. Format retries follow the
same charged retry contract as every other method.

The retrieval block has $A_0=(7\cdot30+3)\cdot4\cdot4=3408$ choices.
Two rerankers, three evidence widths and three orders give
$18A_0=61{,}344$ syntactic terminal workflows. The final answer action is
uniquely determined by the selected width and adds no choice factor.
Input-specific coincidences and width truncation can make different paths
execute the same behavior. The count therefore describes syntactic paths.
The finite-tree theorem concerns the declared terminal grammar.

The first retrieval retains the grounded evidence assessment used in the
loop-value diagnostic: Qwen selects document and sentence identifiers, and the
implementation recovers exact source text. Its costs and retries are recorded.
Identity reranking prioritizes cited documents within the retrieved evidence;
the alternative reranker uses the same Qwen model. Neither controller receives
gold answers or supporting-document labels. WTB checkpoints preserve the selected
documents, their full text, the chosen order and the answer state. Candidates
with different orders share the prefix through evidence selection and execute
their own answer suffix. Compression is excluded from the main grammar.

\subsection{Paper-grounded reference workflows}
\label{app:paper-workflow-references}
We compare searched configurations with three fixed published pipelines,
then examine held-out quality, selected components and physical charges.
Each cited pipeline paper already combines at least two of the
mechanisms represented in our grammar: Query2doc combines generated query
expansion with lexical retrieval; HyDE combines hypothetical-document
generation with dense retrieval; RankGPT combines candidate retrieval with
listwise language-model reranking.

All references use the same candidate documents, grounded evidence
assessment, one retrieval round, retrieval width six, four retained documents,
ranked context and Qwen3.5-4B answer generation. Unspecified slots use the
original query, BM25, relevance selection and identity reranking. The common
identity reranker prioritizes documents cited by the grounded assessment,
as in the search grammar. These are shared-space implementations of the three
published retrieval pipelines, evaluated as complete QA workflows.

\begin{table}[t]
\centering\small
\caption{Three fixed references, each based on a published combination of
implemented mechanisms. All are evaluated on the same 300 primary test
questions without training or validation selection. Test F1 is in percent;
each recipe is a single fixed-workflow measurement.}
\label{tab:paper-workflow-recipes}
\begin{tabularx}{\linewidth}{@{}lp{0.26\linewidth}Xr@{}}
\toprule
ID & Published pipeline & Fixed component choices & Test F1\\
\midrule
P01 & Query2doc--BM25~\citep{wang2023query2doc} & \path{query.query2doc:q=3} then \path{retriever.bm25}. & 61.60\\
P02 & HyDE--E5~\citep{gao2023hyde} & \path{query.hyde} then \path{retriever.hyde_e5}. & 62.89\\
P03 & RankGPT-style BM25--Qwen~\citep{sun2023rankgpt} & \path{retriever.bm25} then \path{reranker.qwen3}. & 62.53\\
\bottomrule
\end{tabularx}
\end{table}

The three reference methods reach 61.60--62.89\% test F1, below the
69.57\% mean for AgentUCT's selected workflows at 32M. The selected workflows
improve on these specified recipes; the matched MCTS-UCT comparison isolates the
allocation change.
Their complete test-execution bills are 3.338M, 2.971M and 5.113M weighted
tokens, respectively, including one excluded sentinel per recipe.
Previously paid input-index preparation is referenced separately. These
measurements supply no across-run SD. The completed four-contrast analysis
includes these three fixed recipes and the mean of all fifty random workflows.

Query2doc uses our Qwen JSON prompt and three query copies, compared with
the original four-shot prompt, text-davinci-003 and five copies.
HyDE uses one Qwen-generated passage with E5 document-side encoding,
compared with the original InstructGPT and Contriever/mContriever setup.
The implementation uses one generated-document embedding with the trained E5
encoder.
P03 uses a single listwise window of at most six documents with Qwen3.5-4B,
compared with RankGPT's BM25 top-100 candidates and GPT sliding-window
protocol. Up to two format attempts are allowed, all executed attempts are
charged, and two invalid orders retain the common fusion order with a
recorded fallback. No retriever or reranker is retrained for this comparison.

The exact nine-action paths are frozen in
\path{configs/rag-followup-v2/fixed-test-manifest.json}. An independent
native-grammar check confirms all three are legal terminal workflows.

Fifty further complete workflows were sampled uniformly without replacement
from all 61,344 legal terminals, using seed 20261004 before their test
execution. Every sampled workflow was evaluated. Mean test F1 was
$60.58\pm5.22\%$ across workflows, the median was $62.11\%$, and the
interquartile range was $57.19$--$65.20\%$. The full observed range was
$48.43$--$68.39\%$; no draw was selected by test quality. Mean EM was
$46.39\pm4.88\%$. This no-search random-composition reference is distinct
from MCTS-Random allocation and from uniform-terminal search, which pays
for training evaluations and uses Top5--validation selection.
The three paper references and fifty random workflows require 15,900
primary question--workflow executions before reuse, plus separate sentinels.
All 15,900 primary outcomes and 53 sentinel outcomes completed. The fifty
random workflows incurred 164.392M weighted-token units of test execution;
the three fixed recipes incurred 11.422M. The combined 175.815M bill
contains 143,630 settled physical calls, with no failed calls or new input
preparation. This is the cost of measuring the reference distribution,
separate from an optimizer's search and validation bill.

\begin{figure}[t]
\centering
\includegraphics[width=3.3in]{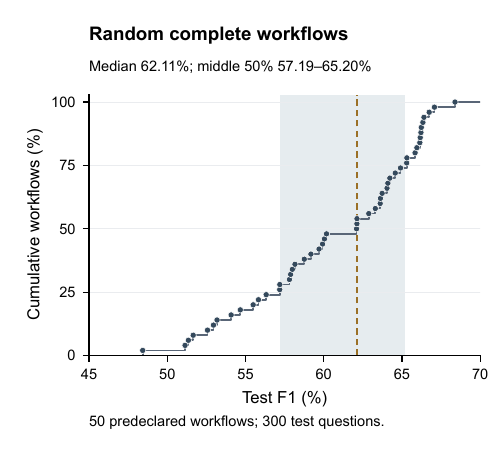}
\caption{Test F1 of all fifty prespecified random complete workflows. Each
point is one workflow on the same 300 primary questions. The gray band marks
the interquartile range and the dashed line the median; neither is a
confidence interval. The distribution describes unoptimized composition;
no draw is promoted to a selected baseline.}
\label{fig:random-composition-distribution}
\Description{All fifty sampled complete workflows are shown on a common test F1 axis, retaining the full observed distribution.}
\end{figure}

Strong-model planning was specified for native \path{deepseek-v4-pro} and
OpenRouter \path{openai/gpt-5.5}, with five plans each and high reasoning
effort~\citep{deepseek2026v4pro,openai2026gpt55}. Both receive identical task
instructions, grammar, component descriptions and three frozen training
examples, with no test inputs or quality feedback. The first generation
requests returned insufficient balance and regional unavailability,
respectively. These provider failures left the deferred references
without a workflow-quality measurement.

The active paired-test family contains four contrasts: the three fixed
recipes and the mean across all fifty random workflows, each against
AgentUCT's mean per-question F1 vector across its five seeds at 32M.
Holm adjustment covers this family.
The question is the paired unit; search-seed and random-workflow variation
are displayed separately. Full design details appear in
Section~\ref{sec:workflow-references}.

\begin{table}[t]
\centering\small\setlength{\tabcolsep}{3pt}
\caption{Question-paired reference comparisons at the fixed 32M endpoint.
$\Delta$ is AgentUCT's five-seed mean minus the reference. Intervals are
pointwise 95\% question-bootstrap intervals conditional on the observed
workflows; Holm adjusts only the four two-sided sign-randomization $p$ values.}
\label{tab:reference-paired-effects}
\begin{tabular}{@{}lrrrr@{}}
\toprule
Reference & F1 (\%) & $\Delta$ (pp) & 95\% CI (pp) & Holm $p$\\
\midrule
P01: Query2doc & 61.60 & +7.97 & [4.86, 11.25] & $6.0\!\times\!10^{-5}$\\
P02: HyDE & 62.89 & +6.68 & [3.59, 9.94] & $6.0\!\times\!10^{-5}$\\
P03: RankGPT-style & 62.53 & +7.04 & [3.72, 10.52] & $6.0\!\times\!10^{-5}$\\
Random mean (50) & 60.58 & +8.98 & [7.17, 10.93] & $4.0\!\times\!10^{-5}$\\
\bottomrule
\end{tabular}

\end{table}

The mean differences are 7.97, 6.68 and 7.04 percentage points for the
three fixed pipelines, and 8.98 points for random composition
(Table~\ref{tab:reference-paired-effects}). The random-composition contrast
has a conditional interval of [7.17, 10.93] points. Resampling random
workflows alone gives [7.52, 10.45] points; resampling both random
workflows and questions gives a sensitivity interval of [6.72, 11.53] points,
with the five AgentUCT seeds still fixed. The AgentUCT seed means range
from 69.20\% to 69.97\%, with sample SD 0.28 points; this search-seed
variation is separate from question uncertainty and the 5.22-point spread
across random workflows.

The paired bootstrap used 10,000 draws with seed 20261002. The two-sided
sign randomization used 100,000 draws with seed 20261005 and the
plus-one correction; the random-mean contrast had no sampled value as
extreme as the observed statistic, giving unadjusted $p=1/100001$, not zero.
Workflow-only and crossed workflow--question sensitivities used 10,000
draws each, with seeds 20261006 and 20261007. All resampling moved complete
question vectors together. These numerical choices were recorded before
resampling, after aggregate outcomes had been inspected. The comparisons
are retrospective on the existing panel and require the stated
question-level sign-exchangeability null. Their estimand is the existing
question panel and the observed workflows and search seeds.

\subsection{Allocation and controls}
Search and validation contain 150 and 300 disjoint hard questions from the
official HotpotQA training split. The primary test contains 300 hard questions
from the official validation split. Training and validation use sampling seed
20261003, excluding earlier search, validation, development and calibration
questions; the primary test remains unchanged. Earlier mixed-difficulty
searches are historical diagnostics outside this cohort.
Each method and seed owns one tree over all 150 training questions. A complete
workflow evaluation returns mean answer F1 over these 150 questions and backs
up that single return to the expanded node and its ancestors. The tree retains
its statistics, random state and qid--prefix cache between evaluations.
At each retained actual-cost snapshot, completed distinct workflows are ranked
by training mean F1, with first evaluation breaking ties. The top five are
evaluated on 300 validation questions. Highest validation mean F1 selects the
endpoint, with training rank breaking ties; 300 held-out questions provide the
primary test score. Validation and test scores never enter reward backup.
An unchanged workflow on an unchanged held-out panel is measured once and
referenced by any repeated measurement request.

The core comparison contains MCTS-Random, MCTS-UCT, AgentUCT with
Policy-complete costs, and AgentUCT-NextC. MCTS-UCT supplies the
standard UCB-guided Monte Carlo tree search baseline. MCTS-Random retains the same
tree and raw-return backup but selects children uniformly, independently of
their reward statistics. This local rule induces the tree's terminal distribution. Expansion and
rollout remain uniform. Five paired seeds (11, 23, 37, 53 and 71) define twenty
completed trials. The first eight trials were audited before continuing the
remaining seeds with unchanged settings; all twenty now have saved-trajectory,
accounting and endpoint-selection checks.
New seeds retain the same primary questions and assess search-seed variation.
If their primary scores are used to change the scientific settings, subsequent
results on that panel are exploratory. Independent confirmation of such a
change requires unused hard test questions fixed before execution.
Each method has its own cache and tree, with the same initial state,
controller, model, resource accounting and validation protocol. The
demand-profile theorem also covers selective panels; this fixed full-panel
experiment instantiates its compression case.

AgentUCT-NextC estimates the first positive unpaid execution per query.
Policy-complete prices the complete reward using 16 legal completions sampled
from smoothed historical prefix transitions. They use the same samples at a
common state, isolating the omitted suffix, and share the physical-price
learning rule and current cache observations. The uniform-complete estimator
control is reported as uniform continuation in
Section~\ref{sec:single-factor-mechanism}. Additional allocation-stage
variants and the no-cache GPU arm are deferred. A comparison
of incidence and residual prefix totals can use saved states without another
model run as a separate offline analysis.

Price priors come from independent fixed-path calibration on questions excluded
from search and evaluation. Each completed training evaluation updates physical
cost means and prefix-transition counts. Cache hits produce no cold-price sample.
The model uses qid--prefix, prefix and component means in that order, with a
calibration pseudo-observation. The historical completion law approximates
subsequent demand under a historical reference law, separately from the
current UCT policy's continuation law.
Evidence selection and CPU ordering have zero model-token price. Width-conditioned
answer prices belong to the final answer edge.

\paragraph{Numerical price bounds.}
For a query--prefix resource $r=(q,p)$ ending in action $a$, the RAG
predictor uses $M_r=M_a$ from Table~\ref{tab:rag-price-caps}.
The same table is frozen across seeds and used by AgentUCT-StaticFull.
Equation~(8) clips observations used to update predicted prices;
the physical ledger retains each actual charge.

\begin{table*}[t]
\centering\small
\caption{Configured RAG price bounds per query--prefix resource in
model-weighted token units. Actions with the same bound are grouped;
values are rounded to six decimals. The frozen configurations retain full precision.}
\label{tab:rag-price-caps}
\begin{tabularx}{\textwidth}{@{}Xr@{}}
\toprule
Action family & $M_a$\\
\midrule
Zero-cost identity, BM25, evidence selection, stopping, context ordering and width choice & 0.000000\\
HyDE E5 retrieval & 56.054907\\
E5 retrieval and BM25--E5 mixtures & 112.109814\\
BGE variants, HyDE BGE retrieval and BM25--BGE mixtures & 291.228353\\
E5--BGE reciprocal-rank fusion & 403.338167\\
HyDE Qwen embedding retrieval & 4880.601186\\
Qwen embedding retrieval and BM25--Qwen mixtures & 9761.202373\\
E5--BGE--Qwen reciprocal-rank fusion, including its legacy alias & 10164.540539\\
Retrieve-size edges and Qwen reranking & 86133.786542\\
Nonidentity query generation and final answer generation & 129200.679813\\
\bottomrule
\end{tabularx}
\end{table*}

The global scale is 150 times the median calibrated one-query completion bill;
tiny retains this same value. Token receipts are weighted by model parameter
count in billion-parameter-token units. Batch charges use logical request tokens,
including retries and excluding padding; padded slots and physical batch sizes
are recorded separately. The budget is a resource proxy, with no equivalence
to a fixed GPU-time or FLOP budget. Model-weighted token bills are the reported resource measure. Estimator elapsed time is instrumented separately; no timing comparison is reported here. Index preparation, search, validation and
held-out measurement have separate receipts.

The frozen protocol specifies five seeds and a 32M maximum weighted-token
search budget, with 16M and 24M reported from the completed trajectory's
affordable snapshots. It fixes the model context/output limits and an
evaluation-count safety cap. Each tiny
configuration changes only its dataset path: it loads three training queries,
one validation query and one development sentinel. Formal measurement excludes
that sentinel from the 300-query primary held-out score. Test F1 and EM,
physical costs and distinct syntactic workflows accompany
the common-budget comparison. Means and sample SDs describe variation across
the five search seeds on the fixed test panel, with the search seed as
the replication unit.

Validation-selected primary-test F1 at a fixed search budget is the primary
outcome. The search backs up training F1, uses cost-aware UCT for allocation,
and selects the final workflow by validation F1 alone. As an auxiliary diagnostic,
we report endpoint utility $J=F-\lambda_J C_{\mathrm{opt}}/\cref$ using
primary held-out F1, the existing reporting weight $\lambda_J=0.05$, and the
fixed calibration scale $\cref=1{,}470{,}690.0198555747$. The full physical
$C_{\mathrm{opt}}$ includes search, its complete crossing evaluation, and
validation, including failed attempts and retries. Test measurement and common
index preparation have separate bills and are excluded from this declared
optimization utility.

A secondary snapshot utility substitutes the retained
search bill and is labeled separately. The full-bill utility and physical
ledger retain the crossing work. Negative $J$ values are possible under
this fixed scale. The reporting weight and scale retain their declared
fixed values, independent of test outcomes. We retain the raw definition and
display same-seed $\Delta J$ against MCTS-UCT in the main five-seed table.
Search uses its declared allocation index and endpoint selection uses
validation quality. Joint improvements in $J$ and F1 remain empirical questions.

Utility comparisons use one full optimization bill per seed, with the seed
as the cost-observation unit. We decompose each difference into
the F1 difference and phase-specific cost contributions, further separating
retained search from complete evaluations after the snapshot cutoff. The latter
is the whole evaluation bill, including the portion within the reporting budget.
A validation-cost
saving contributes to end-to-end utility. The table reports all three baseline comparisons. An unknown physical charge prevents reporting a
complete utility value.
This comparison concerns allocation among budgeted search methods. Comparing
a one-off search investment with a fixed workflow at zero optimization cost
requires an additional economic comparison.

Table~\ref{tab:all-seeds} preserves every primary score and full-bill utility.
All 214 saved adaptive selections occur at the root, so the observed
allocation changes are root-level. Native-checkpoint and reuse checks cover the explicitly sampled trajectories.

\begin{table*}[t]
\centering\small
\caption{All twenty 32M runs. Each entry is primary test F1 (\%) / full-bill $J$.}
\label{tab:all-seeds}
\begin{tabular}{rllll}
\toprule
Seed & AgentUCT & MCTS-UCT & AgentUCT-NextC & MCTS-Random\\
\midrule
11 & 69.5109 / $-1.251646$ & 69.4443 / $-1.242602$ & 69.5109 / $-1.284709$ & 69.3527 / $-1.189892$\\
23 & 69.5228 / $-1.262981$ & 67.3162 / $-1.429808$ & 69.5228 / $-1.262340$ & 67.6926 / $-1.432710$\\
37 & 69.1975 / $-1.164197$ & 68.6871 / $-1.157955$ & 64.8892 / $-1.200446$ & 69.4880 / $-1.165447$\\
53 & 69.9657 / $-1.415753$ & 66.5768 / $-1.434304$ & 69.1490 / $-1.274501$ & 69.4550 / $-1.412869$\\
71 & 69.6305 / $-1.456986$ & 68.0144 / $-1.470387$ & 68.0144 / $-1.480661$ & 68.0144 / $-1.474790$\\
\bottomrule
\end{tabular}
\end{table*}

The five-seed AgentUCT-minus-baseline mean $\Delta J$ is
$+0.03670$ against MCTS-UCT, $-0.00978$ against AgentUCT-NextC and
$+0.02483$ against MCTS-Random. Table~\ref{tab:all-seeds} gives the five underlying
utility pairs for each comparison.
For the random comparison, mean contributions are $+0.00765$ from F1,
$-0.00302$ from retained search, $+0.01273$ from complete crossing evaluations,
and $+0.00747$ from validation. Against MCTS-UCT, the F1, retained-search,
complete-crossing and validation contributions are $+0.01558$, $+0.00084$,
$+0.01450$ and $+0.00578$, respectively. Against AgentUCT-NextC, F1 contributes
$+0.01348$, full search $+0.00294$ and validation $-0.02620$.

Search runs without cost-based admission checks. Every evaluation finishes and
its complete physical bill is recorded. For each reporting budget $B$, retain
the last complete snapshot with cumulative search cost at most $B$. Its tree
statistics, cache membership and shortlist use only that trace prefix.
The evaluation crossing the largest reporting budget also finishes; its work
and overshoot remain in the raw trajectory, and its reward cannot enter an
earlier snapshot. Validation and test measurement use each retained shortlist
and have separately reported costs. Index preparation is also reported
separately. Selection and model calls follow the fixed runtime throughout
the reporting budgets. The finite-tree theorem applies at the retained evaluation
count $n(B)$; a physical-budget rate $O(\log B)$ remains unproved.

\subsection{Multi-seed summaries and reporting units}
All budget summaries use the same five search seeds, with one validation-selected
primary outcome per method, seed and budget. Tables report arithmetic means and
sample standard deviations, $s=\sqrt{\sum_{i=1}^{5}(x_i-\bar x)^2/4}$.
The uncertainty bars in the budget and resource figures show this between-seed
spread. These sample-SD bars summarize seed variation. A reused endpoint
keeps its original complete test measurement and the same sample size.

For each budget and baseline, we subtract its primary score from AgentUCT's
score for the same seed. Tables report the mean and sample SD of these five
differences, together with wins, ties and losses: the numbers of positive,
zero and negative saved differences. F1 is primary and EM auxiliary.
These summaries describe the observed seed variation on the fixed question
panel. Detailed reporting was added to the frozen runs without changing
allocation or endpoint-selection parameters.

\subsection{Budget-dependent held-out quality}
\label{sec:budget-detail}
All three cutoffs use the same five search seeds and the complete
Top5--validation--primary-test procedure. Table~\ref{tab:budget-quality-full}
reports all method--budget means and sample SDs; the individual seed tables
below preserve the complete sixty outcomes. Early snapshots were read from
frozen completed search prefixes. Only missing validation panels and newly
selected primary endpoints were executed, in separate measurement caches.
No later rewards, prices or execution-cache entries were imported into search.
Across both early-budget campaigns, 242,293 new physical calls incurred
375.606M weighted-token units. These are additional measurement costs, not
observed historical early-stopping optimization bills. We therefore do not
construct 16M or 24M values of $J$.

\begin{table}[htbp]
\centering\small
\caption{Complete matched-five-seed budget analysis. Primary test F1 and EM
are reported as mean $\pm$ sample SD. Each method--budget row has five seeds;
reused test measurements across budgets are not additional independent trials.}
\label{tab:budget-quality-full}
\begin{tabular}{rlcc}
\toprule
Budget (M) & Method & F1 (\%) & EM (\%) \\
\midrule
16 & AgentUCT & $67.40 \pm 2.54$ & $52.33 \pm 2.22$ \\
16 & MCTS-UCT & $67.38 \pm 2.52$ & $52.27 \pm 2.14$ \\
16 & AgentUCT-NextC & $67.40 \pm 2.54$ & $52.33 \pm 2.22$ \\
16 & MCTS-Random & $67.40 \pm 2.54$ & $52.33 \pm 2.22$ \\
24 & AgentUCT & $69.57 \pm 0.28$ & $54.60 \pm 0.15$ \\
24 & MCTS-UCT & $68.01 \pm 1.12$ & $52.47 \pm 1.68$ \\
24 & AgentUCT-NextC & $67.83 \pm 2.69$ & $53.33 \pm 2.44$ \\
24 & MCTS-Random & $68.44 \pm 0.89$ & $53.33 \pm 1.49$ \\
32 & AgentUCT & $69.57 \pm 0.28$ & $54.60 \pm 0.15$ \\
32 & MCTS-UCT & $68.01 \pm 1.12$ & $52.47 \pm 1.68$ \\
32 & AgentUCT-NextC & $68.22 \pm 1.96$ & $53.07 \pm 1.64$ \\
32 & MCTS-Random & $68.80 \pm 0.87$ & $53.73 \pm 1.64$ \\
\bottomrule
\end{tabular}

\end{table}

\begin{figure*}[t]
\centering
\includegraphics[width=\textwidth]{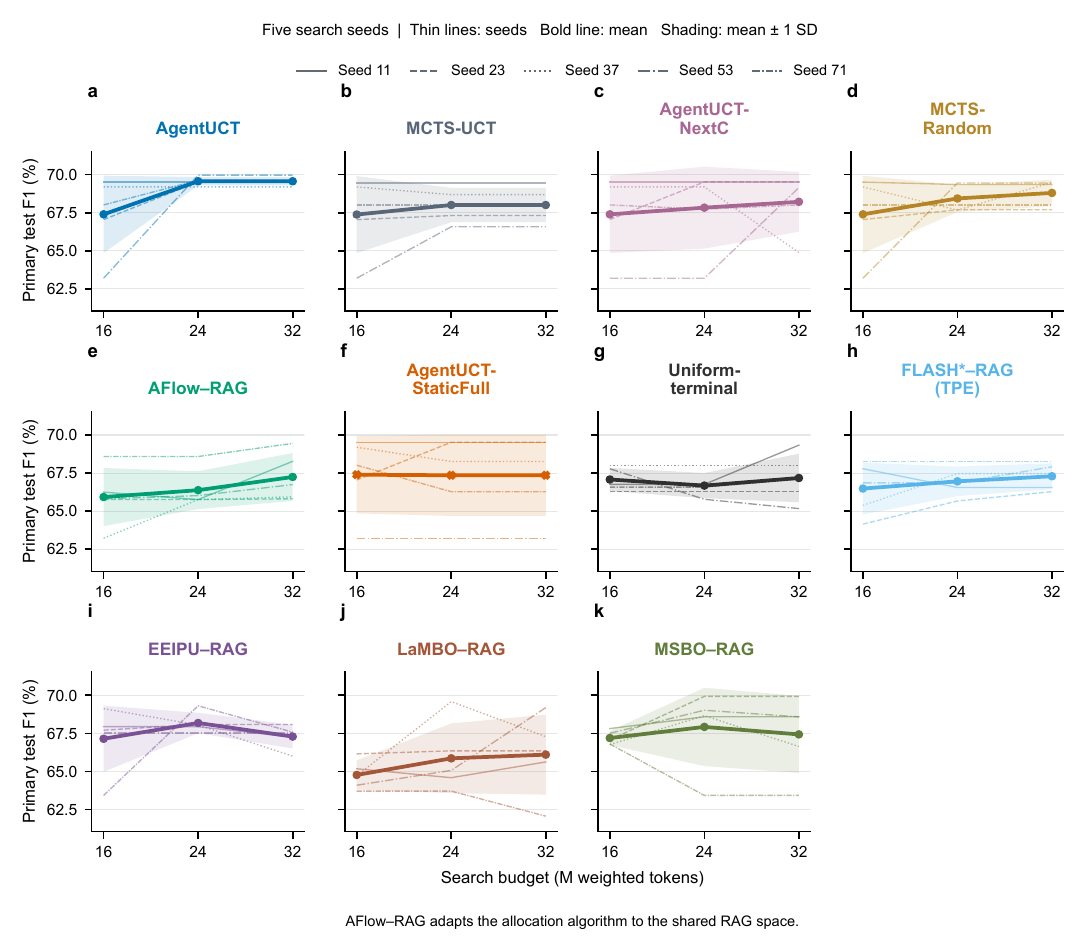}
\caption{Individual seed trajectories behind the main budget comparison.
Each method panel shows the same five seeds as thin lines and their mean as
a bold line; shading is $\pm$ one sample SD. Unchanged endpoints within a
trial share a complete test measurement across budgets.}
\label{fig:budget-quality-seeds}
\end{figure*}

The eight initial root evaluations cost 12.287--16.626M weighted tokens across
runs. At 16M, only seed 23 retained two completed adaptive evaluations;
the other four seeds retained none, and seed 53 retained only seven
initialization evaluations.
The four methods selected the same within-seed endpoint at 16M.
AgentUCT, AgentUCT-NextC and MCTS-Random had identical test F1; MCTS-UCT's mean differed
by 0.0133 points because seed 11 used a fresh execution of the same workflow.
At 24M, mean F1 was 69.5655\%, 68.0077\%, 67.8342\% and 68.4382\% for
AgentUCT, MCTS-UCT, AgentUCT-NextC and MCTS-Random, respectively. Their 32M values were
69.5655\%, 68.0077\%, 68.2173\% and 68.8006\%.

Table~\ref{tab:initialization-paired-evidence} locates the start of adaptive
allocation within the actual budget. For seeds 23, 37 and 53, the cost term
changes the saved AgentUCT action at its first adaptive decision; for seeds
11 and 71, the first change occurs at evaluation 11. These are comparisons
with the cost-free index in the same saved state, not reconstructed MCTS-UCT
trajectories. Four seeds select different endpoints from MCTS-UCT, with test-F1
gains of 0.51--3.39 points. Seed 11 selects the same endpoint, and its small
difference comes from separate executions. The table connects initialization,
allocation and endpoint changes without attributing an endpoint's F1 gain
to a particular changed action.

\begin{table*}[t]
\centering\small
\caption{Where adaptive allocation becomes observable. A and U denote
AgentUCT and MCTS-UCT. Costs are M weighted tokens. ``First A feedback'' is the
cumulative bill after AgentUCT's first complete adaptive evaluation;
``First changed choice'' is the bill before its first decision whose priced
argmax differs from the same-state cost-free argmax. $\Delta$F1 is AgentUCT
minus MCTS-UCT in percentage points at both 24M and 32M; these budgets reuse the
same endpoint measurement. Configuration IDs resolve in the workflow dictionary.}
\label{tab:initialization-paired-evidence}
\begin{tabular}{@{}rrrrrrll@{}}
\toprule
Seed & Init. A & Init. U & First A feedback & First changed choice & $\Delta$F1 & A config. & U config.\\
\midrule
11 & 13.454 & 13.446 & 16.015 & 17.455 & +0.07 & C17 & C17\\
23 & 12.287 & 12.287 & 13.407 & 12.287 & +2.21 & C01 & C16\\
37 & 13.969 & 13.969 & 16.313 & 13.969 & +0.51 & C10 & C06\\
53 & 16.626 & 16.626 & 17.362 & 16.626 & +3.39 & C15 & C02\\
71 & 13.043 & 13.043 & 16.731 & 20.445 & +1.62 & C19 & C05\\
\bottomrule
\end{tabular}

\end{table*}

The lack of monotonicity has a concrete selection explanation. For
AgentUCT-NextC seed 53, a HyDE workflow ranked fourth in the 24M training
shortlist but sixth at 32M, outside the fixed Top5. Its validation F1 of
73.0699\% exceeded the other eligible winner's 72.4927\%; its primary F1 was
63.2059\%, versus 69.1490\% for the 32M winner. Saved shortlist membership,
validation order, original predictions and official scores were checked.
AgentUCT-NextC seed 37 instead decreased from 69.1975\% at 24M to 64.8892\%
at 32M. A larger budget can change the shortlist and validation winner without
improving that winner's test performance. Neither these outcomes nor the flat
AgentUCT curve was used to tune $\beta$, choose a new stopping budget or
extend the search.

\subsection{AgentUCT versus AgentUCT-NextC}
\label{sec:nextc-detail}
This comparison retains the same completion sampler and price-learning rule
while changing which unpaid costs enter selection. All five paired 16M
endpoints coincide; three coincide at 24M and two at 32M.
The paired F1 difference is $1.73\pm2.93$ points at 24M and
$1.35\pm1.79$ points at 32M. These are means and sample SDs of paired
\emph{differences}, not SDs pooled across independent methods.
Table~\ref{tab:nextc-descriptive} reports this contrast together with EM
and the direction of each seed's difference.

At 32M, AgentUCT used 7.385M raw LLM tokens in its retained search versus
7.399M for AgentUCT-NextC, a 0.18\% difference. Their mean logical hit rates
were 6.79\% and 6.27\%, respectively, with the denominator defined below.
AgentUCT's full search bill was 0.086M weighted tokens lower, but validation
was 0.771M higher. Thus its full optimization bill was 0.684M higher, yielding
the negative mean $\Delta J$ despite higher primary F1. Higher endpoint F1 is
accompanied by higher total optimization expense in this comparison.

\begin{table}[htbp]
\centering\small
\caption{AgentUCT minus AgentUCT-NextC. Differences and their sample SDs
are in percentage points across five paired seeds. W/T/L counts positive,
zero and negative differences; the same rows appear in
Table~\ref{tab:budget-descriptive}.}
\label{tab:nextc-descriptive}
\begin{tabular}{rlrrrr}
\toprule
$B$ (M) & Contrast / metric & $\Delta$ (pp) & SD (pp) & W/T/L & $n$\\
\midrule
16 & NextC / F1 & +0.00 & 0.00 & 0/5/0 & 5\\
16 & NextC / EM & +0.00 & 0.00 & 0/5/0 & 5\\
24 & NextC / F1 & +1.73 & 2.93 & 2/3/0 & 5\\
24 & NextC / EM & +1.27 & 2.48 & 2/3/0 & 5\\
32 & NextC / F1 & +1.35 & 1.79 & 3/2/0 & 5\\
32 & NextC / EM & +1.53 & 1.71 & 3/2/0 & 5\\
\bottomrule
\end{tabular}

\end{table}
\begin{table}[htbp]
\centering\small\setlength{\tabcolsep}{3pt}
\caption{All paired budget comparisons, AgentUCT minus each baseline.
Differences and sample SDs are in percentage points over five seeds.
W/T/L gives the numbers of seed-level wins, ties and losses. F1 is primary
and EM auxiliary.}
\label{tab:budget-descriptive}
\begin{tabular}{rlrrrr}
\toprule
$B$ (M) & Contrast / metric & $\Delta$ (pp) & SD (pp) & W/T/L & $n$\\
\midrule
16 & MCTS-UCT / F1 & +0.01 & 0.03 & 1/4/0 & 5\\
16 & NextC / F1 & +0.00 & 0.00 & 0/5/0 & 5\\
16 & MCTS-Random / F1 & +0.00 & 0.00 & 0/5/0 & 5\\
16 & MCTS-UCT / EM & +0.07 & 0.15 & 1/4/0 & 5\\
16 & NextC / EM & +0.00 & 0.00 & 0/5/0 & 5\\
16 & MCTS-Random / EM & +0.00 & 0.00 & 0/5/0 & 5\\
24 & MCTS-UCT / F1 & +1.56 & 1.33 & 5/0/0 & 5\\
24 & NextC / F1 & +1.73 & 2.93 & 2/3/0 & 5\\
24 & MCTS-Random / F1 & +1.13 & 0.74 & 5/0/0 & 5\\
24 & MCTS-UCT / EM & +2.13 & 1.66 & 5/0/0 & 5\\
24 & NextC / EM & +1.27 & 2.48 & 2/3/0 & 5\\
24 & MCTS-Random / EM & +1.27 & 1.44 & 4/0/1 & 5\\
32 & MCTS-UCT / F1 & +1.56 & 1.33 & 5/0/0 & 5\\
32 & NextC / F1 & +1.35 & 1.79 & 3/2/0 & 5\\
32 & MCTS-Random / F1 & +0.76 & 0.92 & 4/0/1 & 5\\
32 & MCTS-UCT / EM & +2.13 & 1.66 & 5/0/0 & 5\\
32 & NextC / EM & +1.53 & 1.71 & 3/2/0 & 5\\
32 & MCTS-Random / EM & +0.87 & 1.57 & 3/0/2 & 5\\
\bottomrule
\end{tabular}

\end{table}

\begin{table}[t]
\centering\small
\setlength{\tabcolsep}{4pt}
\caption{All twenty validation-selected primary outcomes at the 16M search cutoff. F1 and EM are percentages.}
\label{tab:seed-values-b16000000}
\begin{tabular}{rlrrr}
\toprule
Budget (M) & Method & Seed & F1 (\%) & EM (\%) \\
\midrule
16 & AgentUCT & 11 & 69.5109 & 54.6667 \\
16 & AgentUCT & 23 & 67.0514 & 52.3333 \\
16 & AgentUCT & 37 & 69.1952 & 54.0000 \\
16 & AgentUCT & 53 & 63.2059 & 49.0000 \\
16 & AgentUCT & 71 & 68.0144 & 51.6667 \\
16 & MCTS-UCT & 11 & 69.4443 & 54.3333 \\
16 & MCTS-UCT & 23 & 67.0514 & 52.3333 \\
16 & MCTS-UCT & 37 & 69.1952 & 54.0000 \\
16 & MCTS-UCT & 53 & 63.2059 & 49.0000 \\
16 & MCTS-UCT & 71 & 68.0144 & 51.6667 \\
16 & AgentUCT-NextC & 11 & 69.5109 & 54.6667 \\
16 & AgentUCT-NextC & 23 & 67.0514 & 52.3333 \\
16 & AgentUCT-NextC & 37 & 69.1952 & 54.0000 \\
16 & AgentUCT-NextC & 53 & 63.2059 & 49.0000 \\
16 & AgentUCT-NextC & 71 & 68.0144 & 51.6667 \\
16 & MCTS-Random & 11 & 69.5109 & 54.6667 \\
16 & MCTS-Random & 23 & 67.0514 & 52.3333 \\
16 & MCTS-Random & 37 & 69.1952 & 54.0000 \\
16 & MCTS-Random & 53 & 63.2059 & 49.0000 \\
16 & MCTS-Random & 71 & 68.0144 & 51.6667 \\
\bottomrule
\end{tabular}
\end{table}

\begin{table}[t]
\centering\small
\setlength{\tabcolsep}{4pt}
\caption{All twenty validation-selected primary outcomes at the 24M search cutoff. F1 and EM are percentages.}
\label{tab:seed-values-b24000000}
\begin{tabular}{rlrrr}
\toprule
Budget (M) & Method & Seed & F1 (\%) & EM (\%) \\
\midrule
24 & AgentUCT & 11 & 69.5109 & 54.6667 \\
24 & AgentUCT & 23 & 69.5228 & 54.3333 \\
24 & AgentUCT & 37 & 69.1975 & 54.6667 \\
24 & AgentUCT & 53 & 69.9657 & 54.6667 \\
24 & AgentUCT & 71 & 69.6305 & 54.6667 \\
24 & MCTS-UCT & 11 & 69.4443 & 54.3333 \\
24 & MCTS-UCT & 23 & 67.3162 & 52.0000 \\
24 & MCTS-UCT & 37 & 68.6871 & 54.0000 \\
24 & MCTS-UCT & 53 & 66.5768 & 50.3333 \\
24 & MCTS-UCT & 71 & 68.0144 & 51.6667 \\
24 & AgentUCT-NextC & 11 & 69.5109 & 54.6667 \\
24 & AgentUCT-NextC & 23 & 69.5228 & 54.3333 \\
24 & AgentUCT-NextC & 37 & 69.1975 & 54.6667 \\
24 & AgentUCT-NextC & 53 & 63.2059 & 49.0000 \\
24 & AgentUCT-NextC & 71 & 67.7338 & 54.0000 \\
24 & MCTS-Random & 11 & 69.3527 & 55.3333 \\
24 & MCTS-Random & 23 & 67.6926 & 52.3333 \\
24 & MCTS-Random & 37 & 67.6761 & 53.0000 \\
24 & MCTS-Random & 53 & 69.4550 & 54.3333 \\
24 & MCTS-Random & 71 & 68.0144 & 51.6667 \\
\bottomrule
\end{tabular}
\end{table}

\begin{table}[t]
\centering\small
\setlength{\tabcolsep}{4pt}
\caption{All twenty validation-selected primary outcomes at the 32M search cutoff. F1 and EM are percentages.}
\label{tab:seed-values-b32000000}
\begin{tabular}{rlrrr}
\toprule
Budget (M) & Method & Seed & F1 (\%) & EM (\%) \\
\midrule
32 & AgentUCT & 11 & 69.5109 & 54.6667 \\
32 & AgentUCT & 23 & 69.5228 & 54.3333 \\
32 & AgentUCT & 37 & 69.1975 & 54.6667 \\
32 & AgentUCT & 53 & 69.9657 & 54.6667 \\
32 & AgentUCT & 71 & 69.6305 & 54.6667 \\
32 & MCTS-UCT & 11 & 69.4443 & 54.3333 \\
32 & MCTS-UCT & 23 & 67.3162 & 52.0000 \\
32 & MCTS-UCT & 37 & 68.6871 & 54.0000 \\
32 & MCTS-UCT & 53 & 66.5768 & 50.3333 \\
32 & MCTS-UCT & 71 & 68.0144 & 51.6667 \\
32 & AgentUCT-NextC & 11 & 69.5109 & 54.6667 \\
32 & AgentUCT-NextC & 23 & 69.5228 & 54.3333 \\
32 & AgentUCT-NextC & 37 & 64.8892 & 51.0000 \\
32 & AgentUCT-NextC & 53 & 69.1490 & 53.6667 \\
32 & AgentUCT-NextC & 71 & 68.0144 & 51.6667 \\
32 & MCTS-Random & 11 & 69.3527 & 55.3333 \\
32 & MCTS-Random & 23 & 67.6926 & 52.3333 \\
32 & MCTS-Random & 37 & 69.4880 & 55.0000 \\
32 & MCTS-Random & 53 & 69.4550 & 54.3333 \\
32 & MCTS-Random & 71 & 68.0144 & 51.6667 \\
\bottomrule
\end{tabular}
\end{table}

\subsection{Root allocation and selected configurations}
\label{sec:allocation-configurations}
All 214 adaptive selections in the full original trajectories occurred at
the root. The retained 32M prefixes contain 194 of these; the twenty crossing
evaluations account for the remainder. Tables~\ref{tab:root-selections}
and~\ref{tab:root-costs} separate root-allocation counts from complete evaluation
bills. Relative to MCTS-UCT, AgentUCT selected the HyDE root more often
($1.8\pm0.84$ versus $1.2\pm0.84$ times after initialization) and rewriting
less often ($0.8\pm0.45$ versus $1.4\pm0.55$). Relative to AgentUCT-NextC,
it also selected HyDE more often ($1.8$ versus $0.8$ times).
The cost attributed to a root is the entire evaluation beginning at that root,
including all later components, not the cost of query construction alone.
These saved allocations locate the observed policy differences. A causal
contribution of any query operator to test F1 requires a controlled intervention.

\begin{table}[htbp]
\centering\small
\caption{Adaptive root selections in retained 32M prefixes, mean $\pm$
sample SD over five seeds. Common initialization and the crossing evaluation
are excluded. Subsequent expansion and rollout remained uniform.}
\label{tab:root-selections}
\begin{tabular}{@{}lrrrr@{}}
\toprule
Root query action & AgentUCT & MCTS-UCT & AgentUCT-NextC & MCTS-Random \\
\midrule
decomposition & $1.20 \pm 0.84$ & $1.00 \pm 0.00$ & $1.00 \pm 0.00$ & $1.40 \pm 0.89$ \\
hyde & $1.80 \pm 0.84$ & $1.20 \pm 0.84$ & $0.80 \pm 0.45$ & $1.40 \pm 0.89$ \\
identity & $1.00 \pm 0.00$ & $1.40 \pm 0.55$ & $1.20 \pm 0.45$ & $0.80 \pm 0.45$ \\
query2doc & $1.20 \pm 0.45$ & $1.20 \pm 0.45$ & $1.60 \pm 0.55$ & $0.40 \pm 0.89$ \\
query2doc:q=1 & $1.20 \pm 0.45$ & $1.00 \pm 0.00$ & $1.60 \pm 0.55$ & $1.80 \pm 1.30$ \\
query2doc:q=3 & $1.40 \pm 0.55$ & $1.20 \pm 0.45$ & $1.00 \pm 0.00$ & $1.40 \pm 0.89$ \\
rewrite & $0.80 \pm 0.45$ & $1.40 \pm 0.55$ & $1.20 \pm 0.45$ & $1.20 \pm 1.64$ \\
synonyms & $1.00 \pm 0.00$ & $1.40 \pm 0.55$ & $1.20 \pm 0.45$ & $1.40 \pm 1.52$ \\
\bottomrule
\end{tabular}

\end{table}
\begin{table}[htbp]
\centering\small
\caption{Retained 32M cost by initiating root action, in M weighted tokens
(mean $\pm$ sample SD). Each entry includes the entire evaluations in that
group, including initialization. These are not component-specific query costs.}
\label{tab:root-costs}
\begin{tabular}{@{}lrrrr@{}}
\toprule
Root query action & AgentUCT & MCTS-UCT & AgentUCT-NextC & MCTS-Random \\
\midrule
decomposition & $3.33 \pm 1.71$ & $3.00 \pm 0.80$ & $3.20 \pm 0.73$ & $4.18 \pm 1.99$ \\
hyde & $4.75 \pm 1.13$ & $3.82 \pm 2.56$ & $3.06 \pm 1.63$ & $3.39 \pm 1.62$ \\
identity & $3.99 \pm 2.55$ & $5.87 \pm 2.86$ & $5.08 \pm 1.96$ & $3.63 \pm 1.51$ \\
query2doc & $4.44 \pm 2.26$ & $3.26 \pm 2.90$ & $5.02 \pm 2.47$ & $2.29 \pm 1.53$ \\
query2doc:q=1 & $3.95 \pm 1.16$ & $3.09 \pm 1.61$ & $5.14 \pm 1.49$ & $5.70 \pm 3.57$ \\
query2doc:q=3 & $5.34 \pm 1.63$ & $3.80 \pm 1.38$ & $3.99 \pm 1.44$ & $4.60 \pm 2.71$ \\
rewrite & $2.48 \pm 1.21$ & $4.00 \pm 1.89$ & $2.57 \pm 1.12$ & $3.38 \pm 2.26$ \\
synonyms & $2.85 \pm 0.72$ & $4.33 \pm 1.30$ & $3.13 \pm 1.37$ & $3.88 \pm 2.73$ \\
\bottomrule
\end{tabular}

\end{table}

We report ``best configurations'' as validation-selected endpoints, not the
best test scores among inspected workflows. The sixty selected cells contain
20 distinct configurations: five at 16M, fifteen at 24M and fifteen at 32M.
At 24M and 32M all five AgentUCT endpoints are unchanged, as are all five MCTS-UCT
endpoints; AgentUCT-NextC changes three and MCTS-Random one. Those stable cells
reuse their corresponding test measurements. Across all budgets, 37 stored
test measurements support the sixty reported outcomes. Identical workflows in
independent runs need not share a stored test measurement.

At 32M, AgentUCT's five selected workflows use five different query actions:
rewriting, decomposition, document generation, Query2doc with mixture weight
three, and synonyms. All five use the identity reranker; context widths are
two, four, four, six and two in seed order. The observed result spans
these five selected pipelines. Operator frequencies describe their composition;
isolating an operator's effect requires an ablation. The tables below list all sixty
validation winners, their validation/test scores, complete component sequences
and test-source identifiers. The dictionaries retain exact saved operator
parameters, and repeated source codes identify shared measurements.

\begin{table}[p]
\centering
\small
\setlength{\tabcolsep}{4pt}
\caption{Codes used in the complete workflow dictionaries. Each numbered code denotes the exact saved action; the code sequence preserves every component and its order.}
\label{tab:rag-config-code-definitions}
\begin{tabular}{@{}ll@{}}
\toprule
Prefix & Component \\
\midrule
Q & Query transformation \\
R & Retrieval backend and fusion \\
S & Evidence selection \\
K & Retrieved-document count \\
C & Retrieval continuation decision \\
RR & Reranking \\
M & Context-document count \\
X & Context ordering \\
A & Answer generation \\
\bottomrule
\end{tabular}
\end{table}

\begin{table}[p]
\centering
\small
\setlength{\tabcolsep}{4pt}
\caption{Validation-selected RAG configurations at 16M weighted-token search budget. Rank is the auxiliary position in the saved training Top5; validation chooses the endpoint and breaks ties by that rank. F1 and EM are percentages on the fixed 300-question panels; the test sentinel is excluded. Repeated T codes reference the same stored test measurement, not independent replications.}
\label{tab:rag-best-configs-16m}
\begin{tabular}{@{}rlrrrrrl@{}}
\toprule
Seed & Method & Cfg & Rank & Valid F1 & Test F1 & Test EM & Test source \\
\midrule
11 & AgentUCT & C17 & 1 & 74.36 & 69.51 & 54.67 & T01 \\
11 & MCTS-UCT & C17 & 1 & 74.63 & 69.44 & 54.33 & T30 \\
11 & AgentUCT-NextC & C17 & 1 & 74.36 & 69.51 & 54.67 & T10 \\
11 & MCTS-Random & C17 & 1 & 74.36 & 69.51 & 54.67 & T21 \\
23 & AgentUCT & C18 & 1 & 71.54 & 67.05 & 52.33 & T03 \\
23 & MCTS-UCT & C18 & 1 & 71.54 & 67.05 & 52.33 & T32 \\
23 & AgentUCT-NextC & C18 & 1 & 71.54 & 67.05 & 52.33 & T12 \\
23 & MCTS-Random & C18 & 1 & 71.54 & 67.05 & 52.33 & T23 \\
37 & AgentUCT & C12 & 1 & 70.46 & 69.20 & 54.00 & T05 \\
37 & MCTS-UCT & C12 & 1 & 70.46 & 69.20 & 54.00 & T34 \\
37 & AgentUCT-NextC & C12 & 1 & 70.46 & 69.20 & 54.00 & T15 \\
37 & MCTS-Random & C12 & 1 & 70.46 & 69.20 & 54.00 & T26 \\
53 & AgentUCT & C03 & 3 & 73.07 & 63.21 & 49.00 & T07 \\
53 & MCTS-UCT & C03 & 3 & 73.07 & 63.21 & 49.00 & T36 \\
53 & AgentUCT-NextC & C03 & 3 & 73.07 & 63.21 & 49.00 & T17 \\
53 & MCTS-Random & C03 & 3 & 73.07 & 63.21 & 49.00 & T28 \\
71 & AgentUCT & C05 & 1 & 71.70 & 68.01 & 51.67 & T09 \\
71 & MCTS-UCT & C05 & 1 & 71.70 & 68.01 & 51.67 & T37 \\
71 & AgentUCT-NextC & C05 & 1 & 71.70 & 68.01 & 51.67 & T18 \\
71 & MCTS-Random & C05 & 1 & 71.70 & 68.01 & 51.67 & T29 \\
\bottomrule
\end{tabular}
\end{table}

\begin{table}[p]
\centering
\small
\setlength{\tabcolsep}{4pt}
\caption{Validation-selected RAG configurations at 24M weighted-token search budget. Rank is the auxiliary position in the saved training Top5; validation chooses the endpoint and breaks ties by that rank. F1 and EM are percentages on the fixed 300-question panels; the test sentinel is excluded. Repeated T codes reference the same stored test measurement, not independent replications.}
\label{tab:rag-best-configs-24m}
\begin{tabular}{@{}rlrrrrrl@{}}
\toprule
Seed & Method & Cfg & Rank & Valid F1 & Test F1 & Test EM & Test source \\
\midrule
11 & AgentUCT & C17 & 1 & 74.36 & 69.51 & 54.67 & T01 \\
11 & MCTS-UCT & C17 & 2 & 74.63 & 69.44 & 54.33 & T30 \\
11 & AgentUCT-NextC & C17 & 3 & 74.36 & 69.51 & 54.67 & T10 \\
11 & MCTS-Random & C08 & 2 & 75.05 & 69.35 & 55.33 & T20 \\
23 & AgentUCT & C01 & 1 & 74.09 & 69.52 & 54.33 & T02 \\
23 & MCTS-UCT & C16 & 1 & 74.01 & 67.32 & 52.00 & T31 \\
23 & AgentUCT-NextC & C01 & 1 & 74.09 & 69.52 & 54.33 & T11 \\
23 & MCTS-Random & C14 & 1 & 72.36 & 67.69 & 52.33 & T22 \\
37 & AgentUCT & C10 & 1 & 71.06 & 69.20 & 54.67 & T04 \\
37 & MCTS-UCT & C06 & 1 & 73.65 & 68.69 & 54.00 & T33 \\
37 & AgentUCT-NextC & C10 & 1 & 71.06 & 69.20 & 54.67 & T14 \\
37 & MCTS-Random & C13 & 1 & 72.04 & 67.68 & 53.00 & T25 \\
53 & AgentUCT & C15 & 1 & 74.45 & 69.97 & 54.67 & T06 \\
53 & MCTS-UCT & C02 & 1 & 74.00 & 66.58 & 50.33 & T35 \\
53 & AgentUCT-NextC & C03 & 4 & 73.07 & 63.21 & 49.00 & T17 \\
53 & MCTS-Random & C07 & 1 & 73.81 & 69.46 & 54.33 & T27 \\
71 & AgentUCT & C19 & 3 & 74.60 & 69.63 & 54.67 & T08 \\
71 & MCTS-UCT & C05 & 1 & 71.70 & 68.01 & 51.67 & T37 \\
71 & AgentUCT-NextC & C04 & 4 & 71.94 & 67.73 & 54.00 & T19 \\
71 & MCTS-Random & C05 & 1 & 71.70 & 68.01 & 51.67 & T29 \\
\bottomrule
\end{tabular}
\end{table}

\begin{table}[p]
\centering
\small
\setlength{\tabcolsep}{4pt}
\caption{Validation-selected RAG configurations at 32M weighted-token search budget. Rank is the auxiliary position in the saved training Top5; validation chooses the endpoint and breaks ties by that rank. F1 and EM are percentages on the fixed 300-question panels; the test sentinel is excluded. Repeated T codes reference the same stored test measurement, not independent replications.}
\label{tab:rag-best-configs-32m}
\begin{tabular}{@{}rlrrrrrl@{}}
\toprule
Seed & Method & Cfg & Rank & Valid F1 & Test F1 & Test EM & Test source \\
\midrule
11 & AgentUCT & C17 & 1 & 74.36 & 69.51 & 54.67 & T01 \\
11 & MCTS-UCT & C17 & 2 & 74.63 & 69.44 & 54.33 & T30 \\
11 & AgentUCT-NextC & C17 & 4 & 74.36 & 69.51 & 54.67 & T10 \\
11 & MCTS-Random & C08 & 2 & 75.05 & 69.35 & 55.33 & T20 \\
23 & AgentUCT & C01 & 1 & 74.09 & 69.52 & 54.33 & T02 \\
23 & MCTS-UCT & C16 & 1 & 74.01 & 67.32 & 52.00 & T31 \\
23 & AgentUCT-NextC & C01 & 1 & 74.09 & 69.52 & 54.33 & T11 \\
23 & MCTS-Random & C14 & 1 & 72.36 & 67.69 & 52.33 & T22 \\
37 & AgentUCT & C10 & 1 & 71.06 & 69.20 & 54.67 & T04 \\
37 & MCTS-UCT & C06 & 2 & 73.65 & 68.69 & 54.00 & T33 \\
37 & AgentUCT-NextC & C09 & 2 & 71.50 & 64.89 & 51.00 & T13 \\
37 & MCTS-Random & C20 & 1 & 72.61 & 69.49 & 55.00 & T24 \\
53 & AgentUCT & C15 & 1 & 74.45 & 69.97 & 54.67 & T06 \\
53 & MCTS-UCT & C02 & 1 & 74.00 & 66.58 & 50.33 & T35 \\
53 & AgentUCT-NextC & C11 & 1 & 72.49 & 69.15 & 53.67 & T16 \\
53 & MCTS-Random & C07 & 1 & 73.81 & 69.46 & 54.33 & T27 \\
71 & AgentUCT & C19 & 3 & 74.60 & 69.63 & 54.67 & T08 \\
71 & MCTS-UCT & C05 & 1 & 71.70 & 68.01 & 51.67 & T37 \\
71 & AgentUCT-NextC & C05 & 1 & 71.70 & 68.01 & 51.67 & T18 \\
71 & MCTS-Random & C05 & 1 & 71.70 & 68.01 & 51.67 & T29 \\
\bottomrule
\end{tabular}
\end{table}

\begin{table}[p]
\centering
\small
\setlength{\tabcolsep}{4pt}
\caption{Complete saved configurations, dictionary part 1. Codes are defined by component prefix and the exact operator dictionary. Configuration identifiers follow alphabetical workflow order, not quality rank.}
\label{tab:rag-config-dictionary-1}
\begin{tabular}{@{}lp{0.80\linewidth}@{}}
\toprule
Cfg & Complete ordered component sequence \\
\midrule
C01 & \texttt{Q01 R06 S03 K01 C01 RR01 M02 X02 A02} \\
C02 & \texttt{Q02 R03 S01 K01 C01 RR01 M03 X02 A03} \\
C03 & \texttt{Q02 R03 S03 K01 C01 RR02 M03 X03 A03} \\
C04 & \texttt{Q03 R01 S01 K01 C01 RR02 M01 X02 A01} \\
C05 & \texttt{Q03 R02 S01 K01 C01 RR02 M03 X02 A03} \\
C06 & \texttt{Q03 R04 S01 K01 C01 RR01 M02 X02 A02} \\
C07 & \texttt{Q03 R05 S01 K01 C01 RR01 M03 X02 A03} \\
C08 & \texttt{Q03 R07 S03 K01 C01 RR01 M01 X03 A01} \\
C09 & \texttt{Q03 R12 S04 K01 C01 RR01 M03 X01 A03} \\
C10 & \texttt{Q04 R04 S01 K01 C01 RR01 M02 X02 A02} \\
C11 & \texttt{Q04 R05 S01 K01 C01 RR01 M03 X02 A03} \\
C12 & \texttt{Q04 R10 S01 K01 C01 RR02 M02 X02 A02} \\
C13 & \texttt{Q05 R04 S01 K01 C01 RR01 M02 X02 A02} \\
C14 & \texttt{Q05 R06 S03 K01 C01 RR01 M02 X02 A02} \\
C15 & \texttt{Q06 R05 S01 K01 C01 RR01 M03 X02 A03} \\
C16 & \texttt{Q07 R06 S03 K01 C01 RR01 M02 X02 A02} \\
C17 & \texttt{Q07 R09 S01 K01 C01 RR01 M01 X03 A01} \\
C18 & \texttt{Q08 R08 S03 K01 C01 RR01 M02 X01 A02} \\
C19 & \texttt{Q08 R10 S02 K01 C01 RR01 M01 X03 A01} \\
C20 & \texttt{Q08 R11 S02 K01 C01 RR01 M01 X02 A01} \\
\bottomrule
\end{tabular}
\end{table}

\begin{table}[p]
\centering
\small
\setlength{\tabcolsep}{4pt}
\caption{Exact operator definitions, part 1. These are exact saved workflow actions.}
\label{tab:rag-operator-dictionary-1}
\begin{tabular}{@{}lp{0.80\linewidth}@{}}
\toprule
Code & Saved action \\
\midrule
Q01 & \texttt{query.\allowbreak{}decomposition} \\
Q02 & \texttt{query.\allowbreak{}hyde} \\
Q03 & \texttt{query.\allowbreak{}identity} \\
Q04 & \texttt{query.\allowbreak{}query2doc} \\
Q05 & \texttt{query.\allowbreak{}query2doc:\allowbreak{}q=\allowbreak{}1} \\
Q06 & \texttt{query.\allowbreak{}query2doc:\allowbreak{}q=\allowbreak{}3} \\
Q07 & \texttt{query.\allowbreak{}rewrite} \\
Q08 & \texttt{query.\allowbreak{}synonyms} \\
R01 & \texttt{retriever.\allowbreak{}bge\_m3\_hybrid} \\
R02 & \texttt{retriever.\allowbreak{}bge\_m3\_hybrid:\allowbreak{}dense=\allowbreak{}0.\allowbreak{}25} \\
R03 & \texttt{retriever.\allowbreak{}hyde\_e5} \\
R04 & \texttt{retriever.\allowbreak{}rrf} \\
R05 & \texttt{retriever.\allowbreak{}rrf:\allowbreak{}bm25+\allowbreak{}e5+\allowbreak{}bge+\allowbreak{}qwen:\allowbreak{}k=\allowbreak{}100} \\
R06 & \texttt{retriever.\allowbreak{}rrf:\allowbreak{}bm25+\allowbreak{}e5+\allowbreak{}bge+\allowbreak{}qwen:\allowbreak{}k=\allowbreak{}60} \\
R07 & \texttt{retriever.\allowbreak{}rrf:\allowbreak{}bm25+\allowbreak{}e5:\allowbreak{}k=\allowbreak{}10} \\
R08 & \texttt{retriever.\allowbreak{}rrf:\allowbreak{}bm25+\allowbreak{}qwen:\allowbreak{}k=\allowbreak{}10} \\
R09 & \texttt{retriever.\allowbreak{}rrf:\allowbreak{}bm25+\allowbreak{}qwen:\allowbreak{}k=\allowbreak{}60} \\
R10 & \texttt{retriever.\allowbreak{}rrf:\allowbreak{}e5+\allowbreak{}bge:\allowbreak{}k=\allowbreak{}100} \\
R11 & \texttt{retriever.\allowbreak{}weighted:\allowbreak{}bm25+\allowbreak{}qwen:\allowbreak{}alpha=\allowbreak{}0.\allowbreak{}25} \\
R12 & \texttt{retriever.\allowbreak{}weighted:\allowbreak{}bm25+\allowbreak{}qwen:\allowbreak{}alpha=\allowbreak{}0.\allowbreak{}5} \\
S01 & \texttt{selection.\allowbreak{}mmr:\allowbreak{}lambda=\allowbreak{}0.\allowbreak{}3} \\
S02 & \texttt{selection.\allowbreak{}mmr:\allowbreak{}lambda=\allowbreak{}0.\allowbreak{}6} \\
S03 & \texttt{selection.\allowbreak{}mmr:\allowbreak{}lambda=\allowbreak{}0.\allowbreak{}9} \\
S04 & \texttt{selection.\allowbreak{}relevance} \\
\bottomrule
\end{tabular}
\end{table}

\begin{table}[p]
\centering
\small
\setlength{\tabcolsep}{4pt}
\caption{Exact operator definitions, part 2. These are exact saved workflow actions.}
\label{tab:rag-operator-dictionary-2}
\begin{tabular}{@{}lp{0.80\linewidth}@{}}
\toprule
Code & Saved action \\
\midrule
K01 & \texttt{retrieve.\allowbreak{}k=\allowbreak{}10} \\
C01 & \texttt{control.\allowbreak{}stop} \\
RR01 & \texttt{reranker.\allowbreak{}identity} \\
RR02 & \texttt{reranker.\allowbreak{}qwen3} \\
M01 & \texttt{top\_m=\allowbreak{}2} \\
M02 & \texttt{top\_m=\allowbreak{}4} \\
M03 & \texttt{top\_m=\allowbreak{}6} \\
X01 & \texttt{context.\allowbreak{}ends} \\
X02 & \texttt{context.\allowbreak{}ranked} \\
X03 & \texttt{context.\allowbreak{}reverse} \\
A01 & \texttt{answer.\allowbreak{}generate:\allowbreak{}m=\allowbreak{}2} \\
A02 & \texttt{answer.\allowbreak{}generate:\allowbreak{}m=\allowbreak{}4} \\
A03 & \texttt{answer.\allowbreak{}generate:\allowbreak{}m=\allowbreak{}6} \\
\bottomrule
\end{tabular}
\end{table}

\ifdefined\AgentUCTWithAppendix\clearpage\fi

\subsection{Physical tokens, crossing evaluations and cache reuse}
\label{sec:resource-analysis}
We aggregated the physical charge identifiers already attached to each complete
search evaluation. At each cutoff, only identifiers in its saved complete prefix
contributed to the retained bill. Raw input/output tokens were checked against
all recorded model attempts and the frozen model-size weights. Table~\ref{tab:raw-model-tokens}
reports raw model categories, while Table~\ref{tab:search-crossing} separates
weighted retained cost from the entire first completed evaluation after the
cutoff. The latter is observed in the original 32M search, not an independently
rerun early-stop experiment. Additional held-out measurement bills remain
separate. Table~\ref{tab:original-phase-tokens} reports all original phases;
only full search and validation enter the declared $C_{\rm opt}$.

LLM input tokens account for 96.2\% of retained 32M weighted search cost in
each method, pooling its five seed bills. Encoder tokens and LLM output tokens
are reported explicitly alongside the weighted total (Figure~\ref{fig:resource-composition}).
The original receipts contain 300,064 model attempts without semantic
\texttt{purpose}/\texttt{role} fields. Their component-edge labels identify
where a bill was recorded, not necessarily the internal purpose of each call;
for example, a \texttt{retrieve.k} edge can contain generation. We do not infer
an exhaustive query-generation/reranking/answering split from those labels.
Paid failure costs are already included in the total. An edge can contain
several model attempts for productive work as well as for retries.

\begin{table}[htbp]
\centering\small\setlength{\tabcolsep}{3pt}
\caption{Raw retained-search token counts, mean $\pm$ sample SD over five seeds.
M and k denote millions and thousands. Encoder columns are input tokens;
all generative roles share Qwen3.5-4B. These are observed logical request tokens,
including recorded attempts and excluding padding.}
\label{tab:raw-model-tokens}
\begin{tabular}{@{}llrrrrr@{}}
\toprule
$B$ (M) & Method & LLM in (M) & LLM out (M) & E5 (k) & BGE (k) & Qwen enc. (k) \\
\midrule
16 & AgentUCT & $3.167 \pm 0.187$ & $0.147 \pm 0.011$ & $26.7 \pm 10.2$ & $26.9 \pm 15.3$ & $23.9 \pm 18.9$ \\
16 & MCTS-UCT & $3.170 \pm 0.192$ & $0.147 \pm 0.011$ & $26.7 \pm 10.2$ & $26.9 \pm 15.3$ & $25.2 \pm 21.0$ \\
16 & AgentUCT-NextC & $3.170 \pm 0.191$ & $0.147 \pm 0.011$ & $27.3 \pm 10.3$ & $27.6 \pm 16.2$ & $25.5 \pm 21.5$ \\
16 & MCTS-Random & $3.169 \pm 0.191$ & $0.147 \pm 0.011$ & $29.2 \pm 11.9$ & $29.8 \pm 20.1$ & $27.5 \pm 24.9$ \\
24 & AgentUCT & $5.307 \pm 0.156$ & $0.210 \pm 0.014$ & $40.3 \pm 16.2$ & $48.7 \pm 9.9$ & $50.6 \pm 34.5$ \\
24 & MCTS-UCT & $5.263 \pm 0.134$ & $0.207 \pm 0.012$ & $41.2 \pm 9.8$ & $49.7 \pm 10.9$ & $41.9 \pm 27.6$ \\
24 & AgentUCT-NextC & $5.321 \pm 0.159$ & $0.209 \pm 0.014$ & $44.9 \pm 10.9$ & $46.7 \pm 13.6$ & $52.0 \pm 33.9$ \\
24 & MCTS-Random & $5.314 \pm 0.172$ & $0.209 \pm 0.014$ & $45.8 \pm 13.1$ & $51.1 \pm 17.0$ & $48.7 \pm 32.9$ \\
32 & AgentUCT & $7.124 \pm 0.154$ & $0.261 \pm 0.017$ & $57.2 \pm 19.0$ & $62.4 \pm 13.1$ & $66.0 \pm 37.0$ \\
32 & MCTS-UCT & $7.130 \pm 0.164$ & $0.262 \pm 0.018$ & $56.6 \pm 15.8$ & $63.8 \pm 12.3$ & $55.3 \pm 34.2$ \\
32 & AgentUCT-NextC & $7.138 \pm 0.156$ & $0.260 \pm 0.017$ & $59.5 \pm 10.3$ & $60.4 \pm 19.6$ & $68.1 \pm 44.6$ \\
32 & MCTS-Random & $7.105 \pm 0.128$ & $0.260 \pm 0.018$ & $61.6 \pm 10.9$ & $62.7 \pm 20.9$ & $58.4 \pm 34.0$ \\
\bottomrule
\end{tabular}

\end{table}
\begin{table}[htbp]
\centering\small\setlength{\tabcolsep}{4pt}
\caption{Retained search and the first complete crossing evaluation, in M
weighted-token units (mean $\pm$ sample SD). $N_B$ counts complete evaluations;
Selections counts saved adaptive child-selection decisions. Crossing records the
evaluation's entire bill.}
\label{tab:search-crossing}
\begin{tabular}{@{}llrrrr@{}}
\toprule
$B$ (M) & Method & Retained (M) & Next crossing (M) & $N_B$ & Selections \\
\midrule
16 & AgentUCT & $13.970 \pm 0.809$ & $2.429 \pm 0.903$ & $8.2 \pm 1.1$ & $0.4 \pm 0.9$ \\
16 & MCTS-UCT & $13.982 \pm 0.835$ & $2.413 \pm 0.905$ & $8.2 \pm 1.1$ & $0.4 \pm 0.9$ \\
16 & AgentUCT-NextC & $13.983 \pm 0.833$ & $2.468 \pm 0.920$ & $8.2 \pm 1.1$ & $0.4 \pm 0.9$ \\
16 & MCTS-Random & $13.985 \pm 0.837$ & $2.416 \pm 0.931$ & $8.2 \pm 1.1$ & $0.4 \pm 0.9$ \\
24 & AgentUCT & $23.262 \pm 0.705$ & $1.982 \pm 1.244$ & $13.4 \pm 1.5$ & $5.4 \pm 1.5$ \\
24 & MCTS-UCT & $23.067 \pm 0.587$ & $1.875 \pm 1.283$ & $13.2 \pm 1.5$ & $5.2 \pm 1.5$ \\
24 & AgentUCT-NextC & $23.319 \pm 0.709$ & $1.984 \pm 1.240$ & $13.4 \pm 1.5$ & $5.4 \pm 1.5$ \\
24 & MCTS-Random & $23.293 \pm 0.747$ & $1.908 \pm 1.296$ & $13.6 \pm 1.5$ & $5.6 \pm 1.5$ \\
32 & AgentUCT & $31.142 \pm 0.725$ & $1.661 \pm 0.772$ & $17.6 \pm 1.5$ & $9.6 \pm 1.5$ \\
32 & MCTS-UCT & $31.167 \pm 0.777$ & $2.088 \pm 0.653$ & $17.8 \pm 1.8$ & $9.8 \pm 1.8$ \\
32 & AgentUCT-NextC & $31.200 \pm 0.734$ & $1.691 \pm 0.796$ & $17.6 \pm 1.5$ & $9.6 \pm 1.5$ \\
32 & MCTS-Random & $31.054 \pm 0.624$ & $2.036 \pm 0.664$ & $17.8 \pm 1.8$ & $9.8 \pm 1.8$ \\
\bottomrule
\end{tabular}

\end{table}
\begin{table}[htbp]
\centering\small\setlength{\tabcolsep}{4pt}
\caption{Original 32M execution ledger by phase, mean $\pm$ sample SD across five
seeds. LLM counts are raw tokens; Cost is weighted tokens, all in millions.
Full search includes the complete crossing evaluation. Preparation uses encoder
tokens, so its zero LLM count can accompany a positive bill. Test and preparation
are excluded from $C_{\rm opt}$; later snapshot measurement campaigns are separate.}
\label{tab:original-phase-tokens}
\begin{tabular}{@{}llrrr@{}}
\toprule
Phase & Method & LLM in (M) & LLM out (M) & Cost (M) \\
\midrule
Full search & AgentUCT & $7.507 \pm 0.183$ & $0.273 \pm 0.015$ & $32.804 \pm 0.808$ \\
Full search & MCTS-UCT & $7.613 \pm 0.216$ & $0.275 \pm 0.017$ & $33.255 \pm 0.982$ \\
Full search & AgentUCT-NextC & $7.528 \pm 0.189$ & $0.272 \pm 0.015$ & $32.890 \pm 0.817$ \\
Full search & MCTS-Random & $7.576 \pm 0.219$ & $0.272 \pm 0.017$ & $33.090 \pm 0.996$ \\
Validation & AgentUCT & $6.041 \pm 0.841$ & $0.174 \pm 0.013$ & $26.199 \pm 3.576$ \\
Validation & MCTS-UCT & $6.105 \pm 0.856$ & $0.156 \pm 0.015$ & $26.369 \pm 3.650$ \\
Validation & AgentUCT-NextC & $5.867 \pm 0.812$ & $0.169 \pm 0.023$ & $25.429 \pm 3.493$ \\
Validation & MCTS-Random & $6.095 \pm 0.878$ & $0.171 \pm 0.010$ & $26.419 \pm 3.710$ \\
Test & AgentUCT & $0.926 \pm 0.091$ & $0.035 \pm 0.007$ & $4.061 \pm 0.422$ \\
Test & MCTS-UCT & $1.123 \pm 0.448$ & $0.028 \pm 0.007$ & $4.847 \pm 1.884$ \\
Test & AgentUCT-NextC & $1.147 \pm 0.435$ & $0.029 \pm 0.007$ & $4.959 \pm 1.826$ \\
Test & MCTS-Random & $1.105 \pm 0.458$ & $0.029 \pm 0.008$ & $4.779 \pm 1.921$ \\
Preparation & AgentUCT & $0.000 \pm 0.000$ & $0.000 \pm 0.000$ & $1.792 \pm 0.000$ \\
Preparation & MCTS-UCT & $0.000 \pm 0.000$ & $0.000 \pm 0.000$ & $1.792 \pm 0.000$ \\
Preparation & AgentUCT-NextC & $0.000 \pm 0.000$ & $0.000 \pm 0.000$ & $1.792 \pm 0.000$ \\
Preparation & MCTS-Random & $0.000 \pm 0.000$ & $0.000 \pm 0.000$ & $1.792 \pm 0.000$ \\
\bottomrule
\end{tabular}

\end{table}
\begin{figure}[htbp]
\centering
\includegraphics[width=\linewidth]{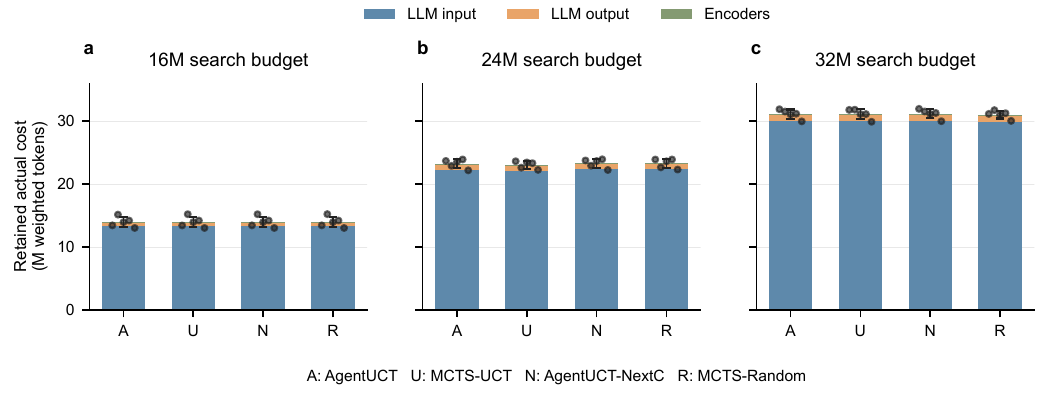}
\caption{Physical composition of each retained search bill. Bars show five-seed
means, dots show each seed's total, and whiskers show sample SD of the total.
The encoder contribution is small on this common scale and is resolved in
Table~\ref{tab:raw-model-tokens}. Complete crossing and held-out bills are excluded
from these bars and reported separately.}
\label{fig:resource-composition}
\end{figure}

\paragraph{Bipartite inventory and logical reuse.}
A saved key is a pair consisting of a query identifier and its complete
execution prefix. We counted these objects from completed-prefix records,
without instantiating an earlier cache or consulting keys introduced by later
search or held-out evaluation. The 32M inventories also matched the saved
materialized inventories. For this fixed full-panel design, every successful
prefix covers all 150 queries, so adjacency density over the materialized
prefix set is one at every measured cutoff. This density uses the materialized
prefix set as its denominator and instantiates the rectangular case of the
information analysis. A graph-density efficiency effect remains unmeasured.

The logical hit fraction is the sum of recorded restored-prefix edges divided
by the sum of all workflow-prefix visits across complete question evaluations.
Its denominator includes zero-token control and ordering edges, alongside
model-bearing workflow-prefix visits. At 32M, AgentUCT, MCTS-UCT, AgentUCT-NextC and MCTS-Random have mean
logical hit fractions of 6.79\%, 6.81\%, 6.27\% and 7.04\%, respectively
(Table~\ref{tab:cache-efficiency}). AgentUCT's 1.35-point test-F1 increase over
AgentUCT-NextC accompanies only 0.183\% fewer raw LLM tokens in retained search.
Its LLM tokens per complete evaluation are higher than MCTS-UCT's (421.7k versus
418.1k). Differences in the selected workflow mix affect both quantities.
A matched no-cache comparison remains unmeasured here, so these observations
alone leave avoided tokens and wall-clock speedup unquantified (Figure~\ref{fig:cache-work}).

\begin{table}[htbp]
\centering\small\setlength{\tabcolsep}{3pt}
\caption{Cache inventory and observed work, mean $\pm$ sample SD over five seeds.
Keys are distinct query--prefix pairs, Prefixes counts distinct full prefixes,
and Visits counts logical prefix accesses. Hit is a per-run ratio averaged
across seeds. LLM/eval is actual input plus output tokens per complete
150-question evaluation. Densities over the observed prefix sets are one.}
\label{tab:cache-efficiency}
\begin{tabular}{@{}llrrrrr@{}}
\toprule
$B$ (M) & Method & Keys (k) & Prefixes & Visits (k) & Hit (\%) & LLM/eval (k) \\
\midrule
16 & AgentUCT & $11.01 \pm 1.36$ & $73.4 \pm 9.0$ & $11.07 \pm 1.48$ & $0.44 \pm 0.99$ & $408.3 \pm 45.9$ \\
16 & MCTS-UCT & $11.01 \pm 1.36$ & $73.4 \pm 9.0$ & $11.07 \pm 1.48$ & $0.44 \pm 0.99$ & $408.6 \pm 45.6$ \\
16 & AgentUCT-NextC & $11.01 \pm 1.36$ & $73.4 \pm 9.0$ & $11.07 \pm 1.48$ & $0.44 \pm 0.99$ & $408.6 \pm 45.6$ \\
16 & MCTS-Random & $11.01 \pm 1.36$ & $73.4 \pm 9.0$ & $11.07 \pm 1.48$ & $0.44 \pm 0.99$ & $408.6 \pm 45.6$ \\
24 & AgentUCT & $17.19 \pm 1.86$ & $114.6 \pm 12.4$ & $18.09 \pm 2.05$ & $4.92 \pm 1.36$ & $415.6 \pm 43.5$ \\
24 & MCTS-UCT & $16.95 \pm 1.80$ & $113.0 \pm 12.0$ & $17.82 \pm 2.00$ & $4.82 \pm 1.09$ & $418.5 \pm 45.3$ \\
24 & AgentUCT-NextC & $17.28 \pm 1.82$ & $115.2 \pm 12.1$ & $18.09 \pm 2.05$ & $4.40 \pm 0.83$ & $416.6 \pm 43.8$ \\
24 & MCTS-Random & $17.34 \pm 1.79$ & $115.6 \pm 11.9$ & $18.36 \pm 2.05$ & $5.45 \pm 2.07$ & $410.0 \pm 44.6$ \\
32 & AgentUCT & $22.14 \pm 1.82$ & $147.6 \pm 12.1$ & $23.76 \pm 2.05$ & $6.79 \pm 0.82$ & $421.7 \pm 30.9$ \\
32 & MCTS-UCT & $22.38 \pm 2.14$ & $149.2 \pm 14.2$ & $24.03 \pm 2.41$ & $6.81 \pm 1.23$ & $418.1 \pm 34.8$ \\
32 & AgentUCT-NextC & $22.26 \pm 1.80$ & $148.4 \pm 12.0$ & $23.76 \pm 2.05$ & $6.27 \pm 0.66$ & $422.5 \pm 30.9$ \\
32 & MCTS-Random & $22.32 \pm 2.05$ & $148.8 \pm 13.7$ & $24.03 \pm 2.41$ & $7.04 \pm 1.56$ & $416.7 \pm 36.0$ \\
\bottomrule
\end{tabular}

\end{table}
\begin{figure}[htbp]
\centering
\includegraphics[width=\linewidth]{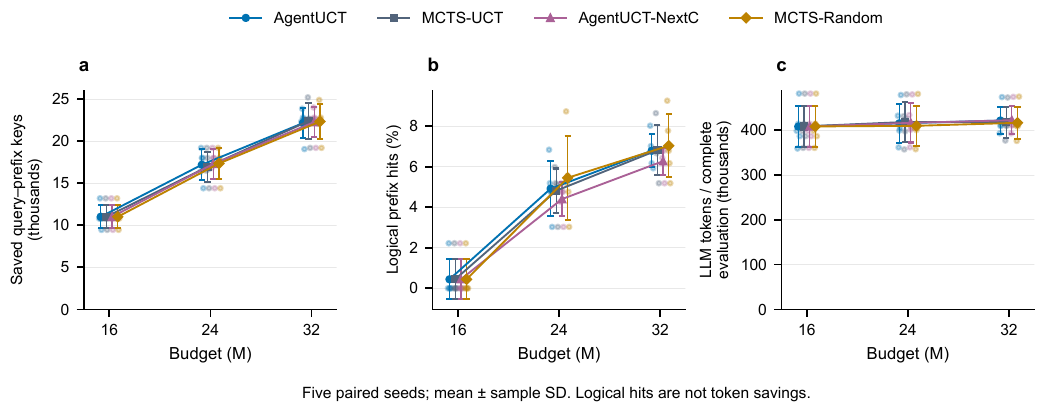}
\caption{Saved objects, observed logical reuse and actual LLM work have different
interpretations. Markers/lines show means, whiskers show sample SD, and faint
points show the five seeds. Horizontal offsets separate methods at each shared
budget. SD measures spread across seeds; a lower whisker can extend below
zero. Cache hits describe observed logical reuse; a matched saving comparison
remains unmeasured.}
\label{fig:cache-work}
\end{figure}

\subsection{Execution variation and trace coverage}
Fresh executions can differ numerically. Across three shared initialization
paths and 450 question evaluations per method, MCTS-UCT differed from AgentUCT on
four answers, two affecting F1; AgentUCT-NextC and MCTS-Random matched AgentUCT.
The largest panel-mean difference was $9.52\times10^{-4}$, with identical
recorded batch prompts and query order at the differing answer. Within one
run, a successful terminal checkpoint supplies its saved prediction; the
sampled main trajectories contained no repeated complete terminal.
The quality-regret theorem assumes a fixed evaluator table, whereas these
empirical comparisons retain observed variation between fresh executions.
After the initial stage, an observer-copy repair and a retrieval-service
memory reservation optimization were deployed with separate provenance.
Fixed-request checks preserved rankings and token charges, without establishing
global bitwise equivalence across fresh runs.

The implementation must pass evidence-order, query-history, stopping,
checkpoint-restoration, physical-price update and recovery regressions.
Model-attempt records retain checkpoint, prompt, output, input/output
tokens and available timing/failure metadata. Semantic role labels were absent
from the original receipts, as noted above. Candidate traces retain raw backup statistics,
probabilities, current cost forecasts and selected actions. Search and tiny
queues perform no trace-reconstruction audit. Offline audits check saved
candidate arithmetic, backup, predictions, checkpoints and bills directly
from saved records, with search, random draws, forecasts and generation
retained as originally observed.
Prediction calibration requires separate evidence. The runtime journal persists
physical receipts and completed items; post-hoc accounting reads these
records separately from component dispatch. Four query workers
process fixed consecutive QID groups inside one candidate. Before each model
dispatch, every unfinished group member either requests generation or finishes;
requests with equal output limits are batched in stable QID order. Cached or
failed items leave the group explicitly, and all submitted work drains before
an error propagates. Provider observers and receipts are per-query; retrieval
cache reset and ranking are atomic. All 150 training items settle before
scoring, price updates and the next search decision.
Call elapsed times may include component queueing; queue time and component
execution time are separately recorded. With concurrent work, their sum
measures accumulated component time; wall-clock duration requires its own measurement. Top-$m$ selection precedes
CPU ordering; ordering preserves selected text, and the final answer stage
generates the workflow's answer.

Runtime diagnostics count each physical generation batch once, using its
session and batch identifiers. Padding efficiency is the ratio of actual
tokens to padded tensor positions, reported separately for inputs and outputs.
Batch timings from a recent receipt window diagnose execution overhead;
end-to-end speed comparisons require completed, matched workloads.

New and resumed follow-up RAG runs reuse the native checkpoint already read
when a new prefix completes, removing a second checkpoint read and graph
compilation. Cache hits and process recovery still read durable checkpoints.
The saved runtime record identifies the implementation used for each launch;
search settings, fixed batches, checkpoints and actual bills are preserved.

Earlier multi-round runs remain archived diagnostics. Independent calibration
of that source pool used 34 fixed paths on three separate questions, producing
102 evaluations, 933 component receipts and 383 Qwen generation attempts.
The shared calibration bill was 1,171,878.42 weighted-token units:
1,165,642.43 for execution and 6,235.99 for index preparation. Its recorded
425.44 seconds include preparation and output persistence and exclude model
loading. The main grammar retains the
corresponding query/retrieval/reranking forecasts and transfers the former
width-conditioned answer prices to explicit final-answer edges. Removed
compression actions receive no mass in the new completion law. These transferred
price priors and the common numeric scale remain fixed across methods and tiny.
Calibrated error bounds remain unavailable. Main-runtime acceptance additionally
checks the complete one-tree, raw-backup, cache and Top-5 protocol, as well as
complete token/time records on the new trajectories.
The frozen parser accepts a singular \texttt{citation} key as an alias for
\texttt{citations} and closes one missing final object brace only when the
result is otherwise valid JSON. It still validates citation identifiers and
rejects unfinished strings or arrays. Raw output and each normalization are
retained. The revised calibration has no query fallback.

\subsection{Single-factor completion-pricing interventions}
\label{sec:single-factor-mechanism}
The two interventions test individual ingredients of AgentUCT's complete-cost
forecast in the original retrospective HotpotQA grammar. Frozen prices retain
the calibrated action prices throughout search, with live cache membership,
historical continuation demand, estimator overhead and the completion sampler
active. Uniform continuation replaces the historical completion law by uniform
legal children while retaining online price updates, live cache membership,
overhead and the original 16 completion samples. Each control preserves the
selector, $\beta$, calibration scale, eight-root initialization, raw-quality
backup, grammar, prompts, Qwen3.5-4B backend, fixed-four batching and panels.
The design was fixed before observing the new control outcomes. New private
execution namespaces keep their cache and bills separate from the original
five-seed AgentUCT reference.

Each control has five complete searches with seeds 11/23/37/53/71. The same
150-question training and 300-question validation panels select a workflow
from up to five training-ranked candidates at each 16M/24M/32M affordable
complete snapshot. Official answer F1 is measured on the original 300 primary
test questions; the preserved diagnostic sentinel is excluded from that
primary score. Test outcomes do not select workflows or update search.
Table~\ref{tab:single-factor-main} gives equally weighted means and sample SD;
Table~\ref{tab:single-factor-direct-seeds} retains every paired seed and budget.
The source seed is the comparison unit. Shared questions and repeated
endpoints correlate budget measurements.

\begin{table}[htbp]
\centering\small
\caption{All 45 direct single-factor/reference score cells. Each entry is
primary test F1 (\%) followed by the saved affordable complete-evaluation
count $N$ in parentheses. No method-difference column is used.}
\label{tab:single-factor-direct-seeds}
\begin{tabular}{@{}llrrr@{}}
\toprule
Method & Seed & 16M & 24M & 32M\\
\midrule
AgentUCT & 11 & 69.51093 (8) & 69.51093 (13) & 69.51093 (18)\\
AgentUCT & 23 & 67.05137 (10) & 69.52280 (15) & 69.52280 (19)\\
AgentUCT & 37 & 69.19516 (8) & 69.19754 (14) & 69.19754 (18)\\
AgentUCT & 53 & 63.20591 (7) & 69.96565 (14) & 69.96565 (18)\\
AgentUCT & 71 & 68.01443 (8) & 69.63048 (11) & 69.63048 (15)\\
Frozen prices & 11 & 69.51093 (8) & 69.51093 (13) & 69.51093 (18)\\
Frozen prices & 23 & 67.05137 (10) & 69.52280 (15) & 69.52280 (19)\\
Frozen prices & 37 & 69.19516 (8) & 68.68706 (14) & 68.68706 (18)\\
Frozen prices & 53 & 63.20591 (7) & 69.84902 (14) & 69.84902 (18)\\
Frozen prices & 71 & 68.01443 (8) & 67.73381 (11) & 67.91143 (15)\\
Uniform continuation & 11 & 69.51093 (8) & 69.51093 (13) & 69.51093 (18)\\
Uniform continuation & 23 & 67.05137 (10) & 67.84516 (15) & 67.84516 (19)\\
Uniform continuation & 37 & 69.19516 (8) & 69.19754 (14) & 65.48153 (18)\\
Uniform continuation & 53 & 63.20591 (7) & 69.84902 (13) & 69.84902 (18)\\
Uniform continuation & 71 & 68.01443 (8) & 69.32492 (11) & 69.32492 (15)\\
\bottomrule
\end{tabular}
\end{table}

At 16M, all five paired endpoint scores tie for both interventions. At 32M,
AgentUCT exceeds frozen prices in seeds 37/53/71 and ties seeds 11/23;
it exceeds uniform continuation in seeds 23/37/53/71 and ties seed 11.
At 24M, frozen prices have lower scores in seeds 37/53/71 and ties in 11/23;
uniform continuation has lower scores in 23/53/71 and ties in 11/37.
Every 16M and 32M pair retains the same $N$. At 24M, uniform-continuation
seed 53 retains 13 complete evaluations compared with AgentUCT's 14; the
other paired counts match. These outcomes identify the endpoint behavior of
the declared forecast interventions in this grammar. The paired counts and
endpoint identities distinguish changes in workflow choice from changes in
evaluation count.

\paragraph{Actual bills and crossing work.}
Table~\ref{tab:single-factor-phase-bills} reports the entire ten-unit bill in
M billion-parameter-token units. Search includes initialization and the whole
complete evaluation crossing 32M. Validation and test cover all three
reporting cutoffs, with native same-endpoint measurement reuse.
Input-only preparation is charged once to frozen-price seed 11:
1.792279M. The ten-unit total is 834.146083M, comprising 327.891294M search,
504.462509M heldout measurement and that unique preparation bill.
The original AgentUCT full phase references cover only its original 32M
heldout measurement, excluding later 16M/24M diagnostics. Their different
measurement scope prevents a full-bill saving comparison with these controls.

\begin{table}[htbp]
\centering\small
\caption{Complete control bills in M billion-parameter-token units. Full
includes search, validation, test and the displayed unique preparation owner.
Crossing work is already included in Search; endpoint measurements can be
reused across budgets.}
\label{tab:single-factor-phase-bills}
\begin{tabular}{@{}llrrrrr@{}}
\toprule
Control & Seed & Search & Validation & Test & Prep & Full\\
\midrule
Frozen prices & 11 & 33.494831 & 42.288278 & 3.640914 & 1.792279 & 81.216302\\
Frozen prices & 23 & 32.086795 & 38.365484 & 8.222437 & 0.000000 & 78.674717\\
Frozen prices & 37 & 32.305456 & 39.231063 & 11.959296 & 0.000000 & 83.495816\\
Frozen prices & 53 & 33.701803 & 33.254490 & 12.920942 & 0.000000 & 79.877235\\
Frozen prices & 71 & 32.212775 & 49.848234 & 23.875805 & 0.000000 & 105.936814\\
Uniform continuation & 11 & 33.546071 & 38.667244 & 3.640914 & 0.000000 & 75.854229\\
Uniform continuation & 23 & 32.153012 & 28.550742 & 8.369418 & 0.000000 & 69.073171\\
Uniform continuation & 37 & 32.302695 & 38.297609 & 15.800652 & 0.000000 & 86.400955\\
Uniform continuation & 53 & 33.919419 & 33.298431 & 12.920942 & 0.000000 & 80.138793\\
Uniform continuation & 71 & 32.168437 & 49.642647 & 11.666966 & 0.000000 & 93.478050\\
\bottomrule
\end{tabular}
\end{table}

\begin{table}[htbp]
\centering\small
\caption{Crossing subsets and saved execution inventory for every control.
Whole and Over are M billion-parameter-token units: Whole is the entire last
crossing evaluation, and Over is the part above 32M, already contained in Whole
and Search. Receipts include zero-charge workflow edges; they differ from
recorded model attempts and physical generation batches. Failed counts failed
physical execution receipts.}
\label{tab:single-factor-crossing-ledger}
\begin{tabular}{@{}llrrrrrr@{}}
\toprule
Control & Seed & Whole & Over & Receipts & Failed & Attempts & Batches\\
\midrule
Frozen prices & 11 & 1.658640 & 1.494831 & 51910 & 0 & 20934 & 6110\\
Frozen prices & 23 & 0.762675 & 0.086795 & 51618 & 0 & 19266 & 5418\\
Frozen prices & 37 & 1.169940 & 0.305456 & 52668 & 0 & 18888 & 5291\\
Frozen prices & 53 & 2.588993 & 1.701803 & 47868 & 0 & 19473 & 5584\\
Frozen prices & 71 & 2.189110 & 0.212775 & 54326 & 0 & 22846 & 6417\\
Uniform continuation & 11 & 1.712323 & 1.546071 & 50109 & 0 & 20606 & 6024\\
Uniform continuation & 23 & 0.706428 & 0.153012 & 49218 & 0 & 18053 & 5105\\
Uniform continuation & 37 & 1.182752 & 0.302695 & 55076 & 0 & 19807 & 5572\\
Uniform continuation & 53 & 2.614015 & 1.919419 & 47568 & 0 & 19490 & 5552\\
Uniform continuation & 71 & 2.195845 & 0.168437 & 51167 & 0 & 20520 & 5741\\
\bottomrule
\end{tabular}
\end{table}

All 511,528 execution receipts settled, including zero-charge edges; none is a
failed physical receipt. The saved complete phase inventory has no separately
charged proposer, repair or retry phase in these two controls. Provider-internal
model attempts remain distinct from those phase labels. The ten whole last
crossing evaluations cost 16.780722M; their 7.891294M overshoot is a subset,
with no additional charge. Earlier snapshots retain their saved rewards and
cache boundary, excluding later crossing observations.

\paragraph{Validation recovery.}
Uniform-continuation seed 37 finished its search before a validation
interruption. Its second paid invocation continued heldout evaluation from
the three frozen saved training-ranked shortlists and reused prior partial
validation. It generated no new search calls and reconstructed no tree,
forecast or search RNG. The two invocations retain all 24,714 original and
30,362 continuation receipts, totaling 55,076. Its full 86.400955M bill includes
32.302695M search, 38.297609M validation and 15.800652M test measurement.
The original failed lifecycle remains archived.

\subsection{AgentUCT-StaticFull ablation and selection depth}
\label{sec:static-complete-design}
AgentUCT-StaticFull is the static complete-cost ablation of AgentUCT.
It tests whether online updating improves on a fixed complete-cost prior. Before its search begins, independent calibration fixes
the same resource-price priors and caps as AgentUCT, giving prices $\bar w_r$.
Any estimator overhead $\bar h(p,a)$ is fixed at its pre-search value.
The continuation law
$\bar\rho(\cdot\mid p,a)$ is the original estimator's zero-history law:
each successive legal child is equally likely, which does not generally
make complete terminals equally likely. The original 16 completion samples
and smoothing are retained. A private seed--state--action random tape fixes
the sampled completions without advancing search randomness. The frozen target is
\[
 \bar b(p,a)=\mathbb E_{X\sim\bar\rho(\cdot\mid p,a)}
                 \left[\sum_{r\in\mathcal D(X)}\bar w_r\right]+\bar h(p,a),
\]
and is estimated by the fixed mean of 16 sampled completions, using an empty
workflow-execution cache as the reference. Structural
zero-cost steps remain free. Shared preparation retains its original separate
account. Prices, continuation probabilities and the sampled completions for
each state--action remain fixed; the estimator reads neither live cache
membership nor later execution observations. Previously unseen states are
priced lazily from these frozen inputs, without tree enumeration, historical
reconstruction or consumption of the search random stream.

Execution still restores and extends the actual WTB cache. Raw F1 backup,
initialization, exploration, rollout, $\beta$, $C_{\rm ref}$, models, panels,
physical batching and Top5--validation--test selection match AgentUCT. Five
additional static-control searches use seeds 11, 23, 37, 53 and 71 and report
the same 16M, 24M and 32M actual-cost cutoffs. Only the price used in selection
is held static; actual failure, retry and crossing charges remain billed.
This intervention jointly evaluates dynamic cache conditioning, price
learning and continuation updates. Its three changes remain combined;
Section~\ref{sec:single-factor-mechanism} separately varies calibrated prices
and the continuation law while preserving live cache conditioning. Across all five paired seeds, AgentUCT reaches $69.57\pm0.28\%$
at both 24M and 32M, compared with $67.36\pm2.67\%$ for AgentUCT-StaticFull.
The paired difference is $2.21\pm2.89$ percentage points, with a median of
0.94 points. Seeds 37, 53 and 71 improve by 0.94, 6.76 and 3.35 points;
seeds 11 and 23 tie. All five 16M endpoints tie. At 32M, each pair retains
the same evaluation count (Table~\ref{tab:completed-static-full-paired}).
The selected workflows differ in all three seeds that improve. At 24M,
seed 53 retains fourteen dynamic evaluations versus thirteen static ones;
the other pairs have equal counts. The 24M and 32M endpoints repeat within
every seed and supply the same five paired quality comparisons.

Seeds 11 and 23 select the same configurations under both estimators: C17 at
every cutoff for seed 11, and C18/C01/C01 for seed 23. In seed 37, the dynamic
method selects C10 at 24M and 32M; the static method selects query decomposition,
\texttt{retriever.rrf}, MMR with $\lambda=0.3$, ten retrieved documents, one
retrieval pass, identity reranking, four documents in ranked order and
\texttt{answer.generate:m=4}. In seed 53, the static method retains C03 at
every cutoff, whereas dynamic pricing selects C15 at 24M and 32M.
Seed 71 selects C05 under both methods at 16M. At 24M and 32M the dynamic
method selects C19; the static method selects query decomposition, weighted
BM25--Qwen retrieval with BM25 weight 0.75, MMR with $\lambda=0.3$, ten
retrieved documents, one retrieval pass, Qwen reranking, two documents in
ranked order and \texttt{answer.generate:m=2}. The static seed-37 and seed-71
recipes are written explicitly because they have no identifier in the
original configuration catalogue.

\begin{table*}[t]
\centering\small
\caption{Completed paired seeds for the AgentUCT-StaticFull ablation. Test F1 is
in percent and is directly compared across methods;
$N_{\rm dyn}$/$N_{\rm stat}$ count affordable complete evaluations. Both
methods use the live execution cache; only the selection-cost estimator is
frozen in AgentUCT-StaticFull. Reused endpoints across cutoffs are correlated
measurements, not additional independent trials. All five planned paired
seeds are complete.}
\label{tab:completed-static-full-paired}
\begin{tabular}{@{}rrrrrr@{}}
\toprule
Seed & $B$ (M) & AgentUCT & \shortstack{AgentUCT-\\StaticFull} & $N_{\rm dyn}$ & $N_{\rm stat}$\\
\midrule
11 & 16 & 69.51 & 69.51 & 8 & 8\\
11 & 24 & 69.51 & 69.51 & 13 & 13\\
11 & 32 & 69.51 & 69.51 & 18 & 18\\
23 & 16 & 67.05 & 67.05 & 10 & 10\\
23 & 24 & 69.52 & 69.52 & 15 & 15\\
23 & 32 & 69.52 & 69.52 & 19 & 19\\
37 & 16 & 69.20 & 69.20 & 8 & 8\\
37 & 24 & 69.20 & 68.26 & 14 & 14\\
37 & 32 & 69.20 & 68.26 & 18 & 18\\
53 & 16 & 63.21 & 63.21 & 7 & 7\\
53 & 24 & 69.97 & 63.21 & 14 & 13\\
53 & 32 & 69.97 & 63.21 & 18 & 18\\
71 & 16 & 68.01 & 68.01 & 8 & 8\\
71 & 24 & 69.63 & 66.28 & 11 & 11\\
71 & 32 & 69.63 & 66.28 & 15 & 15\\
\bottomrule
\end{tabular}

\end{table*}

Table~\ref{tab:static-full-phase-bills} records the complete physical bill for
each static-control run. Whole crossing evaluations remain charged; the
amount exceeding 32M is only part of that work. Validation and test columns
cover the requested cutoff measurements at all three budgets, with complete
measurements reused when endpoints repeat. Previously paid input preparation
is referenced separately and is not charged again as new work.

\begin{table*}[t]
\centering\small\setlength{\tabcolsep}{4pt}
\caption{AgentUCT-StaticFull phase bills in M weighted-token units. Full search
includes the entire crossing evaluation; retained cost excludes it from the
32M snapshot. Overshoot is full search minus 32M. Validation and test include
the three-cutoff measurement campaign. Paid input preparation cites previously incurred physical charges.}
\label{tab:static-full-phase-bills}
\begin{tabular}{@{}rrrrrrrr@{}}
\toprule
Seed & Full search & Retained 32M & Crossing & Overshoot & Validation & Test & Paid input prep.\\
\midrule
11 & 33.693 & 31.982 & 1.711 & 1.693 & 39.807 & 3.641 & 1.792\\
23 & 32.201 & 31.481 & 0.720 & 0.201 & 36.067 & 8.222 & 1.792\\
37 & 32.210 & 30.985 & 1.224 & 0.210 & 35.786 & 12.124 & 1.792\\
53 & 33.851 & 31.277 & 2.574 & 1.851 & 33.264 & 8.372 & 1.792\\
71 & 32.218 & 29.949 & 2.269 & 0.218 & 49.845 & 15.719 & 1.792\\
\bottomrule
\end{tabular}

\end{table*}

The implementation retains the original zero-history sampling tape and
prices the entire root-to-terminal evaluation, including its selected tree
prefix, against the empty cache. Static observation hooks skip receipt-based
price and transition updates; the execution ledger still records every actual
call. A separate run namespace preserves the original dynamic-control results.

\subsection{Complete training-Top1 measurements}
\label{sec:top1-complete}
The no-validation ablation covers all sixty method--seed--budget cells.
For each completed affordable search prefix, the highest training mean F1
selects the endpoint, with the earliest evaluation breaking ties. Its paired
selection effect is
\[
 \Delta_{s,B}^{\mathrm{selection}}
 =F^{\mathrm{train\ Top1}}_{s,B}-F^{\mathrm{Top5\to validation}}_{s,B}.
\]
Forty-nine cells reuse complete tests of the exact workflow from the same
original run. Eight additional method--seed workflow measurements, involving
four distinct configurations, supply the remaining eleven cells. They require
2,400 primary question--workflow executions and eight sentinels. Each
prediction was rescored with the official evaluator and bound to its saved
execution and terminal receipt. The new measurements contain 21,680 settled
component/preparation records and 5,791 generation attempts, costing 29.722M
weighted-token units in the separate ablation-test account. They add no search
observations and do not change historical snapshots.

\begin{figure*}[t]
\centering
\includegraphics[width=\textwidth]{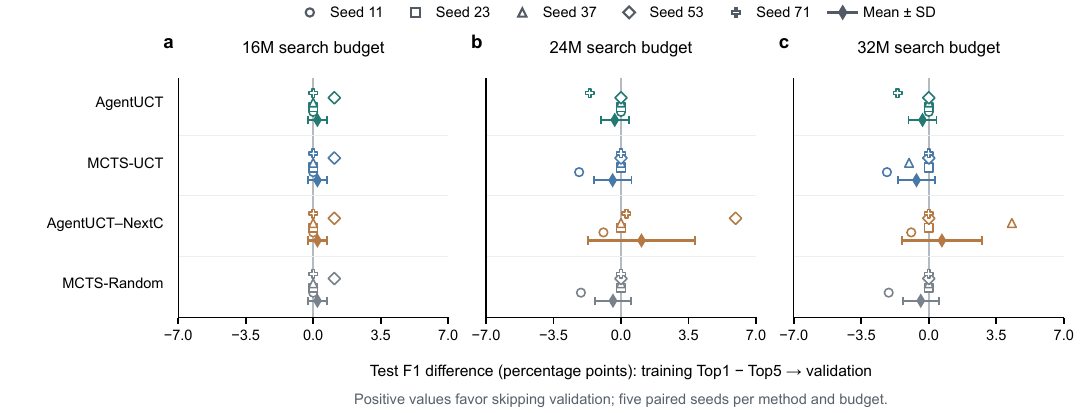}
\caption{Complete effect of skipping validation. Positive paired test-F1
changes favor training Top1. Small symbols retain each seed; the lower filled
diamond and horizontal bar show mean $\pm$ sample SD across five seeds.
Both rules use the same search prefix. Repeated budgets can reuse endpoints
and are not independent replicates.}
\label{fig:top1-ablation-all}
\end{figure*}

\begin{table*}[t]
\centering\small
\caption{Complete selection ablation. Test F1 is in percent; paired
$\Delta$ is in percentage points, positive for training Top1. Scores and
differences are mean $\pm$ sample SD across five matched seeds. Win/tie/loss
compares endpoint-selection rules within a seed.}
\label{tab:top1-results}
\begin{tabular}{llcccc}
\toprule
Budget & Method & Train Top1 & Top5 $\to$ validation & Paired $\Delta$ & Win/tie/loss\\
\midrule
16M & AgentUCT & $67.61\pm2.09$ & $67.40\pm2.54$ & $+0.22\pm0.49$ & 1/4/0\\
16M & MCTS-UCT & $67.60\pm2.08$ & $67.38\pm2.52$ & $+0.22\pm0.49$ & 1/4/0\\
16M & AgentUCT--NextC & $67.61\pm2.09$ & $67.40\pm2.54$ & $+0.22\pm0.49$ & 1/4/0\\
16M & MCTS-Random & $67.61\pm2.09$ & $67.40\pm2.54$ & $+0.22\pm0.49$ & 1/4/0\\
\midrule
24M & AgentUCT & $69.24\pm0.74$ & $69.57\pm0.28$ & $-0.32\pm0.72$ & 0/4/1\\
24M & MCTS-UCT & $67.57\pm0.80$ & $68.01\pm1.12$ & $-0.43\pm0.97$ & 0/4/1\\
24M & AgentUCT--NextC & $68.90\pm0.59$ & $67.83\pm2.69$ & $+1.06\pm2.77$ & 2/2/1\\
24M & MCTS-Random & $68.02\pm0.84$ & $68.44\pm0.89$ & $-0.42\pm0.93$ & 0/4/1\\
\midrule
32M & AgentUCT & $69.24\pm0.74$ & $69.57\pm0.28$ & $-0.32\pm0.72$ & 0/4/1\\
32M & MCTS-UCT & $67.37\pm0.53$ & $68.01\pm1.12$ & $-0.64\pm0.96$ & 0/3/2\\
32M & AgentUCT--NextC & $68.90\pm0.59$ & $68.22\pm1.96$ & $+0.68\pm2.07$ & 1/3/1\\
32M & MCTS-Random & $68.38\pm1.03$ & $68.80\pm0.87$ & $-0.42\pm0.93$ & 0/4/1\\
\bottomrule
\end{tabular}

\end{table*}

At 16M, skipping validation raises all four method means by 0.22 percentage
points, entirely from seed 53. At 32M it lowers AgentUCT by 0.32 points,
MCTS-UCT by 0.64 points and MCTS-Random by 0.42 points, while raising AgentUCT-NextC by
0.68 points. Across the sixty correlated cells, seven improve and nine
decline; the other 44 select the same workflow under both rules. AgentUCT
retains the highest mean at 24M and 32M under either rule. Other method
rankings change: AgentUCT-NextC overtakes MCTS-Random at 32M and both MCTS-UCT and MCTS-Random
at 24M. These changes are descriptive; no test outcome changes the
primary Top5--validation protocol. Omitting validation avoids its physical
bill, while the extra test measurements remain a separate research expense.

\ifdefined\AgentUCTWithAppendix\clearpage\fi
\section{Supplementary RAG allocation and resource diagnostics}
\label{sec:supp-allocation-diagnostics}
These displays retain the complete paired selection ablation, the
quality--cost tradeoff against AgentUCT-NextC, a saved allocation decision,
physical execution accounting and configuration identities. They supplement
the primary budget and task comparisons in the main paper. Five paired
search seeds, sample SDs and all endpoint identities are retained. Reused
measurements across budgets remain correlated.

The original four-MCTS 32M table separates full optimization bills from
budget-retained evaluation counts. AgentUCT's higher test F1 coexists with
lower full-bill utility than AgentUCT-NextC; no endpoint was selected by J.
The same cache and execution protocol was available to every trial, with
private caches and distinct histories.

\begin{table}[t]
\centering\small
\setlength{\tabcolsep}{2.5pt}
\caption{Five-seed results at a 32M search budget. F1 is validation-selected
primary test performance (mean $\pm$ sample SD across five seeds). $C_{\mathrm{opt}}$ is the full optimization bill in
M weighted-token units; $N_B$ counts complete evaluations retained at the
cutoff. $\Delta J$ is relative to paired MCTS-UCT runs.}
\label{tab:stage-results}
\begin{tabular}{lrrrr}
\toprule
Method & F1 (\%) $\uparrow$ & $C_{\mathrm{opt}}$ & $N_B$ & $\Delta J$\\
\midrule
AgentUCT & $\mathbf{69.57}\pm0.28$ & 59.003 & 17.6 & $+0.0367$\\
MCTS-UCT & $68.01\pm1.12$ & 59.624 & 17.8 & $0$\\
AgentUCT-NextC & $68.22\pm1.96$ & 58.319 & 17.6 & $+0.0465$\\
MCTS-Random & $68.80\pm0.87$ & 59.508 & 17.8 & $+0.0119$\\
\bottomrule
\end{tabular}
\end{table}

\begin{figure*}[t]
\centering
\includegraphics[width=\textwidth]{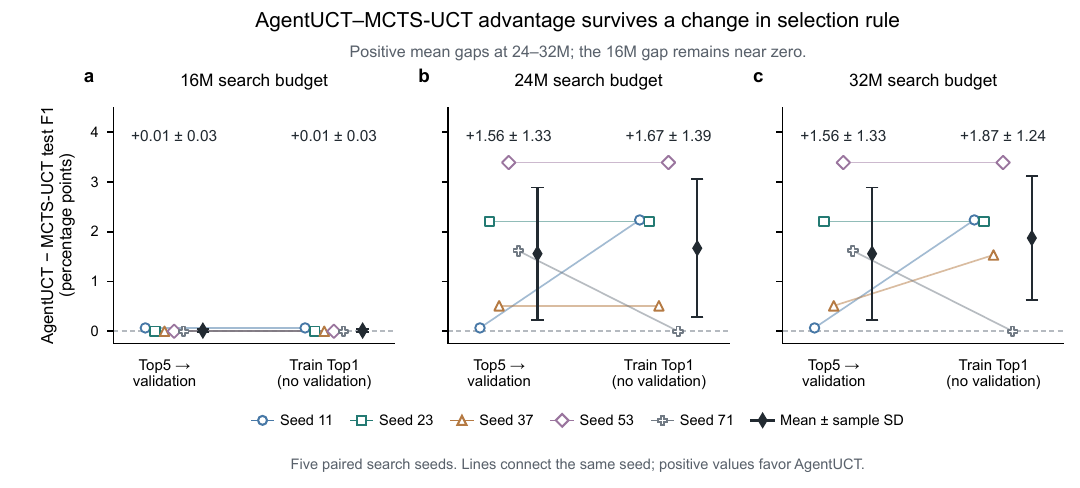}
\caption{On the original retrospective panel, AgentUCT's advantage over MCTS-UCT persists without validation.
Panels compare paired test-F1 differences under both selection rules at each
budget. Symbols retain all five seeds; summaries show mean $\pm$ sample SD.
Positive values favor AgentUCT. The 16M cutoff mostly reflects initialization;
repeated endpoints across budgets reuse complete measurements.}
\label{fig:selection-robustness}
\Description{Paired AgentUCT minus MCTS-UCT test F1 at three search budgets under validation selection and direct training Top1 selection. The positive mean advantage at 24M and 32M persists under both rules.}
\end{figure*}

\begin{figure*}[t]
\centering
\includegraphics[width=\textwidth]{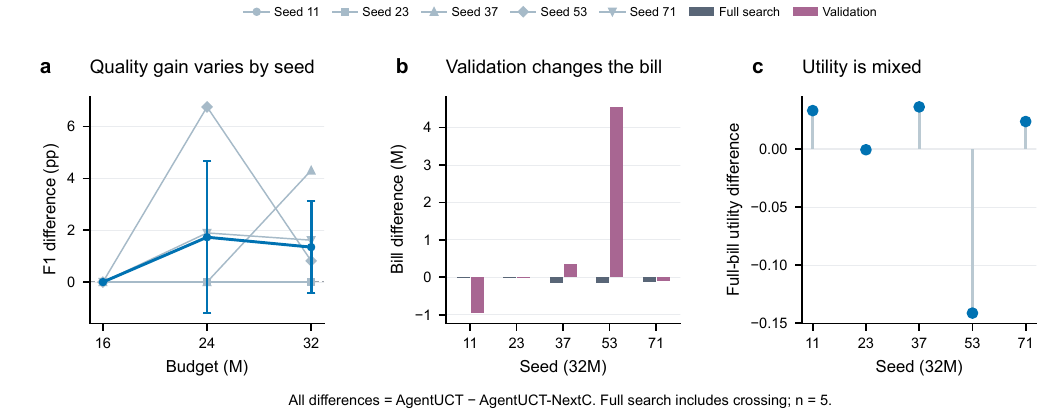}
\caption{Completion pricing relative to AgentUCT-NextC. (a) Paired test-F1
differences at each budget, with all five seeds and the mean $\pm$ sample SD.
(b) Actual full-search and validation-bill differences at 32M, including the
crossing evaluation in search. (c) Auxiliary full-bill utility differences for
the same five seeds. Positive quality and cost differences respectively mean
higher F1 and a higher bill; positive utility means higher $J$.}
\label{fig:completion-tradeoff}
\Description{Three panels show AgentUCT minus AgentUCT-NextC in test F1, full search and validation cost, and full-bill utility, retaining all five seeds.}
\end{figure*}

\begin{table}[t]
\centering\small\setlength{\tabcolsep}{3pt}
\caption{A saved decision at seed 53, iteration 9. Both actions have
$N=1$ and exploration bonus 2.075225. Cost is predicted remaining work in
M weighted tokens; the final index subtracts $2\widehat b/C_{\rm ref}$.}
\label{tab:completion-choice}
\begin{tabular}{lrrr}
\toprule
Query action & $Q+\mathrm{bonus}$ & Cost (M) & Final index\\
\midrule
Identity & 2.7265 & 1.7760 & 0.3112\\
Synonyms (selected) & 2.6912 & 1.0157 & 1.3100\\
\bottomrule
\end{tabular}
\end{table}

\begin{figure*}[t]
\centering
\includegraphics[width=\textwidth]{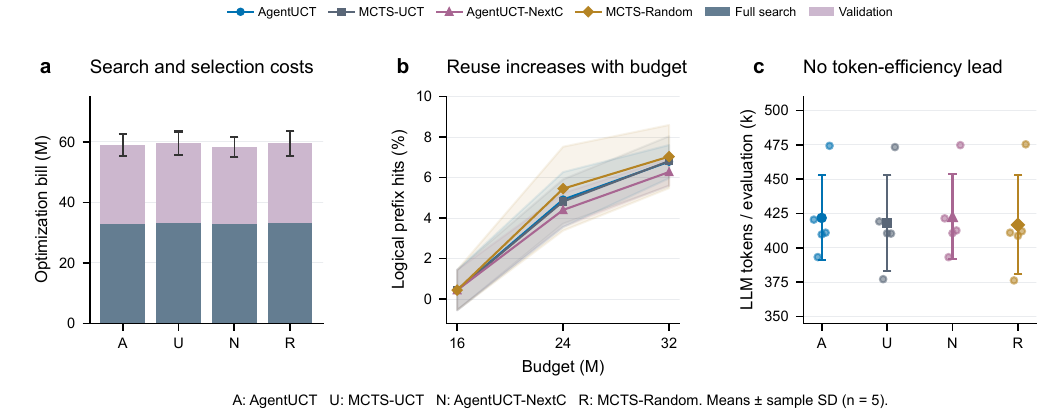}
\caption{Execution accounting for all four methods. (a) Mean full optimization
bill at 32M, separated into search and validation; error bars show the sample
SD of the total over five seeds. Test and preparation are separately reported.
(b) Logical hit fractions over all demanded prefix visits at each cutoff,
including zero-cost stages; bands show sample SD. (c) Actual unweighted LLM
input-plus-output tokens per completed evaluation at 32M, with all seeds and
mean $\pm$ sample SD. Logical hits do not measure a no-cache counterfactual.}
\label{fig:execution-accounting}
\Description{Full optimization cost is decomposed into search and validation, followed by logical prefix hit rates and physical LLM tokens per evaluation for the four algorithms.}
\end{figure*}

\begin{figure*}[t]
\centering
\includegraphics[width=\textwidth]{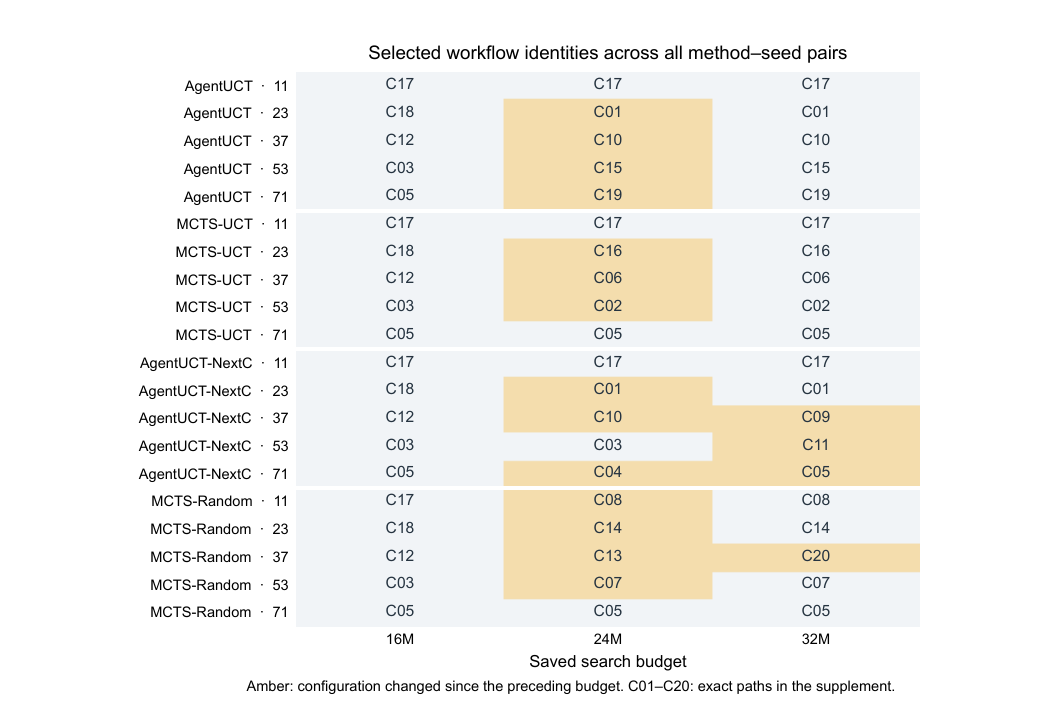}
\caption{Budget migration of validation-selected workflows. Rows retain each
method and search seed; columns use the three complete affordable snapshots.
Codes identify exact component sequences in the supplementary configuration
dictionary. Amber marks a changed configuration since the preceding budget.
Equal codes across independent runs identify the same workflow, while its
physical test execution and prediction may differ.}
\label{fig:configuration-migration}
\Description{Twenty method and seed rows show the selected configuration at 16, 24 and 32 million weighted tokens; amber marks changed configurations.}
\end{figure*}

\ifdefined\AgentUCTWithAppendix\clearpage\fi
\section{Search Algorithms in the Shared RAG Grammar}
\label{sec:rag-optimizer-adapters}

The direct comparison holds the executable workflow family fixed and changes
how the next evaluation is chosen. This section specifies the implemented
algorithms behind Table~\ref{tab:direct-algorithm-coverage}. It documents an
experimental protocol and the completed five-seed comparisons. The original algorithms motivate the selection rules, while the
categorical representation, RAG observations and shared execution service
define their use in this study.

\subsection{Common representation, feedback and accounting}
\label{sec:adapter-common-contract}
Write a legal workflow as $x=(a_1,\ldots,a_9)$. Its actions select the query,
retriever, relevance or diversity rule, retrieval width, stopping rule,
reranker, retained evidence count, context order and answer generator.
The no-loop grammar has 61,344 terminal tuples. A candidate is legal only if
every action is admitted after its preceding prefix; this enforces the
query--retriever coupling for HyDE and the matching evidence-count action
at generation. Candidate sets contain only legal terminals. Complete
workflows already evaluated within a run are excluded from further
objective evaluations.

The scalar objective is mean answer F1 over the fixed 150-question training
panel. Complete-workflow optimizers receive that value only after every
question has finished. EEIPU and LaMBO use three modules ending after
actions 1, 6 and 9: query construction, evidence construction, and context
and answer generation. These module boundaries organize the surrogate;
their objective observation remains a complete-panel F1. MSBO instead
executes two stages ending after actions 6 and 9, with an intermediate
observation explicitly defined below. No optimizer receives validation or
primary-test feedback during search.

All runs use the same persistent WTB prefix service and actual weighted-token
accounting. A prefix is declared available to a panel-level cost model only
when it is materialized for all 150 training questions. Reusing a prefix
restores its execution state. Objective measurements are created at the
complete-panel feedback boundary.
Recorded physical receipts provide both the new incremental bill and, where
needed, an earlier observed cold-stage cost. Surrogate fitting reuses those
receipts while the original physical charge remains counted once. Input-document indexing is
common preparation and is reported separately; sharing those prepared
indices transfers no predictions, search history or executed workflow
prefixes between runs.

For a reporting budget $b$, let $t_b$ be the last complete terminal-workflow
boundary with cumulative search charge $C_{t_b}\leq b$. Training Top5,
the cache view and optimizer history all come from that boundary. Validation
then chooses one of those workflows for primary-test evaluation. Earlier
snapshots cannot use a later observation, cache entry or surrogate update.
The raw account retains all intermediate stages and the full crossing
evaluation. Initialization, proposal generation, repairs and paid failures
are included when the method incurs them. Validation, test and shared
preparation remain separate cost categories. Resource-dependent schedules
use the largest reporting budget, $B=32\mathrm{M}$, throughout a run; the
16M and 24M results are snapshots of that same trajectory.

\begin{table*}[t]
\centering\small
\caption{Fixed algorithm settings. Counts refer to paid complete training
workflows unless stated otherwise. A model-based candidate set is evaluated
by the acquisition function; only its selected workflow is executed on the
training panel.}
\label{tab:rag-adapter-settings}
\begin{tabularx}{\textwidth}{@{}lXX@{}}
\toprule
Algorithm & Initialization and allocation & Fixed numerical settings\\
\midrule
AFlow--RAG & One BM25 workflow; top-four parent mixture; Qwen edits
& Uniform weight 0.3; softmax weight 0.7 and scale 20; temperature 0.7;
top-$p$ 0.9; three attempts per round; 2,048 output tokens\\
FLASH*--RAG (TPE) & D-optimal design, linear EIPS, then pruned categorical TPE
& Phase thresholds $0.15B$ and $0.30B$; at most 1,024 design signatures;
ten pruning signatures; ridge penalty 1; EI margin 0.01; TPE $\gamma=0.25$\\
EEIPU--RAG & Five warmup workflows; Top5-prefix and empty-prefix proposals
& Candidate target 1,024; 128 cost Monte Carlo draws; acquisition floor
one weighted-token unit; cost exponent $\max(0,1-C/B)$\\
LaMBO--RAG & Five warmup workflows; modular partition tree and lazy updates
& Tree depths $(1,1)$; local UCB $\mu+2\sigma$;
bandit learning rate 0.05; fixed categorical partitions\\
MSBO--RAG & Five complete warmup workflows; evidence-start or saved-prefix advance
& 256 new-path candidates; 256 cascade draws; EI fallback threshold 0.005;
$\beta=1+100\max(0,1-C/B)$; full-path window $B/105$\\
Uniform-terminal search & Uniform terminal ranks without replacement
& Same five search seeds and maximum 256 complete workflows\\
\bottomrule
\end{tabularx}
\end{table*}

\subsection{AFlow: parent selection, feedback and legal edits}
\label{sec:aflow-rag-adapter}
The implementation retains AFlow's score-weighted parent mixture and
experience-guided proposal mechanism~\citep{zhang2024aflow}. Its initial
workflow uses the original query, BM25, relevance selection, retrieval
width six, no additional retrieval round, identity reranking, four retained
documents and ranked context. This initial workflow receives the same
complete training evaluation and bill as a proposed workflow.

Completed workflows are sorted by their own training F1, with earlier
evaluation order breaking ties. If $K$ eligible workflows are available,
the parent pool contains the best $\min(4,K)$. Denoting its size by $k$,
the probability of choosing parent $i$ is
\[
 p_i=\frac{0.3}{k}
       +0.7\frac{\exp(20F_i)}{\sum_{j=1}^{k}\exp(20F_j)}.
\]
These scores use each parent's own evaluation, with descendant returns
kept separate from the AFlow parent score. The selected parent's recorded
modifications, resulting scores and unsuccessful proposals form its
experience history.

The Qwen editor receives that history, the parent tuple and the legal
grammar. Training feedback contains up to three examples: the first
imperfect predictions in frozen panel order, or the first examples if all
predictions receive F1 one. Each example includes the question, prediction,
training gold answer and per-question F1. Questions and predictions are
limited to 800 and 400 characters, respectively. Thus the editor is allowed
supervised training feedback, while held-out outcomes remain inaccessible.
The prompt requests one component edit, allowing coupled changes needed
for HyDE or matching evidence-count actions. The executor's task prompts
and model weights cannot be edited.

A response must contain a nonempty modification description and a complete
action tuple. Acceptance checks enforce grammar legality, reject a terminal
tuple already evaluated in the run, and reject an identical modification
description previously attempted from the same parent. A failed check is
returned to the editor for repair, with at most three generated responses
in total per round. Invalid graphs consume bounded repair attempts; only
accepted graphs proceed to a training evaluation. All paid responses remain in the
search bill. A round exhausted by invalid responses creates no F1 value.
The implementation caps both proposal rounds and complete training
evaluations at 256, counting the initial workflow among the latter.

Proposals use temperature 0.7, top-$p=0.9$ and at most 2,048 output tokens.
For reproducibility, proposal attempt $a$ in round $r$ has random seed
$(s+1009r+9176a)\bmod 2^{31}$. Model execution for task answers retains the
common greedy policy. Replacing Python graph edits by finite component
tuples, restricting feedback to these examples and using one fixed-panel
evaluation per valid tuple are the specified workflow-edit rules.

\subsection{FLASH*: design, structural pruning and categorical TPE}
\label{sec:flash-rag-adapter}
FLASH separates algorithm choice from conditional parameter
tuning~\citep{zhang2016flash}. We form a structural signature by retaining
component families and fusion sources while removing numeric settings
such as retrieval width, retained evidence count, mixture weights and
diversity parameters. A second representation retains every categorical
action and its setting. Both representations are one-hot encoded.

The first workflow is uniform over legal terminals. While cumulative actual
cost is below $0.15B=4.8\mathrm{M}$, or fewer than two observations exist,
the design step greedily increases the design Gram matrix's
pseudodeterminant. It considers at most 1,024 randomly sampled structural
signatures when the available set is larger. Numeric settings are drawn
uniformly among unseen legal workflows with the selected signature.
The finite online design replaces the source benchmark's initialization
schedule.

From $0.15B$ to $0.30B=9.6\mathrm{M}$, additive ridge models predict F1
and log observed execution cost from the structural encoding. Both use
penalty one and an unpenalized intercept. Let $\mu_F(x),\sigma_F(x)$ be
the quality prediction and $\widehat\ell_C(x)$ the fitted log-cost
prediction. The source-code EIPS convention implemented here is
\[
 A_{\mathrm{FLASH}}(x)=
 \frac{\operatorname{EI}(\mu_F(x),\sigma_F(x);F^*,\xi)}
      {\max(\widehat\ell_C(x),10^{-6})},\qquad \xi=0.01.
\]
In particular, the denominator is the predicted log-cost used by the
authors' public implementation and retained directly on the log-cost scale. Costs entering the logarithm have a one-unit floor. These floors
protect the predictor and leave physical zero-cost receipts unchanged.
The maximizing structural signature is selected, then an unseen legal
setting within that signature is sampled uniformly.

At $0.30B$, EIPS with $\xi=0$ ranks all structural signatures. The best ten
define the retained choices at each component position. Their Cartesian
closure, intersected with the legal grammar, forms the pruned graph; it
can contain more than ten complete structural combinations. The remaining
budget uses the FLASH* TPE variant. Hyperopt's categorical lower- and
upper-loss densities use $\gamma=0.25$, prior weight one and its linear
forgetting weights. The candidate score is the sum of log density ratios
over categorical variables, maximized over unseen legal points in the
pruned graph. Query and retriever form one joint variable so that HyDE
constraints hold. Other variables encode selection, retrieval width,
reranking, retained evidence and context order. Ties are sampled uniformly.

The 15\%/15\%/70\% phase proportions are actual-cost thresholds, not
numbers of workflow evaluations. Every selected evaluation finishes before
a phase transition. The common persistent cache replaces FLASH's native
cache management, and exhaustion of the pruned finite graph ends search.
These choices define FLASH*--RAG (TPE), distinct from a replication of a
FLASH--SMAC experiment.

\subsection{EEIPU: measured stage costs and memoized prefixes}
\label{sec:eeipu-rag-adapter}
EEIPU combines objective improvement with uncertain remaining execution
cost~\citep{essofi2024memoization}. After five uniformly drawn, distinct
complete warmup workflows, it fits one GP to F1 and three GPs to log
stage costs. Stage costs come from saved physical component calls for the
same question and action prefixes, including recorded failed attempts.
If a prefix is reused, its previously observed cold cost can train the
surrogate while the current incremental bill remains zero. An unobserved
cold cost remains missing from the surrogate's observations.

Candidate generation uses the empty prefix and the distinct available
nonterminal module prefixes of the five best training workflows. With
$P$ such prefixes, each supplies up to $\lceil1024/P\rceil$ uniformly
sampled unseen legal completions, and duplicates are removed. The resulting
pool is approximately 1,024 candidates; overlap and finite completion
sets can reduce it. The empty prefix permits changes to every component.

For candidate $x$, let $\widetilde c_j^{(d)}(x)$ be draw $d$ from stage
$j$'s lognormal cost predictor. A full-panel cached prefix sets that stage's
acquisition cost to $\epsilon=1$ weighted-token unit. The acquisition is
\[
 A_{\mathrm{EEIPU}}(x)=\operatorname{EI}(x)
 \left[\frac{1}{128}\sum_{d=1}^{128}
       \frac{1}{\sum_{j=1}^{3}\widetilde c_j^{(d)}(x)}\right]^{\eta(C)},
 \qquad \eta(C)=\max(0,1-C/B).
\]
Independent stage models use independent normal draws, with common draws
across candidates within a stage to reduce ranking noise. Objective EI is
computed analytically from the quality GP. The one-unit acquisition floor
changes the surrogate denominator; physical cache hits retain zero charge.

Our persistent cache retains every observed prefix. All
materialized full-panel prefixes can discount the acquisition, while only
Top5 prefixes bias candidate generation. This differs from restricting
physical cache capacity to the best five workflows. The categorical
Mat\'ern-$5/2$ GPs, uniform legal completion search and normal Monte Carlo
sampling are additional choices for the shared RAG domain.

\subsection{LaMBO: modular regions and slowly moving allocation}
\label{sec:lambo-rag-adapter}
LaMBO uses modular structure to reduce changes to earlier
computations~\citep{lin2021lambo}. The query and evidence modules are
partitioned separately: their distinct categorical tuples are sorted and
split at the midpoint, with the first group receiving the extra value when
the count is odd. The two binary partitions define a structure-embedding
tree with depths $(1,1)$. Partitions remain fixed throughout search.

Five uniform complete workflows initialize a GP over the full one-hot
configuration. For each tree region, the local candidate minimizes
$1-\mu_F(x)-2\sigma_F(x)$ among its eligible unseen workflows. A lazy
update keeps the previous workflow's modules before the first changed
region fixed. Region selection conditions the current arm probabilities
on the subtree determined by the preceding height. Independent random
signs determine the next height, and the slowly moving bandit recursion
updates multiplicative weights with learning rate 0.05. Only the selected
workflow is executed. Local acquisition values are surrogate predictions;
objective feedback comes from the selected workflow's complete evaluation.

Finite exhaustion requires explicit handling. Regions without eligible
unseen workflows are masked. If the conditioned subtree has no candidate,
selection returns to the available root regions. If every lazy suffix is
exhausted while unseen prefixes remain, a recorded root reset releases
the prefix restriction. The implementation retains the fixed-partition recursion
and omits the source paper's empirical discard/refine heuristic. Its tree
depths are not refitted from test results or realized cost advantages.

LaMBO's preceding-configuration switching model and WTB's cumulative
prefix availability describe different reuse conditions. This experiment
reports the latter's physical calls and charges, including reuse of a
prefix visited earlier than the immediately preceding workflow. A separate
switching-cost diagnostic would require an explicitly specified module
cost vector; it is not substituted for the measured optimization bill.

\subsection{MSBO: native stage execution and cascade decisions}
\label{sec:msbo-rag-adapter}
MSBO chooses between initiating a sample and continuing an existing
sample~\citep{torresi2026multistage}. Here an evidence sample consists of
native execution checkpoints after the first six actions, one for each
training question. Let $A_q$ be the grounded assessor's Boolean
answerability judgment, $E_q$ the materialized evidence-document set, and
$G_q\subseteq E_q$ the distinct document identifiers cited in its grounded
facts. A completed evidence stage supplies
\[
 z_1=\frac{1}{150}\sum_q\mathbf{1}\{A_q=\mathrm{true}\},\qquad
 z_2=\frac{1}{150}\sum_q\frac{|G_q|}{\max(1,|E_q|)}.
\]
These are execution-derived proxies. They contain no gold answer, answer
correctness or partial-panel F1. An invalid native assessment contributes
its recorded false judgment and empty evidence claims. A missing required
assessment is recorded as an execution failure.

Advancing a sample restores all six-action checkpoints and executes evidence
retention, context ordering and answer generation. All 150 predictions must
exist before the official answer F1 is calculated. This terminal observation
alone can enter the training shortlist. A saved evidence sample can support
multiple distinct suffixes under the common persistent execution service.

Five complete warmup workflows, each with a distinct evidence prefix,
initialize the cascade. A warmup prefix is followed by its chosen terminal
suffix. Subsequently, two independent RBF GPs predict $z_1,z_2$ from prefix
features; the terminal GP predicts F1 from the full configuration and both
proxies. Including configuration features in the terminal GP retains
pipeline identity beyond the two aggregated evidence observations. For a
new prefix, 256 normal Monte Carlo draws propagate both proxy uncertainty
and terminal uncertainty. Advancing an observed sample conditions on its
measured proxies. The candidate groups comprise at most 256 new-prefix
paths and every unseen legal suffix of each saved sample.

The selection rule first maximizes Monte Carlo EI relative to the best
completed training F1. If the largest EI is at most 0.005, it uses the
source qUCB fallback. For terminal draws $Y^{(d)}(x)$ and their sample mean
$\bar Y(x)$, the implemented single-candidate score is
\[
 U(x)=\frac{1}{256}\sum_{d=1}^{256}
 \left[\bar Y(x)+\sqrt{\frac{\pi\beta_{\mathrm{MSBO}}(C)}{2}}
                   |Y^{(d)}(x)-\bar Y(x)|\right],
\]
\[
 \beta_{\mathrm{MSBO}}(C)=1+100\max(0,1-C/B),\qquad B=32\mathrm{M}.
\]
Actual cumulative search cost is updated after a complete stage panel.
There is no cost division in this acquisition, consistent with the
authors' non-cost-weighted main configuration. Here resource awareness
enters the stopping account and the stated exploration schedule.

The source implementation also occasionally selects a complete cascade.
We retain its period of 1,000 acquisition decisions and map its
$100+\text{initial-data}$ cost epochs to $100+5=105$ normalized units.
The final full-path window is therefore $B/105$. When $B-C\leq B/105$, or
the acquisition count is a multiple of 1,000, candidates are scored through
the entire cascade posterior, even if a chosen prefix is already cached.
The selected complete tuple is then fixed. A new prefix executes both
stages; a materialized prefix proceeds directly to its selected suffix.
If the evidence stage crosses the budget, its fixed suffix still finishes,
with no intervening acquisition. This rule can cross both the resource
budget and the ordinary stage-count ceiling, because it completes one
already selected full-path decision.

The default ceiling is 256 completed workflows or 512 selected stages,
subject to that full-path completion rule. Every physical stage is billed
in order, including evidence samples never advanced. Primary results use
the terminal boundaries defined in Appendix~\ref{sec:adapter-common-contract}.
Additional stage snapshots retain paid evidence, completed-stage counts
and unfinished sample identities. Intermediate or crossing observations
cannot enter an earlier primary snapshot. Saved decisions and native
checkpoints allow a partially executed stage or fixed suffix to resume
without selecting against later data.

This implementation replaces continuous laboratory controls and measurements by
finite RAG actions and two evidence proxies. It uses scalar-lengthscale RBF
GPs and normal Monte Carlo draws in place of per-coordinate lengthscales
and Sobol sampling. The five-workflow warmup replaces a dimension-scaled
initial design. These choices preserve the start/advance decision and
cascade uncertainty calculation while defining a new application of MSBO.

\subsection{Surrogate computation and uniform-terminal control}
The EEIPU and LaMBO GPs use Mat\'ern-$5/2$ kernels on one-hot action
vectors; MSBO uses RBF kernels. Each target is centered and scaled, with a
minimum scale of $10^{-3}$. A scalar lengthscale and observation-noise
parameter are fitted by marginal likelihood when at least three
nonconstant observations are available, using at most 40 optimization
iterations. Log-lengthscale and log-noise bounds are $[-2.3,3]$ and
$[-13.8,-0.7]$, respectively; the diagonal numerical jitter is $10^{-9}$.
These are finite categorical surrogate choices. Optimizer-selection elapsed time is recorded as a single aggregate, separately from task execution and model-token charges. Fitting and acquisition CPU times remain undecomposed, and no timing comparison is reported here.

Uniform-terminal search provides a complete-workflow allocation control.
Exact descendant counts assign each legal terminal an integer rank. A
seeded sample of distinct ranks is uniform over terminals, even when
different first actions have different numbers of completions. The selected
paths undergo the same paid 150-question training evaluation, three budget
snapshots and validation selection. This differs from random-tree allocation,
whose local child choices induce a different terminal distribution, and
from the fifty frozen random workflows evaluated directly on test in
Section~\ref{sec:workflow-references}.

\section{Historical Generated-Workflow Diagnostics}
\label{sec:gsm-history}
These small pilots and retired formal attempts explain the development of
the common execution contract. Their scores, bills and traces are retained
as historical evidence and are excluded from the current fifteen-run GSM8K
quality summary.
The retained finite-catalogue extension generates no proposals: its outer
generation cost is zero, and the selected program fixes the logical stage
sequence. Its predictor sums physical per-stage means over missing
query--prefix resources, while WTB checks actual invocation-prefix reuse as
call inputs become known. Neither that predictor nor the dynamic pilot fits
prices to validation or test receipts. These settings have separate budgets,
configurations and evidence; the two completed pilot arms both select the
initial workflow and score 22/24, without a search-quality improvement.
\subsection{Exploratory dynamic AFlow pilot}
The completed GSM8K pilot uses LLM-generated Python edits. Two arms pair seed 11: a bounded search policy using AFlow's
original allocator and a program-edit search policy using AgentUCT. Both use
Qwen3.5-4B, the upstream proposal template with its XML output contract,
the actual operator interfaces and the official scorer. Each workflow is
evaluated on eight training questions before a tree update. The two best
distinct training workflows enter an eight-question validation panel; its
winner is measured on 24 primary questions, excluding one separate sentinel.
The panels are disjoint, but were previously sampled for development.

The single reporting budget is 100k raw input-plus-output tokens, with caps
of 32 complete workflows and 32 proposal rounds, each allowing up to three
proposal or repair attempts. Complete evaluations can cross the budget;
the reported snapshot is the last affordable complete state. The full bill
retains that crossing, proposals, repairs, validation and paid failures.
Test costs are separate. The diagnostic utility is
$J_{\mathrm{pilot}}=F-0.05C_{\mathrm{opt,full}}/8192$.
This diagnostic uses the inherited scale 8192 and $\beta=2$; it is reported
separately from RAG's parameter-weighted-token utility.

AFlow samples eligible top-four parents with probabilities
$0.3/K+0.7\,\mathrm{softmax}(20s)_i$. AgentUCT uses progressive widening,
raw descendant-return backup and a prequential estimate of complete edit
cost from earlier completed training proposals, repairs and executions.
The widening limit is $\min(8,\max(1,\lceil\sqrt{N}\rceil))$, with maximum
edit depth 12, exploration coefficient 1.4 and no rollout. Proposals use
temperature 0.7 and top-$p=0.9$; execution is greedy. The context cap is 8192
tokens, with output caps of 2048 for proposals and 1024 for execution.
Each question--workflow execution permits ten model-request slots, counting
retries and cache replays. Each formatted-model invocation permits up to three
format attempts within that limit; an operator can make several such invocations.
A 600-second deadline is checked between requests. An in-flight synchronous
model call settles before timeout handling.
Completed pilot receipts cover 276 physical calls and 251 native WTB result
checkpoints. A generated Programmer workflow executed Python on all eight
training questions. Both arms recorded zero model-response cache hits and selected the initial
workflow. The pilot compares search policies with bounded parent eligibility,
a single training evaluation, bounded repair, feedback and program
deduplication. Its generated-code cost estimates are point forecasts.

Each arm owns its execution cache and ledger. Python reruns while WTB restores
saved model responses at matching complete invocation prefixes. Successful
edits, adaptive selections, actual Programmer execution, failures, the full bill
and validation-to-test selection have been audited in the completed pilot.

\subsection{Retired fast-v2 context-overflow cohort}
The retired fast-v2 GSM8K cohort is separate from the counted-v3 formal comparison. In its diagnostic GSM8K
seed-11 trace, completion pricing also changed a decision below the root.
Through nineteen completed AgentUCT training panels, twenty-one recorded
traversal decisions included eight below the root, two at selection-state
depth two. The cost term changed seventeen choices overall, including five
below the root. MCTS-UCT completed nineteen panels and recorded seventeen
traversal decisions, four below the root. These traces show recursive
allocation in the retired cohort.

Search in this execution cohort terminated before its first reporting budget.
Repeated per-question errors inflated the proposal history beyond the fixed
6,144-token input allowance; repeated context rejections then exhausted the
256-proposal limit. AFlow, MCTS-UCT and AgentUCT completed 20, 19 and 19 training
panels at 4.504M, 5.891M and 5.818M raw search tokens, respectively, all below
6.423M. These runs are diagnostic evidence, not completed target-budget trials.
A separate common cohort encodes identical errors and consecutive identical
unevaluated failures with their occurrence counts. It retains every distinct
error and evaluated sibling, the complete task and parent program, and the
original context limit, model, data, budgets and allocation rules. This change
alters proposal inputs and therefore requires fresh runs for all three methods;
old and new measurements are not pooled.

Table~\ref{tab:gsm-nonroot-choice} shows a cost-induced depth-one decision in the retired fast-v2 diagnostic run.
At parent node 5, the quality-plus-exploration index favoured node 7.
Its estimated complete edit-and-evaluation cost was 1.382M raw tokens, compared with
0.249M for node 6, so AgentUCT selected node 6 as the next edit's input.
The resulting new program, node 9, scored zero on its complete 150-question
training panel. The trace therefore establishes that completion pricing
affected recursive allocation; it supplies no quality-gain evidence for this
particular choice. Candidate statistics are taken from the contemporaneous
saved traversal. Selection-state depth follows saved parent links and is
distinct from the depth of the generated program.

\begin{table}[t]
\centering
\caption{A saved depth-one AgentUCT decision in the retired fast-v2 GSM8K
seed-11 diagnostic run, edit 9. Both candidates are children of node 5. The cost-free
index selects node 7; subtracting $2\widehat b/C_{\rm ref}$ selects node 6,
with $C_{\rm ref}=150{,}954.5$ raw tokens. The edited child subsequently scored
zero on the training panel.}
\label{tab:gsm-nonroot-choice}
\small
\begin{tabular}{lrrrrrr}
\toprule
Node & $N$ & $Q$ & Bonus & $Q+\mathrm{bonus}$ & Est. cost (M) & Net index\\
\midrule
6 (selected) & 2 & 0.3067 & 1.2559 & 1.5626 & 0.2494 & $-1.7424$\\
7 & 1 & 0.2533 & 1.7761 & 2.0294 & 1.3822 & $-16.2831$\\
\bottomrule
\end{tabular}
\end{table}

\fi
\section{Formal Verification and Reproduction Map}
\label{app:lean}
The project pins Lean 4.24.0 and mathlib v4.24.0, with unified entry
\path{theory/lean/AgentUCTPaper.lean}.

\texttt{TreeGraphInformation},
\texttt{TreeDemandQuotient} and \texttt{ReachableDemandSeparation} formalize
reuse information and demand distinctions. \texttt{TreeFiniteBudget} and
\texttt{TreeDecisionValue} formalize the budget and local-choice results.
\texttt{TreeSharedCache}, \texttt{TreeNextUnpaid}, \texttt{DynamicCompletion}
and \texttt{OnlineMean} connect adaptive predictions, shared publication and
whole-tree bounds. The proof dependency inventory is under
\path{theory/evidence/reviewer-revision} and \path{theory/evidence/p0-p1}.

The main theorem quantifies its instance gap and constants before all finite event histories. It discharges child-deficit bounds by induction, without assuming a concentration event, logarithmic drift budget, or desired root-regret conclusion.

The online implementation mapping uses \texttt{OnlineMean}, whose envelope and
sampled-law lemmas connect incremental means and finite completion samples to
\leanname{online_mean_shared_root_log_bound}. The integrated build is recorded
in \path{theory/evidence/algorithm/lean-build.log}; dependencies are listed in
\path{theory/evidence/algorithm/lean-axioms.log}. The source declarations define
the formal claims. Python refinement is not machine-verified. The historical
event audits in \path{evidence/acquisition_update} and
\path{evidence/theory_alignment} concern their recorded runtime revisions.
Current runtime regressions and receipt evidence are under
\path{evidence/main-pair-v1}; full reconstruction of the current experimental
trajectories remains a separate post-run analysis. Paths are relative to the
workspace root.

The shared-cache theorem permits a fixed finite family of query panels with separate reward statistics. Its state-conditioned extension requires successful complete evaluations, persistent local cache entries, a fixed finite nonnegative prediction envelope, and deterministic reward tables. Arbitrarily changing the queries behind a shared reward mean, eviction, stochastic rewards, and action-dependent recurring penalties lie outside that theorem. The archived conditional stochastic results retain their stated assumptions.

\paragraph{Machine-checked proof sources.}
The mathematical sources use the standard logical axioms \texttt{propext},
\texttt{Classical.choice} and \texttt{Quot.sound}. The source modules contain
the proof bodies; \texttt{AgentUCTProofGuideEN} lists their declarations in
dependency order. Proof logs and the inventory are under
\path{theory/evidence/proof-guide-en-20261004}.

\subsection{Scope of Checked Results}
\label{app:results-extensions}
The proofs establish finite-tree quality rates under fixed rewards and
persistent successful publication, exact execution accounting and full-panel
cost quotients, and same-state pricing certificates with explicit error
assumptions. Training, conditional validation and finite-history dynamic
results retain their stated assumptions.

Equal-budget test F1 remains an empirical comparison. The rate counts
completed evaluations, including free replays. A frozen empty-cache penalty
need not vanish after coverage; boundedness alone does not transfer the
theorem to that ablation.
\ifdefined\AgentUCTProofOnly\else
\section{AI Assistance}
\label{app:ai-assistance}
OpenAI Codex assisted with Lean proof development, manuscript writing and
supporting code. Retained development records identify \texttt{gpt-6-astra}
and \texttt{gpt-6.1-sol}, with client versions 0.159.0, 0.159.2 and 0.160.0.
The original prompts, including methodological interactions, are supplied in
\path{evidence/main-pair-v1/ai-assistance-disclosure-20261004.json}.
The framework diagram is a Matplotlib vector drawing. The authors checked
formal statements against Lean outputs and empirical claims against saved
evaluations and receipts, and take responsibility for the work.
\fi

\clearpage
\bibliography{agenticOpt}
\end{document}